\documentclass{article} 
\usepackage{colm2024_conference}

\usepackage{booktabs}
\usepackage{graphicx}
\usepackage{enumitem}
\usepackage{wrapfig}
\usepackage{algorithm}
\usepackage{algpseudocode}
\usepackage{natbib}
\usepackage{makecell}
\usepackage{booktabs}
\usepackage{array}
\usepackage{amsmath}
\usepackage{amssymb}
\usepackage{amsfonts}
\usepackage{multirow}
\usepackage{verbatim}
\usepackage{caption}
\usepackage{longtable}
\usepackage{supertabular}
\usepackage{CJKutf8}
\usepackage[utf8]{inputenc} 
\usepackage[T1]{fontenc}
\usepackage[french,vietnamese,mongolian,greek,english]{babel}
\usepackage{pifont}
\usepackage{tikz}
\usepackage{multirow}

\usepackage{enumitem}
\usepackage{tablefootnote}
\usepackage{xspace}
\usepackage{textcomp}
\usepackage{makecell}
\usepackage{lscape} 
\usepackage{siunitx}
\usepackage{listings}
\usepackage{xcolor}
\usepackage{svg}
\definecolor{dt}{gray}{0.7}
\newcommand{\eg}{\textit{e.g.}\xspace}

\usepackage{pifont}       
\usepackage{bbding}       
\usepackage{fontawesome}
\usepackage{placeins}

\usepackage{scrextend}

\usepackage{tgpagella}
\usepackage{latexsym}
\usepackage[T1]{fontenc}
\usepackage[utf8]{inputenc}
\usepackage{microtype}
\definecolor{mydarkblue}{rgb}{0,0.08,0.45}
\definecolor{citecolor}{HTML}{0071BC}
\usepackage{url}            
\usepackage{nicefrac}       
\usepackage{changepage}
\usepackage{xargs}          
\usepackage{wrapfig,lipsum,booktabs}
\usepackage{longtable}
\usepackage{subcaption}
\usepackage{endnotes}

\usepackage{pgfplots}
\usetikzlibrary{pgfplots.groupplots,fit,backgrounds}
\pgfplotsset{compat=1.3}
\usepackage{tikz}
\usetikzlibrary{patterns}

\usepackage{hyperref}
\definecolor{darkblue}{rgb}{0, 0, 0.5}
\hypersetup{colorlinks=true, citecolor=darkblue, linkcolor=darkblue, urlcolor=darkblue, bookmarksopen=false, bookmarksnumbered=true}

\usepackage[most]{tcolorbox}

\usepackage[capitalize,noabbrev]{cleveref}

\crefname{section}{\S}{\S\S}
\Crefname{section}{\S}{\S\S}
\crefname{subsection}{\S\S}{\S\S}
\Crefname{subsection}{\S\S}{\S\S}
\crefformat{section}{\S#2#1#3}
\crefformat{subsection}{\S\S#2#1#3}

\crefname{table}{Table}{Tables}
\crefname{figure}{Figure}{Figures}
\crefname{algorithm}{Algorithm}{}
\crefname{equation}{eq.}{}
\crefname{appendix}{Appendix}{}
\crefformat{section}{Section #2#1#3}
\usepackage{multicol}
\usepackage{tcolorbox}
\usepackage{titlesec}
\titleformat*{\section}{\large\bfseries}
\usepackage{array} 
\newcolumntype{P}[1]{>{\centering\arraybackslash}p{#1}} 
\usepackage{multirow}

\usepackage[utf8]{inputenc} 
\newcolumntype{L}[1]{>{\raggedright\let\newline\\\arraybackslash\hspace{0pt}}p{#1}}
\newcolumntype{C}[1]{>{\centering\let\newline\\\arraybackslash\hspace{0pt}}p{#1}}
\usepackage{tikz}
\usetikzlibrary{backgrounds, fit}  

\usepackage{adjustbox}
\definecolor{objblue}{RGB}{3,139,221}  
\definecolor{attrred}{RGB}{255,67,67}    
\definecolor{easygreen}{RGB}{0,156,75}  
\definecolor{middleyellow}{RGB}{242,89,34}  
\definecolor{hardred}{RGB}{216,56,58}
\usepackage{colortbl}
\usepackage{array}
\usepackage{arydshln}

\usepackage[most]{tcolorbox}
\definecolor{BoxBackground}{RGB}{240, 240, 240} 
\definecolor{BoxFrame}{RGB}{0, 0, 0} 
\definecolor{TitleBackground}{RGB}{0, 0, 0} 
\definecolor{TitleText}{RGB}{255, 255, 255} 
\tcbset{
  academicbox/.style={
    boxsep=5pt,
    left=2pt,
    right=2pt,
    bottom=0.5pt,
    boxrule=0.5pt,
    colback=BoxBackground,
    colframe=BoxFrame,
    colbacktitle=TitleBackground,
    coltitle=TitleText,
    enhanced,
    attach boxed title to top left={yshift=-0.1in,xshift=0.1in},
    boxed title style={boxrule=0pt,colframe=white},
    title={#1},
  }
}
\newtcolorbox{AcademicBox}[1][]{academicbox=#1}

\title{Qwen-RobotManip Technical Report: Alignment Unlocks Scale for Robotic Manipulation Foundation Models}

\author{
\bf Qwen Team}

\newcommand{\ours}{\textsc{Qwen-RobotManip}\xspace}

\begin{document}

\maketitle

  
\begin{abstract}
Foundation models in language and multimodality achieve strong generalization because heterogeneous data sources can be aligned under a unified formulation, and abundant low-cost data from the internet allows diverse training signals to reinforce one another at scale.
In this report, \textbf{we investigate whether this scaling recipe can be applied to robotic manipulation to achieve genuine generalization.}
This is challenging because, unlike text, manipulation data is heterogeneous by nature,
expensive to collect, and narrow in diversity, making alignment and scale simultaneously difficult to achieve.
We present \ours, a generalizable Vision-Language-Action foundation model built upon Qwen-VL.
\ours introduces a unified alignment framework across the representation, motion, and behavioral dimensions of manipulation, making large-scale multi-source training coherent rather than conflicting.
This alignment capability in turn enables \ours to absorb manipulation data at a scale that prior training regimes could not sustain.
To provide a scaling engine for manipulation data, a human-to-robot synthesis pipeline converts egocentric hand demonstrations into robot trajectories across 15 platforms, and a rigorous curation pipeline harmonizes heterogeneous real-robot and synthetic datasets.
To our surprise, by leveraging \textbf{only} open-source robotic manipulation datasets and human demonstration videos without any proprietary data collection, \ours constructs a $\sim$38,100-hour pretraining corpus and already exhibits emergent generalization capabilities, including zero-shot instruction following, robustness to perturbations, reactive error recovery, and cross-embodiment knowledge transfer.
In experiments, we further find that most standard benchmarks systematically fail to capture the quality of pretraining. Thus, we instead adopt OOD evaluation settings, including RoboCasa365, LIBERO-Plus, EBench, RoboTwin-Clean2Rand, RoboTwin-IF (our new instruction-following benchmark), and RoboTwin-XE (our new cross-embodiment transfer benchmark), as our north star for measuring genuine generalization.
\ours achieves substantially better performance than prior state-of-the-art models, including $\pi_{0.5}$, across all OOD settings, ranks 1st in RoboChallenge with a \textbf{20\%} relative improvement, and is validated on real-robot platforms including AgileX ALOHA, Franka, UR, and ARX.

\end{abstract}

\begin{figure}[!h]
    \vspace{10pt}
    \centering
    \includegraphics[width=1.0\linewidth]{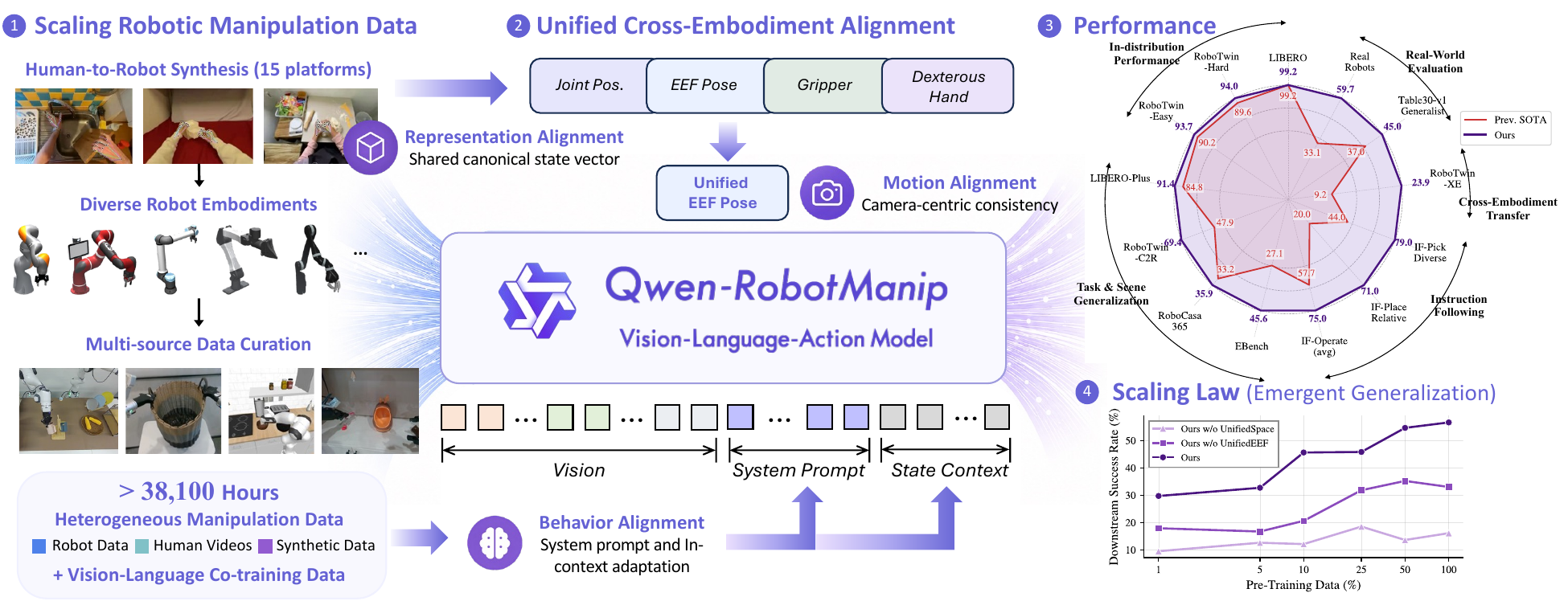}
    \vspace{-3pt}
    \label{fig:teaser}
\end{figure}

\clearpage

\section{Introduction}
\label{sec:intro}


Foundation models in language and multimodality~\citep{gpt3,gpt4,dubey2024llama,gemini,yang2025qwen3,qwen3-vl,qwen35blog} achieve strong generalization because heterogeneous data sources can be aligned under a unified formulation, and abundant low-cost data from the internet allows diverse training signals to reinforce one another at scale. 
In this report, we investigate whether this scaling recipe can be applied to robotic manipulation to achieve genuine generalization. 
This is challenging because, unlike text, manipulation data is heterogeneous by nature, expensive to collect, and narrow in diversity, making alignment and scale simultaneously difficult to achieve.


The current state of Vision-Language-Action (VLA) models~\citep{openvla,bjorck2025grootn1,black2024pi0,pi05,community2026starvla,ddpvla,kim2025fine} illustrates how far this gap remains. 
Despite a rapid succession of models reporting competitive numbers on standard benchmarks~\citep{liu2023libero,nasiriany2024robocasa,mu2025robotwin}, the generalization being demonstrated is largely superficial~\citep{zhang2026world}. 
OOD evaluations in most works involve only minor visual perturbations while preserving the same embodiment, task structure, and workspace layout as data collection, and performance degrades sharply when models are tested beyond these narrow settings.

The reason these pretrained priors fail to transfer is twofold. 
First, existing robotic demonstration corpora~\citep{padalkar2023openxembodiment,khazatsky2024droid,fang2023rh20t} are concentrated in narrow teleoperation setups, far too limited in embodiment and task diversity for a scaling recipe to take hold. Second, and more fundamentally, data diversity alone is insufficient without alignment~\citep{luo2026joint, wang2026rethinking}.
When demonstrations from different embodiments arrive with incompatible observation and action representations, scaling data volume produces interference rather than synergy. Prior cross-embodiment efforts~\citep{bjorck2025grootn1,zheng2025xvla,pi05} have adopted shared architectures, embodiment tokens, or unified action tokenization, but without a formulation that makes the same physical motion numerically consistent across embodiments, additional data cannot be converted into improved capability.
Alignment is therefore not an independent engineering choice but a prerequisite for data scaling itself.

We present \textbf{\ours}, a Vision-Language-Action foundation model built upon Qwen-VL~\citep{yang2025qwen3, qwen3-vl,qwen35blog}, designed around this principle: \emph{alignment first, then scale}. 
\ours introduces a unified alignment framework across the representation, motion, and behavioral dimensions of manipulation, making large-scale multi-source training coherent rather than conflicting: a canonical state-action representation with per-dimension binary masking accommodates diverse robot morphologies within a single template, a camera-frame delta pose parameterization makes visually similar motions numerically proximate across coordinate frames, and an in-context policy adaptation mechanism reads intra-episode execution history as an implicit embodiment identifier for kinematic-aware behavioral adjustment. 
Training is conducted under a dual-stream co-training strategy that jointly optimizes over manipulation data and a curated vision-language stream, preventing the VLM backbone's perceptual and reasoning capabilities from eroding under action prediction pressure.

This alignment capability in turn enables \ours to absorb manipulation data at a scale that prior training regimes could not sustain. 
A human-to-robot synthesis pipeline converts egocentric hand demonstrations into robot trajectories across 15 platforms, and a rigorous curation pipeline harmonizes heterogeneous real-robot and synthetic datasets, together accumulating $\sim$38,100 hours of manipulation data. 
Notably, this entire corpus is constructed from \textbf{only} open-source robotic manipulation datasets and egocentric human videos without any proprietary data collection, yet \ours already exhibits emergent generalization capabilities, including robustness to perturbations, zero-shot instruction following, reactive error recovery, and cross-embodiment transfer.

We further argue that the field's evaluation methodology must evolve alongside its models. Standard in-domain benchmarks, where models without large-scale robot pretraining match or exceed pretrained ones~\citep{Yan_2025_ICCV,community2026starvla}, systematically fail to distinguish genuine generalization from in-distribution pattern memorization.
\ours therefore introduces OOD evaluation settings including LIBERO-Plus~\citep{fei2025liberoplus}, RoboTwin-Clean2Rand~\citep{mu2025robotwin}, RoboCasa365~\citep{nasiriany2026robocasa365}, and EBench~\citep{ebench2026}, alongside two new benchmarks: \textbf{RoboTwin-IF}, an instruction-following benchmark that tests whether policies condition on language as a genuine control signal rather than exploit visual shortcuts, and \textbf{RoboTwin-XE}, which evaluates zero-shot transfer to morphologically distinct robots. 
\ours achieves state-of-the-art on standard benchmarks and substantially outperforms existing VLA models including GR00T-N1.7 and $\pi_{0.5}$ across all OOD evaluation axes, ranks 1st on the RoboChallenge Table30-v1 generalist track with a 20\% relative improvement, and is validated on various real-robot platforms and tasks. We hope these results and benchmarks together raise the standard for how VLA generalization is measured across the field.

Our contributions are as follows.

\textbf{A unified alignment framework for cross-embodiment training.} We address representational heterogeneity through three complementary mechanisms. A canonical state-action representation with per-dimension binary masking, a camera-frame delta pose parameterization that grounds end-effector actions in the visual domain, and an in-context policy adaptation mechanism that treats intra-episode execution history as an implicit embodiment identifier together enable consistent signal extraction across diverse embodiments. A dual-stream co-training strategy jointly optimizes manipulation and vision-language objectives to preserve the perceptual and reasoning capabilities that underpin generalization.

\textbf{A scalable multi-source data corpus.} We consolidate $\sim$38,100 hours of manipulation data from open-source robot datasets and egocentric human demonstrations. A human-to-robot synthesis pipeline converts any egocentric demonstration into trajectories across 15 robot platforms, providing a scalable and embodiment-rich data source. A multi-stage curation pipeline ensures signal quality across all heterogeneous sources.
Beyond manipulation data, a curated vision-language mixture including novel embodied chain-of-thought and egocentric video understanding data preserves the VLM backbone's perceptual and reasoning capabilities during VLA training.

\textbf{A new standard for evaluating VLA generalization.} We introduce OOD evaluation settings including LIBERO-Plus, RoboTwin-Clean2Rand, RoboCasa365, and EBench, alongside RoboTwin-IF, a benchmark that diagnoses genuine language conditioning, and RoboTwin-XE, a benchmark for zero-shot cross-embodiment transfer. We argue that in-domain metrics are insufficient proxies for foundation model capability and that OOD transfer is the correct measure. \ours substantially outperforms existing VLA models across all OOD settings while achieving state-of-the-art on standard benchmarks.

\textbf{Real-robot validation across diverse deployment scenarios.} 
We validate \ours on four physical platforms (AgileX ALOHA, Franka, UR, and ARX) across in-domain, out-of-domain, few-shot adaptation, and zero-shot cross-embodiment transfer settings.
On the RoboChallenge Table30 v1 generalist track, \ours ranks 1st.
Results confirm that the generalization capabilities of \ours hold under real-world deployment conditions.

\section{Data Sources for Robotic Manipulation}
\label{sec:data}

The quality and diversity of training data are foundational to the generalization ability we seek.
To build such a VLA model with strong generalization across embodiments, tasks, and environments, we curate a large-scale heterogeneous training corpus 
comprising three complementary data modalities: robotic manipulation demonstrations across diverse hardware platforms, egocentric human manipulation videos, and synthetic robot data generated by our human-to-robot pipeline.
A unified curation and pre-processing pipeline processes all sources to ensure high-quality and consistent state, action, video, and language annotations.
Table~\ref{tab:data_overview} summarizes the composition of the full corpus.


\begin{table}[h]
\centering
\caption{\textbf{Overview of the training data corpus.}}
\vspace{-4pt}
\label{tab:data_overview}
\tabcolsep 3pt
\resizebox{\textwidth}{!}{
\begin{tabular}{lllll}
\toprule
\textbf{Data Type} & \textbf{Embodiment Type} & \textbf{Data Sources} & \textbf{Task Setting} & \textbf{Total Time} \\
\midrule
\multirow{3}{*}{Robot}
  & Single-arm         & \multirow{3}{60mm}{\raisebox{-2.9ex}{\parbox{60mm}{\{OXE, RoboMIND, DROID, RH20T, AgibotWorld-Beta, RoboCOIN, RDT, InternData-A1, Galaxea Open-World\}}}}
                                          & Tabletop              & 3,808 h \\[2pt]
  & Dual-arm           &                 & Tabletop              & 6,744 h \\[2pt]
  & Mobile \& humanoid &                 & Tabletop \& indoor    & 868 h   \\
\midrule
Human          & Human hands           & EgoDex, VITRA, EgoVerse      & Tabletop \& in-the-wild & 1,933 h  \\
\midrule
Human-to-Robot & 15 dual-arm platforms & Synthesized from human data. & Tabletop \& in-the-wild & 24,808 h \\
\bottomrule
\end{tabular}}
\end{table}

\subsection{Robotic Datasets}

Robot manipulation demonstrations constitute the core of our pretraining corpus, spanning single-arm and bimanual tabletop manipulation, dexterous manipulation, mobile manipulation, and humanoid loco-manipulation in both simulation and real world.
We incorporate nine open-source datasets, totaling over 11,000 hours of demonstrations.

\textbf{Open X-Embodiment (OXE)}~\citep{padalkar2023openxembodiment} aggregates real-world robotic datasets from diverse research institutions.
We retain three subsets (\emph{Fractal}, \emph{Bridge}, and \emph{BC-Z}) of high-quality single-arm tabletop manipulation data across the Google Robot and WidowX platforms, contributing about 600 hours.

\textbf{AgiBotWorld-Beta}~\citep{bu2025agibotworld} is a large-scale real-world humanoid manipulation dataset collected on the AgiBot G1 bimanual platform. We use datasets collected with grippers, covering about 200 task types and 2,400 hours.

\textbf{RoboMIND}~\citep{wu2024robomind} and \textbf{RoboMIND 2.0}~\citep{wu2025robomind2} provide large-scale real-world datasets covering single-arm, dual-arm, ALOHA~\citep{zhao2023aloha}, and humanoid robots across diverse tabletop manipulation tasks. RoboMIND spans four embodiments including Franka Emika Panda, UR5e, AgileX Cobot Magic V2.0, and Tien Kung humanoid; RoboMIND 2.0 extends coverage to six platforms including Franka, UR5, AgileX, ARX, Tien Kung, and Tian Yi. Together they contribute about 1,400 hours of demonstrations.

\textbf{Galaxea Open-World Dataset}~\citep{galaxea2025openworld} provides $\sim500$ hours of real-world bimanual mobile manipulation demonstrations collected on Galaxea dual-arm robots across diverse household tasks.

\textbf{RoboCOIN}~\citep{wu2025robocoin} is a large-scale multi-embodiment real-world dataset covering a wide range of bimanual and humanoid platforms.
We retain 10 embodiment types including AgiBot G1, Airbot MMK2, Alpha Bot 2, AgileX Cobot Magic, Unitree G1edu, Leju, Realman R1 Lite, Realman RMC-AIDA-L, ALOHA, and Tianqin A2, resulting in about 430 hours of demonstrations.

\textbf{DROID}~\citep{khazatsky2024droid} is an in-the-wild single-arm dataset collected with Franka Panda robots across 86 diverse real-world environments, contributing about 95,000 trajectories totaling 500 hours.

\textbf{RH20T}~\citep{fang2023rh20t} is a large-scale contact-rich real-world dataset spanning 4 embodiments (Flexiv, UR5, Franka, and Kuka) with multi-modal sensing including visual, force-torque, audio, and proprioception. 
It covers 140+ tasks across 42 skill categories, resulting in about 1,100 hours of demonstrations.

\textbf{RDT-1B}~\citep{liu2024rdt1b} provides 29 hours of bimanual manipulation demonstrations collected on the ALOHA platform.

\textbf{InternData-A1}~\citep{tian2025interndataa1} is a large-scale dataset generated in high-fidelity simulation environments, covering various single-arm and dual-arm embodiments across pick-and-place, articulated manipulation, and long-horizon tasks, totaling over 3,600 hours.


\subsection{Egocentric Human Datasets}

Egocentric human hand manipulation data is naturally aligned with the perspective of robot-mounted cameras, serving as an efficient source for expanding manipulation data~\citep{kareer2025egomimic,qiu2025humanoid,egoscale,beingh0,beingh05,easymimic}.
We collect egocentric data from three sources with hand pose annotations.

\textbf{EgoDex}~\citep{egodex}, collected using Apple Vision Pro, contains 338K demonstrations across 194 tabletop manipulation tasks totaling 829 hours of 30\,Hz egocentric video.
It provides SE(3) annotations for 25 joints of both hands per frame, tracked on-device using multiple calibrated cameras and visual--inertial SLAM.
We utilize 732 hours for training.

\textbf{VITRA}~\citep{vitra} performs fully automated 3D hand reconstruction, camera trajectory estimation, and atomic action segmentation on unstructured egocentric videos.
It draws from five video sources: Ego4D~\citep{grauman2022ego4d} (cooking-and-cleaning and general activity subsets), EPIC-KITCHENS~\citep{damen2018epic}, EgoExo4D~\citep{egoexo4d}, and Something-Something~v2~\citep{ssv2}, totalling approximately 1M trajectories.
We utilize the Ego4D and EPIC-KITCHENS subsets, contributing about 247 hours of video.

\textbf{EgoVerse}~\citep{egoverse} is a large-scale collaborative egocentric manipulation dataset spanning 1,362 hours across 1,965 tasks, 240 scenes, and 2,087 demonstrators.
Hand poses (21 keypoints per hand) and 6-DoF head poses are recovered via vision-based methods including visual--inertial SLAM and model-based pose estimation.
We utilize the industry-contributed portion, contributing approximately 954 hours of video.

The three sources collectively amount to approximately \textbf{1,933} hours of video.
All hand poses are converted into a unified representation of MANO~\citep{mano} parameters and 21 keypoints; for sources lacking native MANO annotations, parameters are recovered via optimization-based fitting.



\subsection{Human-to-Robot Data Synthesis}

\begin{figure}[t]
    \centering
    \includegraphics[width=0.98\linewidth]{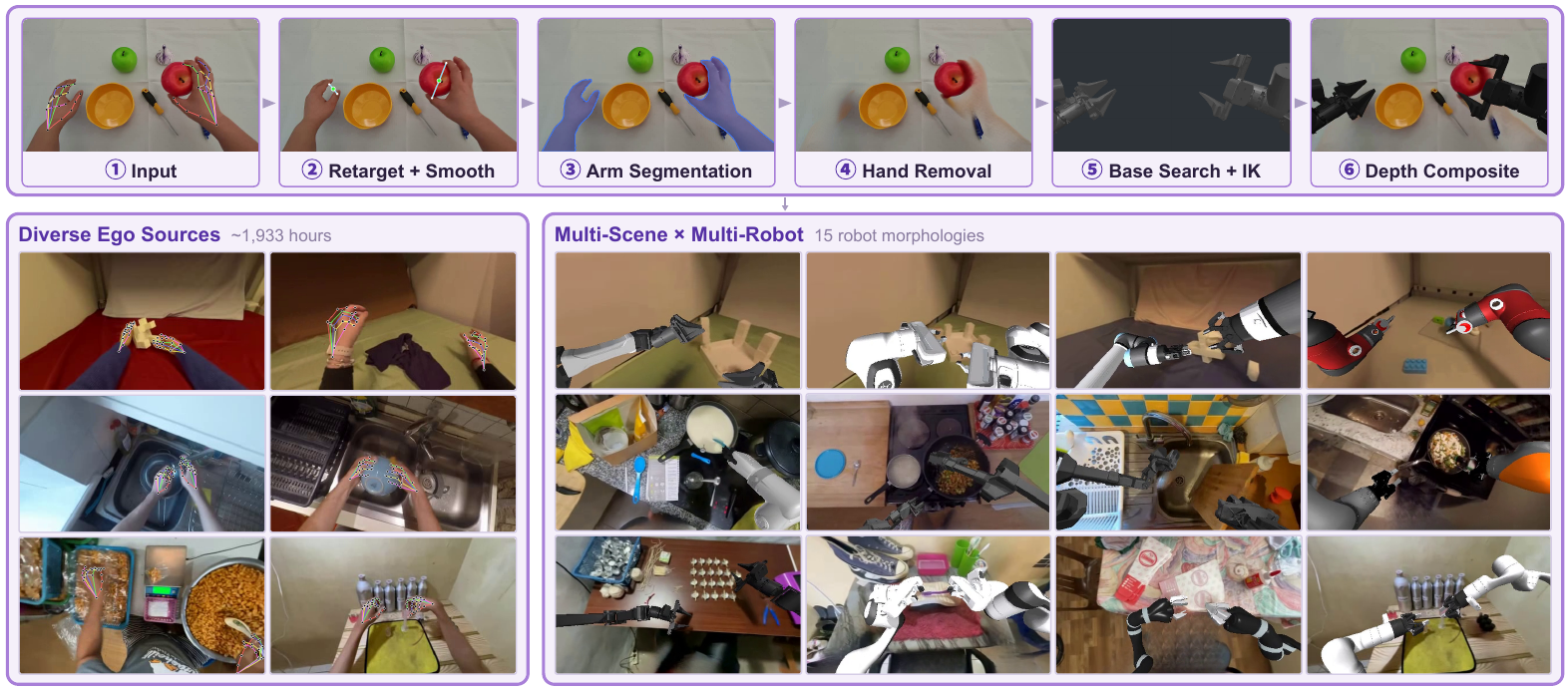}
    \vspace{-5pt}
    \caption{\textbf{Human-to-robot data synthesis pipeline.} 
    \textit{(Top)} Given egocentric video, the pipeline performs action retargeting and smoothing, SAM3-based hand segmentation, ProPainter inpainting, base pose search with MuJoCo IK, and depth-guided compositing. 
    \textit{(Bottom)} $\sim$1,933h of egocentric data from 3 sources is rendered across 15 robot morphologies, yielding $\sim$24,808h of synthesized demonstrations.}
    \label{fig:h2r_pipeline}
    \vspace{-1pt}
\end{figure}

A significant gap exists between egocentric human data and robot data in both morphology and visual domains. 
To bridge this gap, we map human hand trajectories to the robot action space, and replace human hands in videos with robot models. 
Inspired by prior work~\citep{lepert2025phantom,lepert2025masquerade}, we design an end-to-end synthesis pipeline that explicitly separates the process into action alignment and visual alignment (Figure~\ref{fig:h2r_pipeline}).

\textbf{Action Alignment.} 
This stage focuses on trajectory retargeting and smoothing to bridge the morphology gap between human hands and parallel-jaw grippers.
We define the robot action at frame $t$ as $\mathbf{a}_t = (\mathbf{p}_t,\,\mathbf{R}_t,\,w_t)$, where $\mathbf{p}_t\in\mathbb{R}^3$ is the end-effector position, $\mathbf{R}_t\in SO(3)$ is the gripper orientation, and $w_t\in\mathbb{R}_{\geq 0}$ is the gripper width.
Using the 3D hand keypoints $\mathbf{k}_i$ from MANO, we define a virtual finger $\mathbf{k}_\text{vf}$ as a weighted combination of the index and middle fingertips. 
The end-effector position $\mathbf{p}_t$ is retargeted as the midpoint between the thumb tip and the virtual finger, and the gripper width $w_t$ is their Euclidean distance:
\begin{equation}
  \mathbf{k}_\text{vf} = 0.7\,\mathbf{k}_\text{index} + 0.3\,\mathbf{k}_\text{middle},\quad
  \mathbf{p} = \tfrac{1}{2}\bigl(\mathbf{k}_\text{thumb} + \mathbf{k}_\text{vf}\bigr),\quad
  w = \|\mathbf{k}_\text{thumb} - \mathbf{k}_\text{vf}\|_2.
  \label{eq:eef}
\end{equation}

The gripper orientation is constructed as a right-handed orthonormal frame $\mathbf{R}=[\mathbf{x}\;\mathbf{y}\;\mathbf{z}]$. We first establish the grasp axis $\mathbf{z}$ along the jaw-line direction (the line connecting the thumb tip and the virtual finger). 
Together with the wrist-to-fingertip direction $\mathbf{d}=\mathbf{k}_\text{vf}-\mathbf{k}_\text{wrist}$, these two vectors define the jaw plane; the gripper-normal axis $\mathbf{y}$ is the normal of this plane, and the approach axis $\mathbf{x}$ completes the right-handed frame:
\begin{equation}
    \mathbf{z} = \frac{s \cdot (\mathbf{k}_\text{thumb} - \mathbf{k}_\text{vf})}{w}, \quad
    \mathbf{y} = \frac{\mathbf{z} \times \mathbf{d}}{\|\mathbf{z} \times \mathbf{d}\|}, \quad
    \mathbf{x} = \mathbf{y} \times \mathbf{z}
\end{equation}
where $s=+1$ for the right hand and $s=-1$ for the left hand. This sign flip ensures that $\mathbf{z}$ points in a consistent direction regardless of handedness, so that both hands map to the same gripper frame. The three axes correspond to: $\mathbf{x}$ -- approach direction, $\mathbf{y}$ -- gripper normal (perpendicular to the jaw plane), $\mathbf{z}$ -- grasp axis (along the jaw line).

Per-frame hand detection introduces high-frequency noise. 
We apply Savitzky--Golay~\citep{savitzky1964smoothing} filtering to positions and widths, and Gaussian-weighted SLERP to orientations, producing smooth trajectories while preserving motion structure.

\textbf{Visual Alignment.} 
This stage replaces the human appearance with a robot model through a sequence of masking, inpainting, and rendering steps to bridge the visual domain gap. 
First, SAM3~\citep{carion2025sam3segmentconcepts} generates a binary mask $M_t\in\{0,1\}^{H\times W}$ for the human arm using text prompts. 
Next, ProPainter~\citep{propainter} inpaints the masked regions guided by optical flow, creating a clean background sequence $\{\hat{I}_t\}$ without human hands. 

A fundamental challenge in converting ego videos to robot data is determining the robot base placement. Unlike robot-to-robot transfer where a source base position is available, egocentric hand trajectories are embodiment-free: there is no physical robot base to reference. We formulate this as an optimization over base placements: given $N$ target end-effector poses $\{\mathbf{T}_i^\text{ee}\}_{i=1}^{N}$ and a robot with maximum reach $r_\text{max}$, we seek:
\begin{equation}
    \mathbf{T}_\text{base}^* = \arg\max_{\mathbf{T}_\text{base}} \frac{1}{|\mathcal{K}|} \sum_{k \in \mathcal{K}} \mathbb{1}\left[\text{IK}(\mathbf{T}_\text{base}^{-1} \mathbf{T}_k^\text{ee}) \text{ is feasible}\right]
\end{equation}
where $\mathcal{K} \subset \{1, \dots, N\}$ is a set of representative keyframes covering the spatial extremes of the trajectory. Candidate base placements are generated via grid search around the trajectory centroid, constrained by the per-morphology kinematic reach $r_\text{max}$. 
This search is performed independently for each of the 15 robot morphologies, as different arm lengths and joint configurations require different base placements for the same trajectory.

Given the optimized base pose, we run inverse kinematics in a MuJoCo~\citep{mujoco,Zakka_Mink_Python_inverse_2026} virtual environment to track the smoothed action trajectory, rendering the robot image $I_t^\text{robot}$ and its depth map $D_t^\text{robot}$. 
Depth Anything v3~\citep{da3} estimates a metric depth map $D_t$ for the scene. 
We compute an occlusion mask $M_t^\text{occ} = \mathbb{1}[D_t^\text{robot} \leq D_t]$ to naturally composite the robot onto the clean background:
\begin{equation}
  I_t^\text{syn} = M_t^\text{occ} \odot I_t^\text{robot}
                 + \bigl(1 - M_t^\text{occ}\bigr) \odot \hat{I}_t.
  \label{eq:composite}
\end{equation}

Each human demonstration is rendered into 15 bimanual robot configurations (each composed of two identical arms from: Panda, UR5e, ARX-L5, xArm7, Sawyer, Kinova~Gen3, IIWA, Jaco, FR3, UR10e, ViperX, WidowX, Piper, YAM, AgileX ALOHA), yielding approximately \textbf{24,808} hours of synthesized demonstrations in total.

\textbf{Action Speed Alignment.}
Egocentric hand manipulation exhibits significantly higher action speeds than robot teleoperation data. 
To align the action speed distributions, we apply per-source frame subsampling during training to match the robot data speed. 
Specifically, EgoDex is downsampled to 60\% of its original frame rate ($\sim$1.7$\times$ slower), EgoVerse to 45\% ($\sim$2.2$\times$ slower), and ViTRA to 25\% ($\sim$4$\times$ slower).

\begin{figure}[t]
    \centering
    \includegraphics[width=0.98\linewidth]{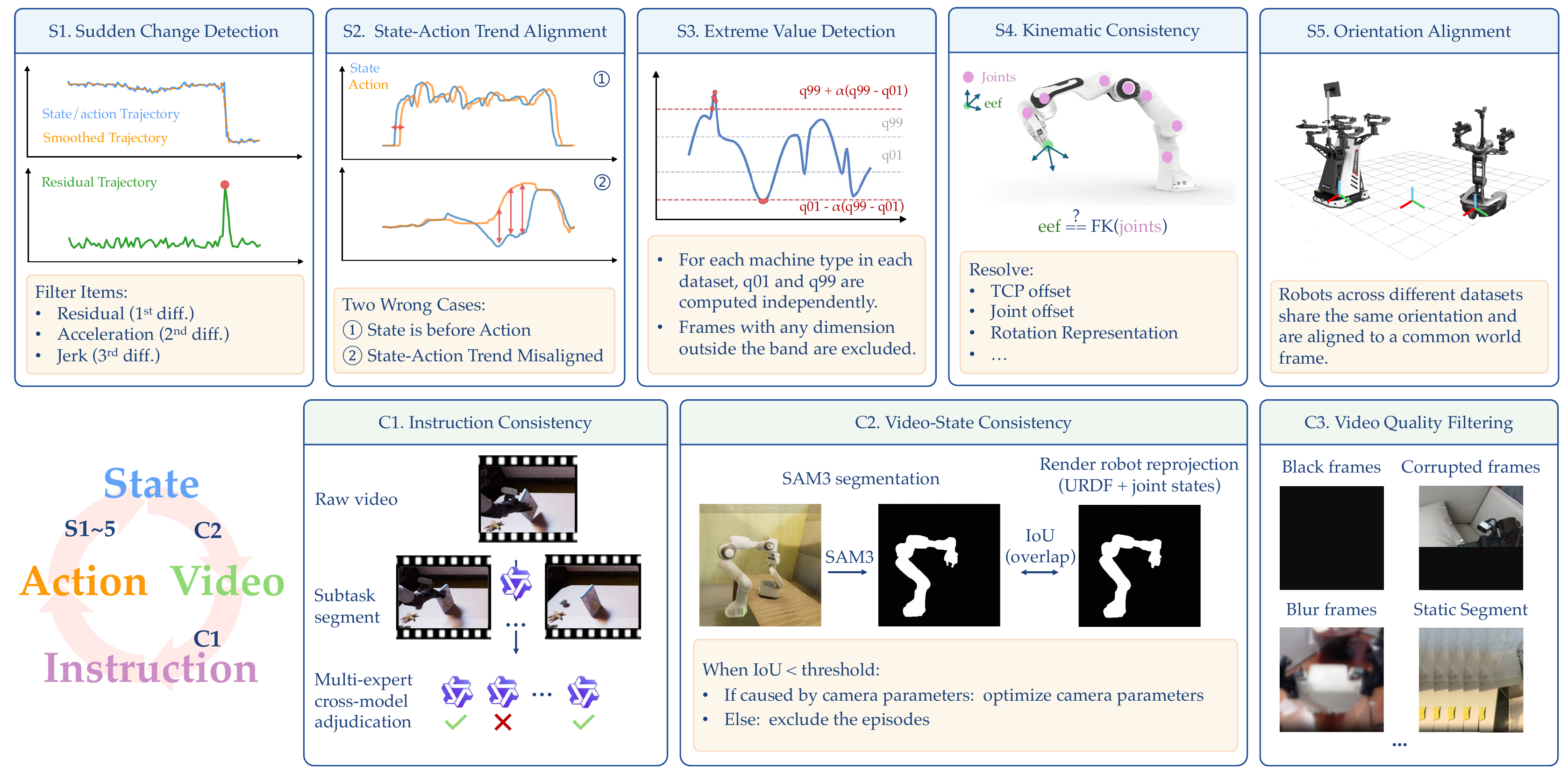}
    \vspace{-3pt}
    \caption{\textbf{Multi-stage data curation pipeline.} Five-stage state-action signal filtering (sudden change, trend alignment, extreme value removal, FK consistency, and base-frame alignment) followed by three cross-modal quality checks (instruction consistency, video-state consistency, and video quality filtering).}
    \label{fig:data preprocess}
\end{figure}

\subsection{Data Curation and Pre-Processing}
\label{sec:state_action_filtering}  
Aggregating manipulation data across diverse embodiments, simulators, and collection pipelines introduces heterogeneous noise in recorded state and action signals, including discrete outliers from physical collisions, temporal misalignment between state and action streams due to unsynchronized clocks or packet loss, extreme values that destabilize optimization, and inconsistent end-effector conventions across datasets sharing the same robot embodiment. We address these through a five-stage filtering pipeline applied to all datasets prior to training. 

\paragraph{Stage 1: Sudden Change Detection.}

For each signal dimension, we extract a smooth trend via cascaded median filtering and Savitzky--Golay smoothing~\citep{savitzky1964smoothing}, then compute three complementary deviation signals: the absolute residual between the raw and smoothed trajectory, the second-order finite difference (acceleration), and the third-order finite difference (jerk). A frame is flagged when the residual exceeds a scaled threshold \emph{and} either acceleration or jerk also exceeds its threshold, reducing false positives from slow drift while preserving sensitivity to abrupt transients. Thresholds are set per dataset according to embodiment type, rotation representation, data source (real vs.~simulation), and base mobility. Exclusion strategies range from frame-level removal to full episode discard. For instance, in InternData-A1~\citep{tian2025interndataa1} where sudden changes arise exclusively from physical collisions (\eg, a gripper contacting a rigid object), the corrupted episode is discarded entirely.


\paragraph{Stage 2: State-Action Trend Alignment.}
In a correctly recorded episode, action commands should temporally lead or coincide with resulting state changes, which is a causal invariant violated when timestamps are unsynchronized or there is packet loss. 
For each shared joint dimension, we smooth both the state and action trajectories, then estimate the optimal temporal lag via cross-correlation, then compute a \emph{directional agreement} (DA) metric on lag-aligned first-order differences. 
Dimensions with DA below a dataset-specific threshold (typically $0.6$-$0.7$) are flagged and their episodes excluded. For datasets using delta actions, we first integrate the action sequence to recover absolute values before comparison. This stage revealed severe quality issues in certain subsets: 81\% of episodes in the RoboMIND UR-type data failed this check and were excluded.


\paragraph{Stage 3: Extreme Value Filtering.}
Frames with state or action values outside the expected range are removed to prevent distortion of the quantile-based normalization ($[q_{01}, q_{99}] \to [-1, 1]$) used during training. Per-dimension $q_1$ and $q_{99}$ percentiles are computed per embodiment type, and frames outside the band $[q_1 - \alpha(q_{99}-q_1),\ q_{99}+\alpha(q_{99}-q_1)]$ are excluded. Gripper dimensions are exempt due to their bimodal distributions.


\paragraph{Stage 4: Joint-End-Effector Forward Kinematics Consistency.}
We compute forward kinematics (FK) via Pinocchio~\citep{carpentier2019pinocchio} from each robot's URDF and compare against logged end-effector poses. The discrepancies can arise from differing joint-angle sign conventions, differing end-effector frame definitions, incorrect rotation representations, incorrect base-frame assumptions, and erroneous end-effector logging. Rather than aggressively filtering, this stage primarily performs \emph{data correction}: constant positional offsets are resolved by adjusting the tool-center-point (TCP) definition, and shoulder-relative bimanual poses are transformed into the world frame. This process revealed that the same robot model can carry different joint-angle conventions across datasets, further motivating the unified state-action representation of Sec.~\ref{sec:unified_state_action}.


\paragraph{Stage 5: Base Frame and End-Effector Orientation Alignment.}
We apply per-dataset rotation corrections to align world-frame orientation conventions, ensuring the positive $x$-axis consistently corresponds to the robot's forward-facing direction and that the unified state-action representation is geometrically consistent across embodiments.


Beyond this five-stage state and action signal quality filtering, we apply three additional checks to ensure cross-modal consistency across video, language, and proprioceptive observations.


\paragraph{Check 1: Instruction Consistency.} 
We verify semantic consistency between each demonstration and its language annotation via a three-stage VLM-based pipeline. First, long episodes are decomposed into subtask-level segments~\citep{lei2026towards} so that each clip corresponds to a temporally localized action unit, keeping the visual evidence focused and tractable for automated assessment (Temporal Normalization for Evaluation Units). Second, each segment is evaluated through structured reasoning-guided prompting: rather than requesting an immediate binary label, the VLM is directed to attend to manipulated objects, action semantics, temporal ordering, and agent-environment interaction, producing an intermediate analytical judgment before issuing a final consistency decision. This structured prompting reduces superficial or heuristic responses and improves label interpretability (Structured Reasoning-Guided Annotation). Third, clips flagged as non-aligned or ambiguous by the initial model are adjudicated by multiple VLMs as independent evaluators, with the final label determined by cross-model voting, reducing single-model bias and improving label robustness (Multi-Expert Cross-Model Adjudication). Inconsistent samples are excluded from training.

\paragraph{Check 2: Video-State Consistency.} 
We perform rigorous data cleaning to remove low-quality or misaligned samples.
To verify video-state consistency, we render the robot projection into the image plane using the URDF and recorded joint states, segment the actual robot mask with a fine-tuned SAM3~\citep{carion2025sam3segmentconcepts} model, and measure their overlap. Samples with low overlap are filtered out.


\paragraph{Check 3: Video Quality Filtering}
We apply video-level data cleaning to remove frames that may degrade policy learning. We remove visually invalid frames including black, corrupted, blurred, and prolonged static segments, using image processing checks applied jointly with state and action signals to detect redundant static periods typically at episode boundaries. Task-critical key frames such as gripper closure events are explicitly preserved to avoid discarding visually subtle but semantically important transitions.






\subsection{Vision-Language Co-training Datasets}
\label{sec:vlm_data}

Prior works have demonstrated that co-training VLAs with vision-language (VL) data mixtures can significantly improve their generalization ability~\citep{pi05, driess2025knowledge, fang2026molmoact2}. By incorporating VL data during VLA training, the model retains the rich visual and semantic knowledge acquired from web-scale multimodal pretraining and transfers this knowledge to action generation through the action expert. For example, this enables the model to follow novel language instructions, operate in unfamiliar scene backgrounds, or manipulate previously unseen objects.

To this end, we curate a comprehensive VL dataset from multiple sources, including proprietary data, open-source datasets (e.g., RoboPoint~\citep{yuan2024robopoint}, RefSpatial~\citep{zhou2025roborefer}, PixMo~\citep{deitke2025pixmo}, and CapsFusion~\citep{yu2024capsfusion}), and carefully
synthesized embodied-centric data. The resulting VL mixture comprises approximately 28M data points, spanning the following categories:

(1) \textbf{General Visual Understanding}, including visual question answering, multi-image reasoning, and image captioning at varying granularities (from single-sentence summaries to paragraph-level detailed descriptions), which preserves the model's broad visual perception and commonsense reasoning capabilities;

(2) \textbf{Spatial Perception and Reasoning}, covering 2D/3D visual grounding, point localization, counting, spatial relationship reasoning (depth comparison, distance estimation, camera viewpoint inference), and manipulation feasibility reasoning, which are directly transferable to robotic spatial understanding;

(3) \textbf{OCR and Document Understanding}, which helps maintain the VLM's ability to recognize text, numbers, and symbols, a capability that is also required in robotic tasks involving labeled objects (e.g., identifying a block marked with a specific number);

(4) \textbf{Multimodal Specialized Knowledge}, covering domain-specific visual reasoning tasks such as STEM problem solving, chart interpretation, and visual puzzle reasoning, which helps prevent catastrophic forgetting of the VLM's general multimodal reasoning capabilities during VLA fine-tuning;

(5) \textbf{Instruction Following, Multilingual, and Pure Text data}, which strengthens the model's ability to follow diverse natural language instructions, a capability that is critical for generalizing to novel robot tasks, while also enabling multilingual robot control and preserving text generation quality;

(6) \textbf{Embodied-Centric VL Data}, which we specifically curate to bridge the gap between web-scale VL knowledge and robotic manipulation. This subset includes:   (a) embodied chain-of-thought (ECoT) reasoning data derived from robot manipulation trajectories, where the model performs structured reasoning in three stages: first describing the current scene state from multi-view observations (including gripper status, object positions, and spatial layout), then assessing task progress by comparing the current state against the overall goal, and finally predicting the next atomic manipulation action; (b) egocentric video understanding data, where the model describes fine-grained hand/arm movements, hand-object interactions, and object state changes from short clips of first-person human manipulation videos; and (c) 2D trajectory prediction data, where the model predicts future movement trajectories of human hands or robot end-effectors as sequences of normalized 2D coordinates, conditioned on visual observations and task instructions.

Among these, categories (1)–(5) serve a dual purpose: they prevent catastrophic forgetting of the pretrained VLM's general capabilities, while certain subsets, such as spatial reasoning, visual grounding, and OCR, directly benefit robotic manipulation by strengthening the model's spatial understanding, object recognition, and scene generalization. Category (6) is specifically curated to bridge VL understanding and action generation: ECoT data teaches the model to perceive embodiment-specific scene states, track task progress, and reason about next actions in language. This encourages the VLM backbone to build richer embodied representations that are more directly useful for downstream continuous action generation~\citep{zawalski2024robotic, Chen2025TrainingSF}; egocentric video data exposes the model to fine-grained human manipulation patterns and object state transitions from a first-person perspective, grounding its understanding of how physical interactions unfold, including how objects deform, slide, or topple under contact and how hand-object configurations evolve during grasping, placing, and tool use. This knowledge transfers to robotic manipulation despite the embodiment gap. In addition, 2D trajectory prediction data directly connects visual observations to spatial motion reasoning in image coordinates, and together these data sources establish a shared representational foundation that facilitates knowledge transfer to low-level action prediction through the action expert.






We describe the synthesis procedure for each type of embodied-centric VL data below.

\begin{table}[tb]
\centering
\caption{\textbf{Taxonomy of atomic action types.}}
\vspace{-5pt}
\label{tab:atom_actions}
\small
\begin{tabular}{llL{9.5cm}}
\toprule
\textbf{Category} & \textbf{Action Type} & \textbf{Example} \\
\midrule
\multirow{3}{*}{Movement}
  & Reach (and grasp)         & ``Reach toward the red cup on the left side of the table and grasp it.'' \\
  & Move (and release) & ``Move the held cup onto the wooden tray and release it.'' \\
\midrule
\multirow{9}{*}{Manipulation}
  & Flip          & ``Flip the golden pancake 180 degrees in the frying pan.'' \\
  & Rotate        & ``Rotate the black knob on the oven door clockwise.'' \\
  & Toggle        & ``Toggle the red power switch to on.'' \\
  & Open          & ``Open the wooden drawer on the left side of the cabinet.'' \\
  & Close         & ``Close the lid of the black laptop.'' \\
  & Push          & ``Push the yellow block forward along the table.'' \\
  & Pull          & ``Pull the silver drawer handle away from the cabinet.'' \\
  & Insert        & ``Insert the red peg into the circular hole in the board.'' \\
  & Press         & ``Press the sponge against the table surface.'' \\
  & Click         & ``Click the red button on the control panel.'' \\
  & Strike        & ``Strike the silver nail with the wooden hammer.'' \\
\midrule
\multirow{3}{*}{Special}
  & Handover      & ``Move the held cable toward the right arm and release it.'' \\
  & Return to home & ``Return the arm to its home position.'' \\
  & Other         & ``Pour the water from the red cup into the glass.'' \\
\bottomrule
\end{tabular}
\end{table}

\paragraph{Embodied Chain-of-Thought (ECoT) Data.}
Inspired by prior work~\citep{zawalski2024robotic, feng2026@procvlm, li2026@robointer}, we construct ECoT supervision that trains the VLM to jointly perform three forms of embodied reasoning: describing the current scene, assessing task progress, and predicting the next atomic manipulation action. For a sampled timestamp $t$ in a manipulation trajectory with task instruction, we synthesize one ECoT annotation using both the current multi-view observation and additional trajectory context available only at annotation time.

Specifically, at timestamp $t$, we first extract synchronized images from all available camera views (e.g., front, wrist, and side views). We then construct three forms of additional context. First, we build a \textit{memory summary} by uniformly subsampling frames from the beginning of the episode up to $t$ and prompting a strong VLM to summarize completed actions and visible state changes. This memory mainly supports task progress assessment during annotation, since many intermediate goals or prior manipulations are no longer directly observable in the current frame. Second, we construct a \textit{future action preview} from 6 frames sampled at 1-second intervals starting from $t$ and ask a strong VLM to summarize the robot's immediate future behavior in this short clip. This future preview provides direct annotation-time evidence for predicting the next atomic action. Third, we compute a coarse \textit{temporal progress} estimate from the relative position of $t$ within the full trajectory, which serves as a weak auxiliary cue for judging whether the task is likely close to completion.

Given the multi-view images at $t$, the task instruction, and the auxiliary annotation-time context described above, we prompt a strong VLM\footnote{In practice, we use Qwen3.6-Plus with thinking mode throughout the ECoT data synthesis process.} to generate a structured three-part ECoT response: (1) a Scene Description summarizing observable objects, spatial relations, robot arm positions, and gripper states; (2) a Task Progress Assessment evaluating completed subgoals and ending with an explicit completion judgment (\textit{Task complete.} or \textit{Task not yet complete.}); and (3) a Next Action predicting a single atomic manipulation step from Table~\ref{tab:atom_actions}. Although the annotating VLM has access to privileged trajectory context, the prompt requires the generated ECoT text to be expressed using only evidence from the current observation and task instruction. During training, each ECoT data sample is converted into a standard VL input-output pair: the model receives only the multi-view images and task instruction, and is trained to generate the full three-part ECoT text. The privileged signals are used only during data synthesis to improve annotation quality and are excluded from training inputs.

\paragraph{Egocentric Video Understanding Data.}
We construct egocentric video understanding data from the human manipulation videos. For each episode, we split the main-camera video into non-overlapping clips of random duration (1.5--3 seconds) and extract 4 uniformly spaced frames per clip (at 0\%, 33\%, 67\%, and 100\% of the clip duration). A strong VLM is prompted to describe the fine-grained manipulation actions observed across the 4-frame sequence, including hand and arm movement directions, hand-object interactions with spatial relationships, and any object state changes. Clips with very little visual change, as determined by the VLM, are filtered out to avoid training on uninformative static segments. The resulting annotations teach the model to perceive and describe the dynamics of physical manipulation from an egocentric perspective, complementing the ECoT data, which operates on robot observations.

\paragraph{2D Trajectory Prediction Data.}
To further facilitate the VLM in learning motion planning from both robotic and human manipulation demonstrations, while alleviating ambiguity in depth perception, we project the trajectories of the robot end-effector (EEF) and the human hand in egocentric data onto the image using the estimated camera parameters. In addition, we filter out samples with little motion using a bounding-box-based criterion, thereby removing uninformative data points in which the EEF or human hand barely moves.

\section{Qwen-RobotManip: The Generalizable Vision-Language-Action Model Design}
\label{sec:model}

\subsection{Main Architecture}

\ours follows a decoupled architecture consisting of a vision-language backbone for multimodal perception and semantic reasoning, and a flow-matching action expert for continuous action generation. This decoupling allows the action expert to specialize in high-frequency, fine-grained motor control while the backbone retains and extends its pretrained perceptual and reasoning capabilities through joint end-to-end training.

\paragraph{Vision-language backbone.}
We adopt Qwen3.5-4B~\citep{qwen35blog} as the vision-language backbone.
Qwen3.5 is a natively multimodal model trained with early vision-language fusion: visual tokens from a Vision Transformer with dynamic-resolution spatial merging are interleaved directly into the text token stream and processed uniformly across images and language instructions within a single transformer.
Given one or more camera views together with a natural language task instruction, the backbone encodes them jointly into contextual representations (\eg, last-layer hidden states $D_{\mathrm{vlm}}{=}2560$) that capture both fine-grained visual features and task-level semantics, which are then consumed by the action expert via cross-attention.

\paragraph{Action expert.}
We attach a Diffusion Transformer (DiT)~\citep{peebles2023scalable} as a flow-matching action expert~\citep{chi2023diffusion,black2024pi0,adaptdiffuser} for learning precise continuous actions from both robot trajectory data and egocentric human demonstrations.
The expert consists of $N{=}10$ transformer blocks with hidden dimension $D_{\mathrm{act}}{=}768$ and 12 attention heads.
Each block performs self-attention over the concatenated state-and-action token sequence, followed by cross-attention to VLM hidden states and a SwiGLU feed-forward network.
Cross-attention layers alternate between attending to \emph{visual} tokens (even-indexed blocks) and \emph{language} tokens (odd-indexed blocks), both extracted from the final layer of the VLM, letting the expert separately ground action predictions in spatial observations and linguistic instructions at each processing stage.
The robot's proprioceptive state is encoded by a two-layer MLP and prepended to the noisy action token sequence before entering the DiT blocks. The expert is further conditioned on denoising timestep embeddings and additional learned camera embeddings detailed in \cref{subsec:cam_delta_action}.


The expert is trained with a flow-matching objective~\citep{lipman2023flow,esser2024scaling}. Given a ground-truth action chunk $\mathbf{a}$, a timestep $t \sim \mathrm{Beta}(1, 1.5)$ is sampled and an interpolant $\mathbf{x}_t = (1{-}t)\,\boldsymbol{\epsilon} + t\,\mathbf{a}$ is constructed from Gaussian noise $\boldsymbol{\epsilon}\sim\mathcal{N}(\mathbf{0},\mathbf{I})$. The model is then trained to minimize mean squared error on the predicted velocity field $\mathbf{x}_1 - \mathbf{x}_0$. At inference, action sequences are produced via 4 Euler integration steps, enabling low-latency real-time control. 

\begin{figure}[t]
    \centering
    \includegraphics[width=0.99\linewidth]{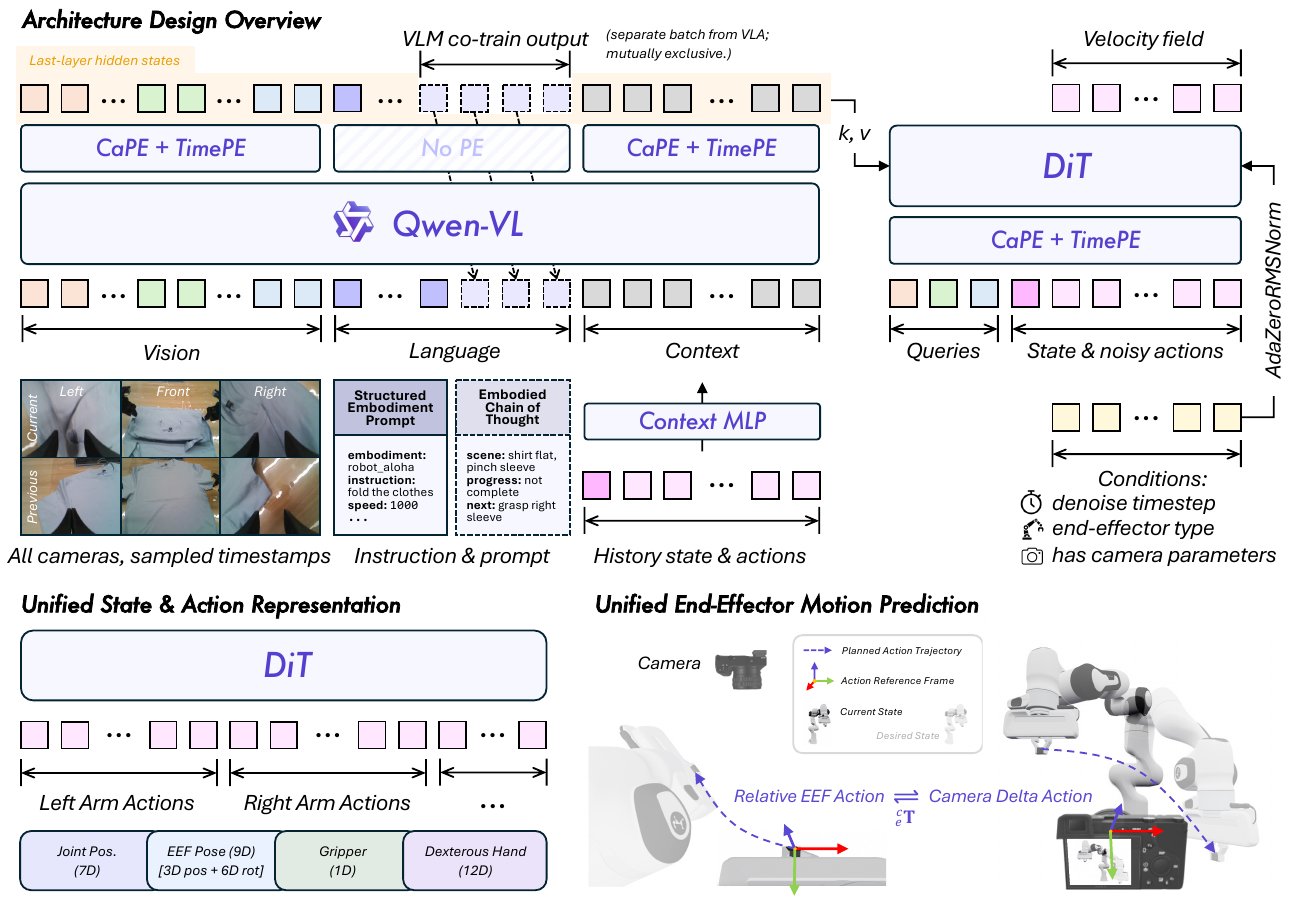}
    \caption{\textbf{Overview of \ours.} The model couples a Qwen-VL backbone with a flow-matching Diffusion Transformer (DiT) action head. The backbone jointly encodes multi-view visual tokens, structured embodiment prompts, and historical context tokens, with last-layer hidden states injected into the DiT via alternating cross-attention. States and actions share a unified 80-dimensional canonical representation, with end-effector actions expressed as camera-frame delta poses to align the action space with visual observations across embodiments, conditioned on camera and end-effector type embeddings for embodiment-aware denoising. 
    VLM co-training and VLA training use separate batches, where each batch contains either (Vision, QA) or (Vision, Language, Context, Action) data.}
    \label{fig:method}
\end{figure}

\subsection{Cross-Embodiment State and Action Representation}
\label{sec:unified_state_action}

Heterogeneous proprioceptive states and action spaces across embodiments make scalable training on mixed multi-embodiment datasets a key challenge.
We address this by introducing an \textbf{80-dimensional canonical vector representation} for states and actions.  
The representation is structured as two 29-dimensional per-arm blocks followed by 22 reserved dimensions.
Each per-arm block is organized into the following semantic groups:
\begin{itemize}[leftmargin=*,topsep=2pt,itemsep=1pt]
    \item \textbf{Joint positions} (7 dims): joint positions for the robot arm;
    \item \textbf{End-effector pose} (9 dims): Cartesian position (3) and orientation in a 6D continuous rotation representation~\citep{zhou2019continuity} (6);
    \item \textbf{Gripper state} (1 dim): joint position for the parallel gripper;
    \item \textbf{Dexterous hand joints} (12 dims): active hand joint positions for embodiments equipped with multi-fingered dexterous hands.
\end{itemize}
The trailing 22 reserved dimensions are shared across both arms and are available for additional degrees of freedom such as mobile-base velocity.
For the \emph{state} vector, all values are expressed in absolute coordinates.
For the \emph{action} vector, joint actions are expressed as absolute values and end-effector actions are expressed as relative deltas from the current state. In particular, end-effector orientation deltas are parameterized as 3D rotation vectors rather than the 6D representations used for states. \cref{subsec:cam_delta_action} further details the camera-frame delta representation for end-effector actions.

Different robot embodiments populate different subsets of this canonical template.
For example, a 7-DOF single-arm gripper (\eg, Franka Panda) fills the joint, end-effector, and gripper fields of one arm, leaving the remaining dimensions as zero.
A dual-arm system (\eg, ALOHA) fills both per-arm blocks. A robot with dexterous hands additionally populates the hand-joint dimensions. Zero-padded dimensions are excluded from the training loss via a per-dimension binary mask, ensuring that gradients flow only through semantically populated entries and preventing spurious supervision on structurally absent degrees of freedom.

\subsection{Unified End-Effector Motion Prediction}\label{subsec:cam_delta_action}

The canonical state-action representation of \cref{sec:unified_state_action} unifies the structural layout of states and actions across embodiments, but does not yet address a subtler source of fragmentation: end-effector poses recorded in different coordinate frames across datasets~\citep{lyu2026lda}. When the same motion is expressed relative to different base frames or camera frames depending on the data source, the model must learn to reconcile these geometric inconsistencies rather than focusing on the underlying manipulation skill. We address this by grounding all end-effector actions in a shared camera-frame delta pose representation and injecting camera geometry directly into the action expert via positional encodings. Together with embodiment-aware conditioning, this ensures that actions which appear visually similar are numerically proximate across embodiments, enabling the model to extract cross-embodiment synergies from heterogeneous data.

Concretely, we extract $N_{\mathrm{ee}} \in \{1, 2\}$ independent \textbf{40-dimensional per-end-effector tokens} from the 80-dimensional state-action vector. The 29 active dimensions of each arm are packed into a 40-dimensional slot with 11 reserved for future extension, and the DiT processes these tokens jointly via self-attention.


\paragraph{Camera-frame delta pose action representation.}
Rather than representing end-effector motion as an absolute pose in the robot base frame, a relative pose in the end-effector local frame, or a world-frame delta, we adopt a \emph{camera-frame delta pose} representation~\citep{chen2025toward}. 
Its key property is that actions appearing visually similar in the image are also numerically proximate in the action space, directly aligning the action representation with the visual observation space and facilitating cross-embodiment transfer.
This requires calibrated camera intrinsics and extrinsics at both training and inference.

Formally, let $c$ denote the reference camera frame, $e$ the current end-effector frame, and $e^*$ the desired end-effector frame at a future step.
The pose component of the predicted action is:
\begin{equation}
\label{eqn:cam_delta_rot_inv}
    \mathbf{a}_p = \begin{bmatrix}
        \ ^c _e \mathbf{R} \ ^e_{e^*} \mathbf{R} \ ^e _c \mathbf{R} & \ ^c_e \mathbf{R} \ ^e \mathbf{t}_{e^*} \\
        \mathbf{0} & 1
    \end{bmatrix}
\end{equation}
The rotational block ${}^c_e\mathbf{R}\ {}^e_{e^*}\mathbf{R}\ {}^e_c\mathbf{R}$ expresses the relative end-effector rotation ${}^e_{e^*}\mathbf{R}$ in the camera frame by conjugating with the camera-to-end-effector extrinsics, while the translational block ${}^c_e\mathbf{R}\ {}^e\mathbf{t}_{e^*}$ projects the desired end-effector displacement into camera coordinates. The full expression is thus geometrically equivalent to projecting the relative end-effector action into the camera frame via the extrinsics.
A more compact alternative is~\citep{zhang2026grounding}:
\begin{equation}
\label{eqn: cam_delta_full_inv}
    \mathbf{a}_p = \ ^c _{e^*} \mathbf{T} \ ^e _c \mathbf{T}
\end{equation}
While \cref{eqn: cam_delta_full_inv} eliminates end-effector definition inconsistencies entirely, its translational component is coupled with the relative end-effector rotation ${}^e_{e^*}\mathbf{R}$ and the camera-to-end-effector offset ${}^e\mathbf{t}_c$, making it more susceptible to long-tail distributions and more sensitive to calibration errors.
We therefore adopt \cref{eqn:cam_delta_rot_inv} in implementation.

\paragraph{Camera-aware positional encoding.}
To enable the action expert to reason about camera geometry, we inject camera parameters into the DiT's cross-attention layers via Camera Positional Encoding (CaPE)~\citep{kong2024eschernet}.
Camera pose is encoded via CaPE, occupying 32 of each 64-dimensional attention head's dimensions, with the remaining 32 used by RoPE~\citep{heo2024rotary} for temporal indexing.
Each image token's positional encoding is derived from the extrinsics of its corresponding camera, while each state/action token uses the extrinsics of its selected reference camera.
Because CaPE is a rotational positional encoding, the global world-frame origin cancels algebraically in the dot-product attention, leaving only the relative pose between each visual token and the querying state/action tokens.

Following the practice of GTA~\citep{miyato2024gta} and PRoPE~\citep{li2025cameras}, we apply CaPE not only to keys and queries but also to values and attention outputs, strengthening the geometric consistency of the cross-attention.
Camera intrinsics are incorporated by projecting the normalized image-plane coordinates of each visual patch through a learned linear layer and adding them to the corresponding image token, providing per-token field-of-view awareness.

\paragraph{End-effector-aware conditioning.}

Beyond the denoising timestep, the DiT is further conditioned on two additional signals, both applied via additive embeddings through adaptive layer normalization~\citep{peebles2023scalable}.
\begin{enumerate}[leftmargin=*,topsep=2pt,itemsep=1pt]
    \item \emph{End-effector type embedding}: a learned codebook entry per end-effector category (single-arm, dual-arm left, dual-arm right, egocentric head, or mobile base) associated with each state/action token, allowing the model to apply embodiment-specific action priors.
    \item \emph{Auxiliary flag embedding}: a binary embedding indicating whether calibrated camera parameters are available for the current sample, switching the predicted pose action space between camera-frame delta mode and robot-base relative mode.
\end{enumerate}

\paragraph{Multi-view reference camera selection.}
In multi-view settings, the end-effector action is expressed relative to a chosen reference camera frame.
During training, for single-arm datasets we randomly select any available external or wrist-mounted view as the reference. For dual-arm datasets, we randomly apply one of two strategies:
(1) both arms share a head-mounted camera or any available third-person view as the common reference frame;
(2) the left arm uses the left wrist camera and the right arm uses the right wrist camera as their respective reference frames.


Within the DiT's cross-attention, each image token uses the pose of its corresponding camera for CaPE, while each state/action token uses the pose of its selected reference camera for CaPE, guiding the DiT to denoise the camera delta action expressed in that reference frame.
Because CaPE is a rotational encoding, inter-view relative poses are encoded implicitly, with the world-frame superscript canceling algebraically.

\subsection{Embodiment Prompt}

We adopt a structured prompt to condition the policy on both task semantics and execution context. 
Each prompt consists of the following fields:
\begin{itemize}[leftmargin=*,topsep=2pt,itemsep=1pt]
    \item \textbf{Embodiment}: the robot platform (\eg, \texttt{robot\_aloha}), enabling the model to account for morphological and control differences across embodiments.
    \item \textbf{Instruction}: the high-level task description, defining the overall objective of the episode.
    \item \textbf{Speed}: the episode length in timesteps, discretized into bins of 500 steps.
    \item \textbf{FPS}: the temporal sampling rate of the input sequence.
    \item \textbf{Camera View Direction}: the camera's position relative to the robot arm, either \texttt{arm side} or \texttt{opposite side}.
    
\end{itemize}
    
    



\begin{center}
\begin{tcolorbox}[
  width=0.72\linewidth,
  colback=gray!6!white,
  colframe=gray!50!black,
  colbacktitle=gray!40!black,
  coltitle=white,
  title={\small\bfseries Structured Embodiment Prompt Example},
  arc=5pt,
  boxrule=0.8pt,
  titlerule=0pt,
  left=8pt, right=8pt, top=6pt, bottom=6pt,
]
{\color{gray!60!black}\textbf{embodiment:}}~{\color{gray!60!black}\texttt{robot\_aloha}}\\[2pt]
{\color{gray!60!black}\textbf{instruction:}}~{\color{gray!60!black}\texttt{Take the toy off the table and put it on the mat.}}\\[2pt]
{\color{gray!60!black}\textbf{speed:}}~{\color{gray!60!black}\texttt{1000}}\\[2pt]
{\color{gray!60!black}\textbf{fps:}}~{\color{gray!60!black}\texttt{30}} \\[2pt]
{\color{gray!60!black}\textbf{camera view direction:}}~{\color{gray!60!black}\texttt{arm side}}
\end{tcolorbox}
\end{center}

Together, these fields allow the model to capture not only what task should be performed, but also which robot is acting and how the behavior is temporally structured, 
reducing ambiguity in policy learning, improving adaptability across embodiments, and increasing robustness to variations in execution speed and frame rate. 
To further improve robustness to incomplete inputs, we randomly drop the embodiment, speed, and fps fields with probability 15\% during training, encouraging the model to generalize when prompt information is partially unavailable at test time.




\subsection{In-Context Policy Adaptation}
\label{sec:in_context}

Despite the strong generalization enabled by cross-embodiment pretraining, deploying a VLA policy to a new robot or environment often requires rapid behavioral adaptation without parameter updates. Inspired by in-context learning in large language models, we equip \ours{} with an in-context policy adaptation mechanism that conditions current action prediction on a structured window of recent execution history (observation-action pairs) from the same episode, enabling the policy to adapt its behavior at deployment time without any parameter update or task-specific fine-tuning.

\paragraph{Execution context representation.}
A key design question is what information constitutes a useful policy context. We draw a direct analogy from the model's own inference procedure. At each decision step, \ours{} observes the current visual observations and proprioceptive state and predicts a complete action chunk of $K$ steps. We therefore define one context chunk as exactly this triplet $(\mathbf{o}_h, \mathbf{s}_h, \mathbf{a}_h)$, consisting of the visual observation, proprioceptive state, and the $K$-step action sequence executed during chunk $h$. This records what the robot saw, was in, and did. A context of $H$ such chunks thus provides the policy with a structured window of recent behavior it can directly reason about.

The two modalities within each context chunk are processed through complementary pathways, owing to their fundamentally different representational structures. Historical frames $\mathbf{o}_h$ are prepended to the current frame and processed jointly by the VLM visual encoder within a single forward pass, with an image-count annotation appended to the language instruction to help the VLM attribute each visual token to its correct temporal position. Proprioceptive states and action chunks, which cannot be processed by the visual pathway, are projected into the VLM hidden space by two lightweight MLP encoders. The state encoder $\mathrm{MLP}_s$ and action encoder $\mathrm{MLP}_a$ produce per-chunk token representations with learned temporal position embeddings $\mathbf{e}_h^{\mathrm{temp}}$ to distinguish chunks and slot embeddings $\mathbf{e}_{0:K'}^{\mathrm{slot}}$ to distinguish action tokens within each chunk:
\begin{align}
    \mathbf{t}_h^s
        &= \mathrm{MLP}_s(\mathbf{s}_h)
           + \mathbf{e}_h^{\mathrm{temp}}
        \;\in\; \mathbb{R}^{D_\mathrm{vlm}},
    \label{eq:state_tok}\\
    \bigl[\mathbf{t}_h^{a,0},\ldots,\mathbf{t}_h^{a,K'-1}\bigr]
        &= \mathrm{reshape}\!\left(
             \mathrm{MLP}_a\!\left(\mathrm{flatten}(\mathbf{a}_h)\right)
           \right)
           + \mathbf{e}_h^{\mathrm{temp}}
           + \mathbf{e}_{0:K'}^{\mathrm{slot}}
        \;\in\; \mathbb{R}^{K' \times D_\mathrm{vlm}}.
    \label{eq:action_tok}
\end{align}
All $H$ chunks are serialized chronologically into a single context token sequence:
\begin{equation}
    \mathbf{C} = \bigl[
        \underbrace{\mathbf{t}_0^s,\;\mathbf{t}_0^{a,0{:}K'}}_{\text{chunk }0},\;
        \underbrace{\mathbf{t}_1^s,\;\mathbf{t}_1^{a,0{:}K'}}_{\text{chunk }1},\;
        \ldots,\;
        \underbrace{\mathbf{t}_{H-1}^s,\;\mathbf{t}_{H-1}^{a,0{:}K'}}_{\text{chunk }H-1}
    \bigr]
    \;\in\; \mathbb{R}^{H(1+K') \times D_\mathrm{vlm}}.
    \label{eq:ctx_seq}
\end{equation}
The current state $\mathbf{s}_t$ is not encoded here and continues to flow through the action head's dedicated state encoder unchanged, preserving full backward compatibility with the base \ours{} model.

\paragraph{History integration.}
We study two strategies for injecting context token sequence into the policy. In the \emph{unified} mode, context tokens $\mathbf{C}$ are appended to the end of the VLM input sequence and processed jointly with visual and language tokens under causal self-attention, allowing the VLM to reason over history, task description, and visual observations together. The resulting history-fused last-layer hidden states are passed to the DiT action head via cross-attention. In the \emph{dual} mode, the state-action context is injected directly into the DiT action head rather than the VLM, keeping the VLM context length unchanged at the cost of shallower history integration. Unified injection allows the VLM's full self-attention to jointly reason over behavioral history, task description, and visual observations, enabling richer cross-modal context integration than is possible when history is confined to the action head alone. We therefore adopt unified injection as the default configuration.

\paragraph{Stochastic context sampling.}
A naive strategy of always providing the $H$ most recent chunks leads to a degenerate shortcut. Because the last context chunk is temporally closest to the current step, the model can achieve low training loss by simply copying the most recent action chunk rather than genuinely reasoning about the episode's behavioral dynamics. This collapses the context mechanism into a trivial action-copy heuristic, which breaks down whenever the immediate history is ambiguous, atypical, or does not reflect the robot's broader execution style. What we want the model to learn is the behavioral profile of the current episode~\citep{huang2025dadu}, its velocity patterns, grasping strategies, and interaction signatures, not a shortcut based on temporal proximity. 
To prevent this, we introduce \textit{stochastic context sampling} during training. Rather than always supplying the $H$ chunks immediately preceding the current step, the context window is drawn from a random position within the episode. The sampled chunks may therefore be temporally distant from the current step, forcing the model to reason about the robot's behavioral profile across the full episode rather than exploiting recency as a shortcut. 

This randomization serves as a form of curriculum diversification. The model must learn to extract consistent behavioral style from \emph{any} subset of the episode history, making it robust to missing, partial, or temporally displaced context at inference time. 
At deployment, we supply a rolling window of the most recent $H$ chunks as context, allowing the model to leverage the full available history.
Empirically, stochastic context sampling proves critical for preventing this collapse. Without it, the policy achieves low training loss but poor task success, a clear sign that the model has learned to copy recent actions rather than reason about execution context. With it, the model genuinely exhibits in-context adaptation, adjusting its behavior based on the broader behavioral dynamics of the current episode.

\section{Training}
\label{sec:training}
\subsection{Pre-training Recipe}

\subsubsection{Dual-Stream Co-Training}
We train \ours on two complementary data streams simultaneously. The VLA stream is built from the full multi-source manipulation corpus described in \cref{sec:data}, comprising real-robot demonstrations, egocentric human-hand manipulation videos, and human-to-robot synthesized trajectories. 

The VLM stream consists of large-scale vision-language supervision data (\cref{sec:vlm_data}), co-trained alongside the VLA stream to prevent the pre-trained perceptual and language capabilities from degrading under action prediction optimization, which would directly weaken the model's ability to interpret novel instructions and generalize to unseen visual contexts~\citep{driess2025knowledge}. In practice, we adopt a 9:1 ratio of robot data to VL data.




\subsubsection{Training Objectives}

\textbf{Flow matching loss.}
For each VLA sample, we are given a ground-truth action chunk $\mathbf{a} \in \mathbb{R}^{T \times D}$. Following the flow-matching formulation, we construct a noisy interpolant
\[
\mathbf{x}_t = (1-t)\,\boldsymbol{\epsilon} + t\,\mathbf{a},
\]
where $\boldsymbol{\epsilon} \sim \mathcal{N}(\mathbf{0}, \mathbf{I})$ and $t \sim \mathrm{Beta}(1, 1.5)$. The action expert is trained to predict the corresponding velocity field $\mathbf{v} = \mathbf{a} - \boldsymbol{\epsilon}$, minimizing:
\begin{equation}
    \mathcal{L}_{\mathrm{FM}} = \mathbb{E}_{\mathbf{a},\,\boldsymbol{\epsilon},\,t}
        \left\| f_\theta\!\left(\mathbf{x}_t,\, t,\, \mathbf{s},\, \mathbf{o}\right) - \left(\mathbf{a} - \boldsymbol{\epsilon}\right) \right\|_2^2,
    \label{eq:fm_loss}
\end{equation}
where $f_\theta$ denotes the full model conditioned on the proprioceptive state $\mathbf{s}$ and the visual-language observation $\mathbf{o}$. Gradients from $\mathcal{L}_{\mathrm{FM}}$ are applied to both the VLM backbone and the action expert.

Because different robot embodiments populate different subsets of the 80-dimensional canonical action space (\cref{sec:unified_state_action}), 
we apply a composed binary mask $\mathbf{m} \in \{0,1\}^{T \times D}$ that restricts the objective to only the dimensions and time steps carrying valid supervision.
The mask is constructed from three complementary sources.
The \emph{per-dimension slot mask} identifies actively populated dimensions for the current embodiment. For instance, a single-arm gripper populates the joint, end-effector, and gripper fields of one arm while leaving the opposite arm and hand slots as zeros.
The \emph{step validity mask} excludes time steps flagged as anomalous by the data curation pipeline (\cref{sec:state_action_filtering}) or outside episode boundaries, with all subsequent steps also masked once any step is deemed invalid to preserve causal consistency.
For egocentric human data, a \emph{per-hand validity mask} zeros out an entire arm slot from the moment the corresponding hand exits the camera view, preventing the model from being trained on occluded hand trajectories.
The three masks are AND-combined, and the masked flow matching loss becomes a per-sample average over valid entries only:
\begin{equation}
    \mathcal{L}_{\mathrm{FM}} = \frac{1}{B}\sum_{i=1}^{B}
        \frac{\sum_{t,j}\, m_{i,t,j}\,\bigl(f_\theta(\mathbf{x}_{i,t},\, t_i,\, \mathbf{s}_i,\, \mathbf{o}_i)_{j} - v_{i,t,j}\bigr)^2}
             {\sum_{t,j}\, m_{i,t,j}},
    \label{eq:fm_loss_masked}
\end{equation}
where $B$ is the batch size and subscripts $t,j$ index the time step and dimension.
This formulation ensures that every sample in the batch contributes equally to the gradient regardless of how many dimensions are active, preventing embodiments with more populated slots from dominating optimization.

\textbf{VLM next-token prediction loss.}
For each VLM sample, the backbone is trained with the standard autoregressive next-token prediction objective:
\begin{equation}
    \mathcal{L}_{\mathrm{VLM}} = -\mathbb{E} \sum_{i} \log p_\phi\!\left(y_i \mid y_{<i},\, \mathbf{c}\right),
    \label{eq:vlm_loss}
\end{equation}
where $y_i$ is the target response token at position $i$, and $\mathbf{c}$ is the input context, which may be text-only or an interleaved sequence of images and text.
The overall objective is
\begin{equation}
    \mathcal{L} = \mathcal{L}_{\mathrm{FM}} + \lambda\,\mathcal{L}_{\mathrm{VLM}},
    \label{eq:total_loss}
\end{equation}
where $\lambda$ controls the relative weight of the two losses.
We set $\lambda = 0.1$ so that VLM supervision provides stabilizing regularization without overshadowing action learning. Separate learning rates are used for the backbone and the action expert to account for their different initialization scales.
To amortize the cost of the VLM forward pass, the action expert performs $K_{\mathrm{repeat}}{=}8$ repeated diffusion steps per training sample, drawing 8 independent noise samples and timesteps for the same action chunk, and substantially improving training efficiency without increasing data consumption.

\subsection{Post-Training Recipe}

\subsubsection{Domain-specific Supervised Fine-tuning}
\label{sec:domain_sft}
Once the foundation model has been pre-trained on the full heterogeneous corpus, we adapt it to specific deployment scenarios through supervised fine-tuning (SFT). 
Rather than training specialist policies for individual tasks, we adopt a \emph{generalist SFT} paradigm. 
For each benchmark or real-world deployment scenario, all available demonstration data is combined into a single training set, 
producing one unified fine-tuned model that can execute every task within the target domain. 

Compared with pre-training, SFT differs in several aspects.
First, the SFT procedure only optimizes the flow matching objective $\mathcal{L}_{\mathrm{FM}}$ of \cref{eq:fm_loss_masked} without the VLM next-token prediction loss.
We disable the multi-stage data curation filtering of \cref{sec:state_action_filtering} and train on the complete unfiltered data, preserving every valid demonstration for post-training.
We apply color jitter augmentation to the input images.
In addition, with the model initialized from a pre-trained checkpoint, SFT is conducted on fewer GPUs with fewer training steps than pre-training. 

\subsubsection{Co-Training in Post-Training}
Domain-specific SFT has become an important protocol for quantitatively evaluating the quality of a pre-trained VLA model. 
A strong pre-trained model is expected to adapt efficiently to a target domain after fine-tuning on the benchmark training set. However, this evaluation protocol can also expose a critical failure mode. After extensive benchmark-specific SFT, a VLA model may improve task performance by exploiting repeated visual and task patterns in the benchmark, while becoming less sensitive to the language instruction. In this case, the policy is no longer strongly conditioned on both vision and language. Instead, it behaves more like a vision-action pattern matcher. We refer to this phenomenon as VLA-to-VA degradation.

This degradation stems from three compounding factors. First, domain-specific SFT datasets are limited in diversity with concentrated visual layouts and instruction expressions, making shortcut correlations between scene patterns and actions easy to learn. Second, training and test splits share similar visual patterns, allowing high benchmark scores to be achieved through pattern memorization rather than genuine language grounding. As a result, such benchmarks overestimate instruction-following ability and fail to distinguish true language-conditioned control from benchmark-pattern overfitting. Third, models without strong compositional grounding ability tend to treat language as a weak context signal during SFT, with actions increasingly dominated by visual shortcuts learned from the benchmark data.

To mitigate this risk, we propose a mixed post-training strategy as an optional enhancement to the standard domain SFT described in \cref{sec:domain_sft}: rather than fine-tuning solely on the benchmark training set, the model is co-trained with a subset of pre-training data filtered by distributional proximity to the target domain. 
This provides broader adaptation signals while preserving the base model's robust execution capabilities, without introducing unrelated data that might dilute domain-specific learning. 
In our main experiments, we follow the standard domain-specific SFT protocol to ensure fair comparison with baselines; the mixed post-training strategy is validated as an additional enhancement in \cref{sec:mixposttrain}.

To directly evaluate whether the language-following capability is preserved, we construct a new benchmark, RoboTwin-IF (\textbf{I}nstruction \textbf{F}ollowing), based on RoboTwin 2.0~\citep{chen2025robotwin}. 
The benchmark examines whether the model performs the instructed action in the same or similar visual scenes with different instructions, rather than relying on visual pattern matching to select a default behavior. 
The detailed benchmark design is presented in \cref{sec:eval_protocol}.

\section{Deployment}
\label{sec:deployment}
In our deployment setup, inference is performed on a remote server with observations and actions transmitted between the robot and the server over a WiFi connection. To mitigate the latency introduced by cloud-based inference and network transmission, we adopt Real-Time Chunking (RTC)~\citep{black2026real}, which asynchronously generates the next action chunk while the robot executes the current one, effectively hiding the round-trip latency and enabling smooth real-time control.

\section{Experiments}
\label{sec:experiments}

\subsection{Are Standard Benchmarks Enough?}
\label{subsec:are_std_bench_enough}

We evaluate \ours on robotic manipulation benchmarks that span a diverse range of embodiments and task types, and compare against several recent VLA models~\citep{black2024pi0, bjorck2025grootn1, community2026starvla, pi05}. 
We begin with LIBERO~\citep{liu2023libero} and RoboTwin~\citep{mu2025robotwin}, two standard benchmarks that are widely used for VLA evaluation.
LIBERO~\citep{liu2023libero} comprises four single-arm tabletop manipulation suites across 130 combinations of tasks and scenes. 
RoboTwin~\citep{mu2025robotwin} presents 50 dual-arm manipulation tasks in easy and hard modes, requiring adaptation to varied backgrounds, objects, and spatial layouts.

\begin{figure}[!h]
    \centering
    \includegraphics[width=1.0\linewidth]{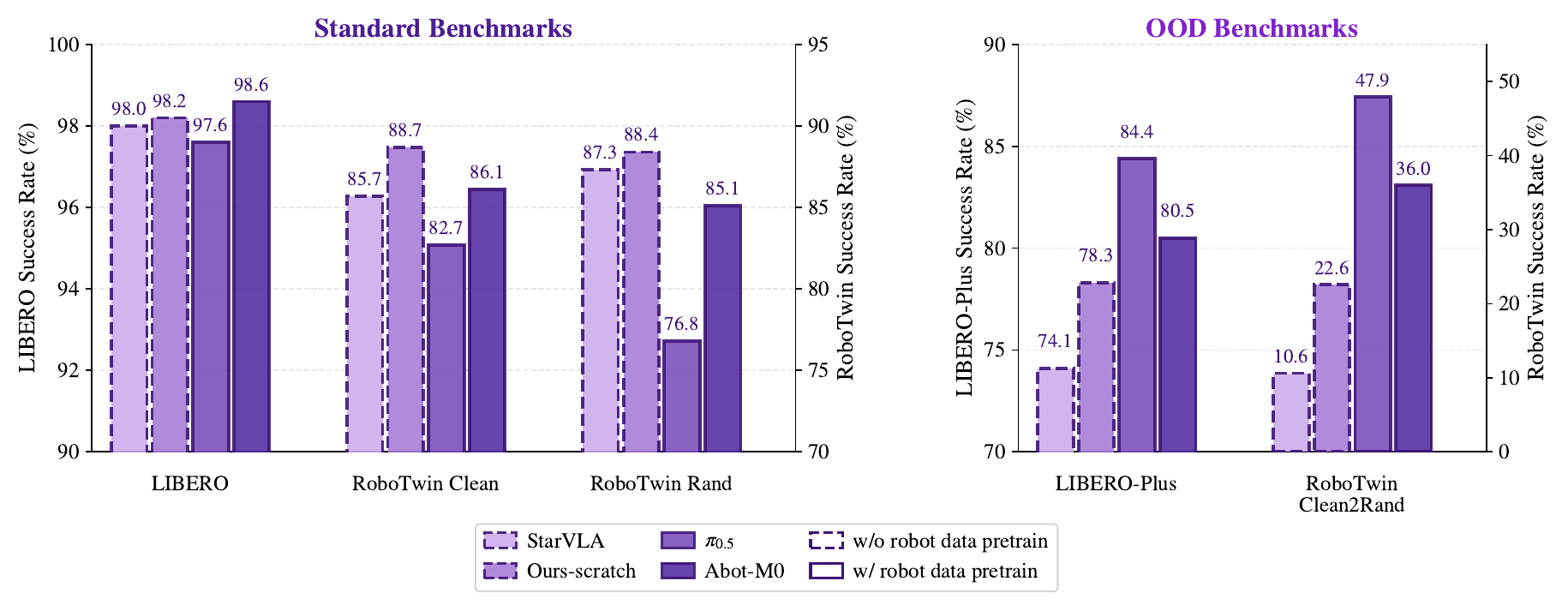}
    \caption{
    \textbf{Standard in-distribution benchmarks cannot reveal whether a model benefits from large-scale robot data pretraining.}
    Left: on in-distribution benchmarks (LIBERO, RoboTwin), models without large-scale robot pretraining (dashed borders) match or exceed pretrained ones. Right: on OOD benchmarks (LIBERO-Plus, RoboTwin-Clean2Rand), a clear separation emerges---pretraining provides genuine generalization that training-from-scratch cannot replicate. 
    }
    \label{fig:iid_diagnostic}
\end{figure}


Figure~\ref{fig:iid_diagnostic} (left) reports the results. A notable pattern is that models trained from scratch---StarVLA and Ours-scratch---attain competitive or even superior results compared to well-pretrained models such as $\pi_{0.5}$ and Abot-M0 on LIBERO and RoboTwin, despite lacking large-scale robotic pretraining. This is not a coincidence but a structural property of these benchmarks. Because training and evaluation data are drawn from the same environment and task distributions, high success rates can be achieved through in-distribution pattern matching alone. Models that lack genuine generalization can perform well simply by memorizing recurring visual and behavioral patterns, and the benchmark cannot distinguish this from real capability. Prior work has further shown that fine-tuning a pretrained VLA on a benchmark's own training split yields performance comparable to training from scratch~\citep{Yan_2025_ICCV}, confirming that the pretrained prior contributes negligible transferable value under in-domain evaluation.

Figure~\ref{fig:iid_diagnostic} (right) tells a different story. On OOD benchmarks---LIBERO-Plus and RoboTwin-Clean2Rand---where evaluation conditions diverge from training, a clear separation emerges: $\pi_{0.5}$ substantially outperforms StarVLA and Ours-scratch, with the gap widening as perturbation severity increases. StarVLA collapses from 85.7\% (RoboTwin Easy, IID) to 10.6\% (RoboTwin-Clean2Rand, OOD). This confirms that OOD evaluation is the correct north star for measuring foundation model quality: it reveals the transferable structure that pretraining provides and that in-domain metrics systematically fail to capture.

The way practitioners actually use a foundation model reinforces this conclusion. A researcher or engineer deploying a VLA model does not have access to the same distribution of tasks, objects, and environments present in any benchmark. They have a handful of demonstrations collected on their own hardware, in their own workspace, for their own task. What determines whether the foundation model helps them is not its in-domain benchmark rank but how much generalizable structure it has internalized, and how efficiently that structure transfers under minimal fine-tuning on unfamiliar data. The following section therefore adopts OOD evaluation as the primary measure and reports comprehensive benchmarking results for \ours across both in-distribution and out-of-distribution settings.

\subsection{Generalization Capabilities}

\begin{figure}[tb]
    \centering
    \includegraphics[width=1.0\linewidth]{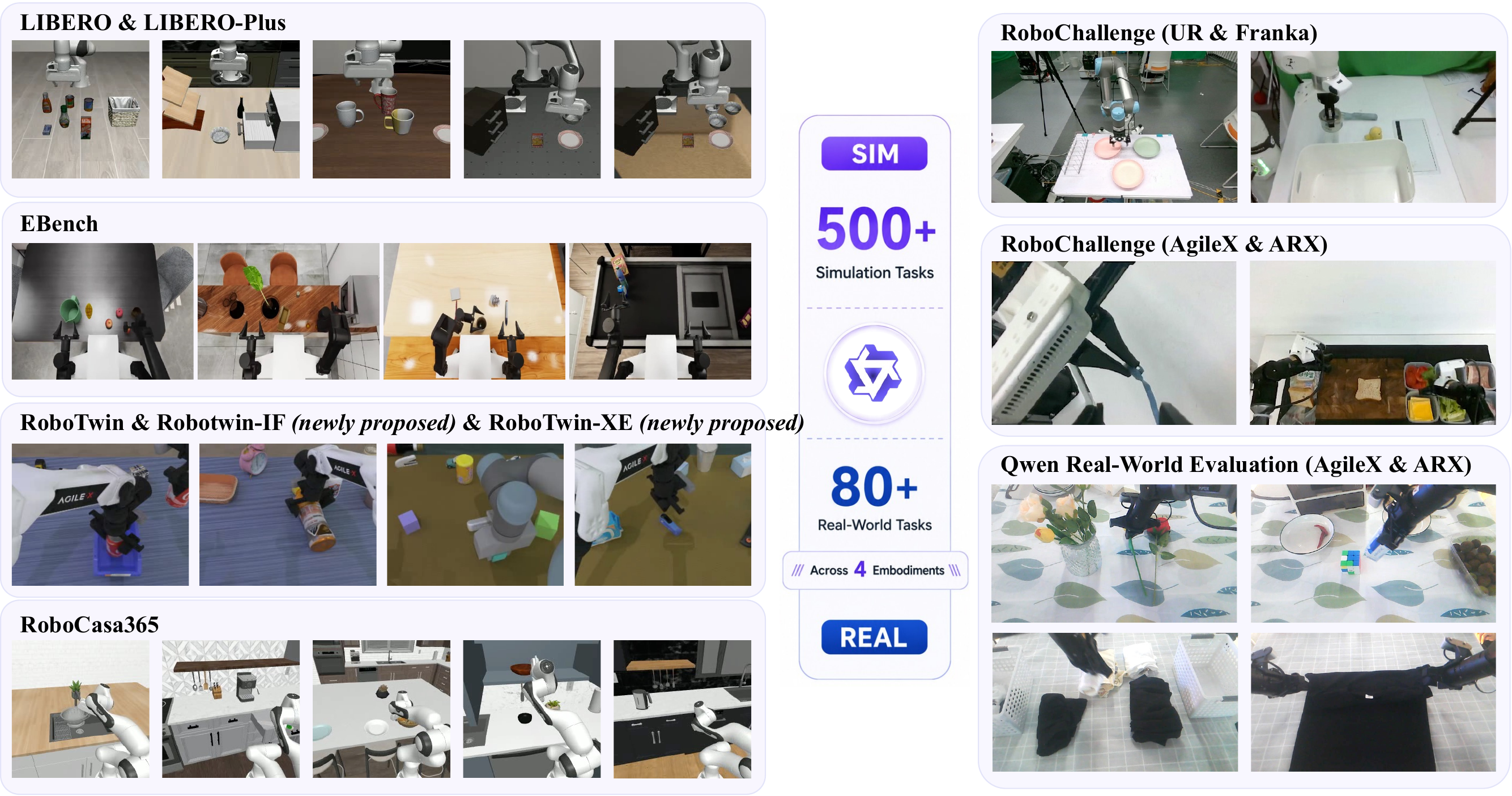}
    \caption{\textbf{Evaluation settings for \ours}, spanning 500+ simulation tasks across LIBERO, LIBERO-Plus, EBench, RoboTwin, RoboTwin-IF, RoboTwin-XE, and RoboCasa365, and 80+ real-world tasks across 4 embodiments including UR, Franka, AgileX (ALOHA), and ARX.}
    \label{fig:eval_setting_overall}
\end{figure}

The following sections evaluate \ours on a suite of out-of-distribution settings that directly measure the generalization capabilities a foundation model is expected to provide. We organize these evaluations around three axes (Figure~\ref{fig:eval_setting_overall}): task and scene generalization under controlled perturbations, instruction following with novel language and tasks, and zero-shot cross-embodiment transfer.

\subsubsection{Evaluation Protocol}
\label{sec:eval_protocol}

\paragraph{Task and Scene Generalization.}
We evaluate robustness of VLA models to visual and physical changes on four benchmarks.
\textbf{LIBERO-Plus}~\citep{fei2025liberoplus} introduces controlled perturbations along seven orthogonal dimensions on top of the original LIBERO benchmark---background textures, camera viewpoints, language instructions, lighting conditions, object layouts, robot initial states, and sensor noise---with models fine-tuned on the standard LIBERO training set and evaluated under each perturbation.
\textbf{RoboTwin-Clean2Rand} constructs an OOD evaluation protocol on top of RoboTwin~\citep{mu2025robotwin}: all models are fine-tuned exclusively on a Clean dataset with fixed white background, default lighting, no distractors, and a fixed table height, then evaluated under controlled randomizations along individual axes (background, lighting, clutter, table height) as well as a Hard setting that applies all randomizations simultaneously.
\textbf{RoboCasa365}~\citep{nasiriany2026robocasa365} provides a large-scale kitchen manipulation benchmark with three evaluation suites of increasing difficulty: Atomic (18 basic manipulation skills with diverse object and layout variations), Composite-Seen (multi-step long-horizon tasks seen during training), and Composite-Unseen (long-horizon tasks absent from the training set).
\textbf{EBench}~\citep{ebench2026} is an indoor mobile manipulation benchmark built on NVIDIA Isaac Sim, spanning 26 task types and 794 evaluation instances. The benchmark evaluates generalization along perturbation dimensions: background, instruction, object, and a mixed setting that combines all perturbations, reporting both success rate and a process score for each.

\paragraph{Instruction Following.}
Existing OOD benchmarks primarily probe robustness to visual and physical perturbations, while leaving generalization to unseen language instructions largely untested. We develop \textbf{RoboTwin-IF} (\textbf{I}nstruction \textbf{F}ollowing), a benchmark built on RoboTwin~\citep{mu2025robotwin} that systematically evaluates instruction-following capabilities across five task suites, each targeting a distinct dimension of language grounding:
\begin{itemize}[leftmargin=*,topsep=2pt,itemsep=1pt]
    \item \textbf{Pick-Diverse-Object}: Four objects are randomly sampled from a pool of 12 everyday items. The instruction names one object by color and noun. The robot must identify and lift the correct target among three distractors, testing \emph{target-object grounding}.
    \item \textbf{Place-Relative}: Two named objects and 1--3 distractors are on the table. The instruction specifies picking up object A and placing it in a spatial relation (``beside'' or ``on top of'') with respect to object B, testing \emph{spatial-relation understanding}.
    \item \textbf{Operate-Mic-Drawer}: A microphone and a cabinet with a functional drawer are present. The instruction specifies a multi-step bimanual sequence: open the drawer with one arm, then pick up the microphone with the other arm and place it inside. Some instructions further specify which arm performs which sub-task, testing \emph{multi-step sequencing and bimanual coordination}.
    \item \textbf{Operate-Stapler}: A stapler, a colored pad, and 1--2 distractors are on the table. The instruction specifies either pressing the stapler or moving it onto the colored pad. The pad is always present regardless of the verb, acting as a distractor in press episodes and as the placement target in move episodes, testing \emph{verb discrimination with shared scene elements}.
    \item \textbf{Operate-Tabletop}: A bell, a stapler, and 1--2 pickable objects are all present simultaneously. The instruction specifies one of three actions: ring the bell, press the stapler, or pick up a named object. Only one action is correct; the other interactive objects are distractors, testing \emph{three-way verb-and-target discrimination} in a multi-affordance scene.
\end{itemize}

\begin{figure}[tb]
    \centering
    \includegraphics[width=1.0\linewidth]{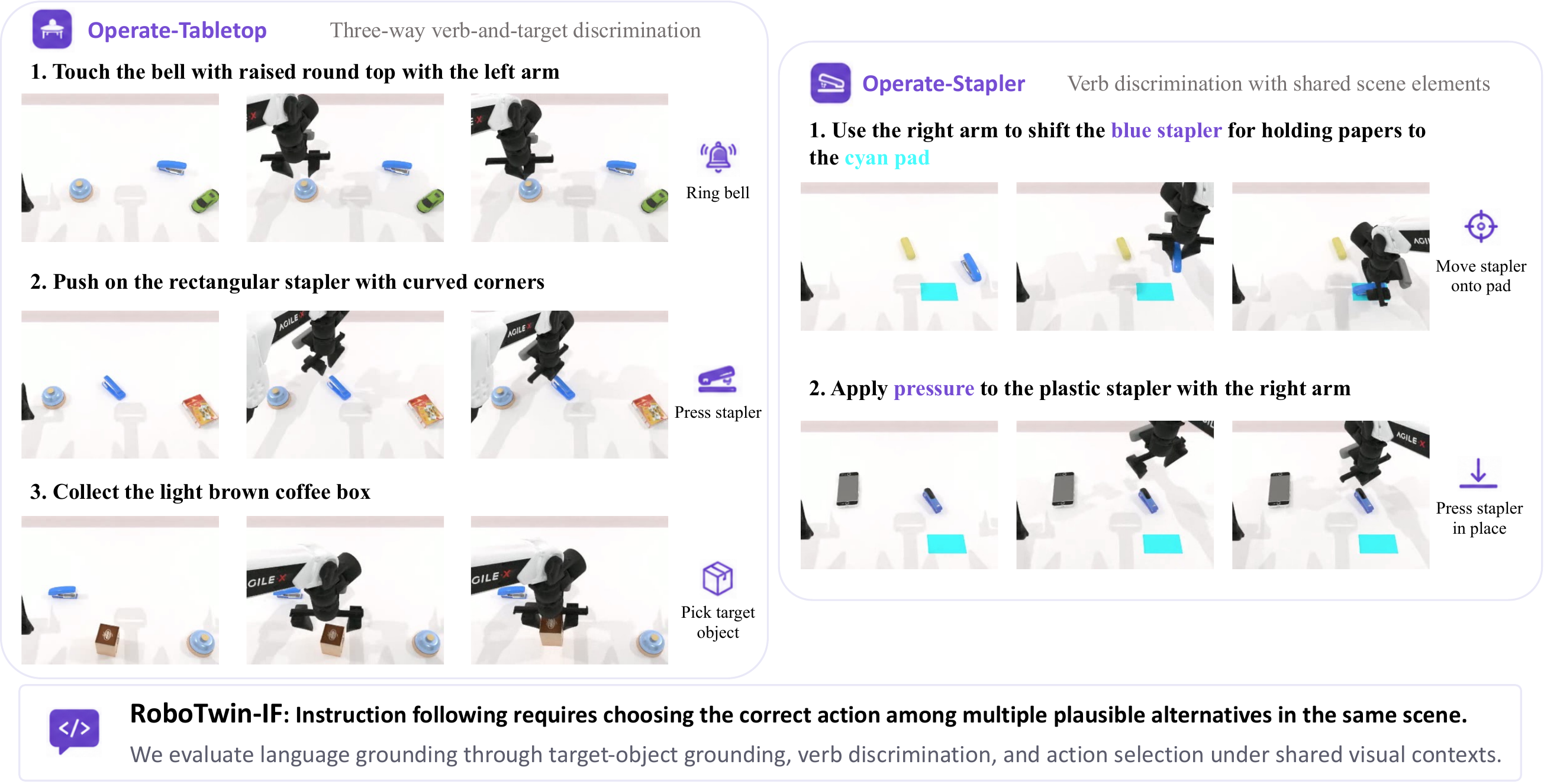}
    \caption{\textbf{Representative task suites from RoboTwin-IF.} Operate-Tabletop (left) requires three-way verb-and-target discrimination where a bell, stapler, and pickable objects are all present and only the instruction-specified action is correct. Operate-Stapler (right) requires verb discrimination under shared visual context, where the colored pad and stapler are always present regardless of the instructed verb. Both suites are evaluated on unseen instruction templates held out from training.}
    \label{fig:eval_robotwin_if}
\end{figure}

All models are fine-tuned exclusively on RoboTwin Clean. Each suite uses a two-tier language diversity system: ``seen'' instruction templates (used during training data collection) and ``unseen'' templates (held out for evaluation), combined with per-object description variants that further diversify noun phrases. At evaluation time, each episode is paired with a held-out unseen instruction template, ensuring zero overlap with the training distribution.

\paragraph{Cross-Embodiment Generalization.}

Camera-frame relative EEF actions express deltas in the camera coordinate frame rather than a robot-specific joint space, enabling a single policy to potentially control morphologically distinct robots without re-training. 
We develop \textbf{RoboTwin-XE}, a benchmark based on RoboTwin that evaluates zero-shot transfer to unseen robot embodiments. 
The model is fine-tuned on the RoboTwin Clean dataset and tested under the RoboTwin Hard setting, replacing the default AgileX platform with ARX-X5, UR5-WSG, and Franka Panda. 
Initial EEF poses are aligned via IK, head and wrist camera extrinsics are kept identical, and task scenes, object layouts, and perturbation seeds are shared across embodiments. 
The model must therefore generalize across two types of embodiment-specific differences: the visual appearance of the robot arm in camera observations, and kinematic differences (DOF count, joint arrangement, link lengths, and workspace geometry) that affect reachability and motion dynamics. 
The model is fine-tuned exclusively on AgileX demonstrations; no target-embodiment data is used.


\subsubsection{Summary of Results}

On standard in-distribution benchmarks, \ours achieves state-of-the-art or competitive performance (Table~\ref{tab:iid_results}) on LIBERO (99.2\%) and RoboTwin Easy/Hard (93.7\%/94.0\%). However, as argued in \cref{subsec:are_std_bench_enough}, these metrics do not reliably distinguish genuine generalization from in-distribution pattern matching. We therefore focus on OOD evaluation as the primary measure of robotic foundation model capabilities.

\begin{table}[!t]
\centering
\caption{\textbf{In-distribution benchmark results.} \ours matches or exceeds prior state-of-the-art across all standard benchmarks.}
\label{tab:iid_results}
\small
\tabcolsep 5pt
\begin{tabular}{lccc}
\toprule
& \textbf{LIBERO} & \textbf{RoboTwin-Easy} & \textbf{RoboTwin-Hard} \\
\midrule
$\pi_0$~\citep{black2024pi0} & 94.4 & 65.9 & 58.4 \\
$\pi_{0.5}$~\citep{pi05} & 97.6 & 82.7 & 76.8 \\
StarVLA~\citep{community2026starvla} & 98.0 & 85.7 & 87.3 \\
Abot-M0~\citep{yang2026abot} & 98.6 & 86.1 & 85.1 \\
Being-H0.7~\citep{beingh07} & \textbf{99.2} & 90.2 & 89.6 \\
\midrule
\ours-scratch & 98.2 & 88.7 & 88.4 \\
\ours & {99.1} & {93.4} & {92.5} \\
\ours-Context & \textbf{99.2} & \textbf{93.7} & \textbf{94.0} \\
\bottomrule
\end{tabular}
\end{table}

\begin{figure}[!t]
    \centering
    \includegraphics[width=0.9\linewidth]{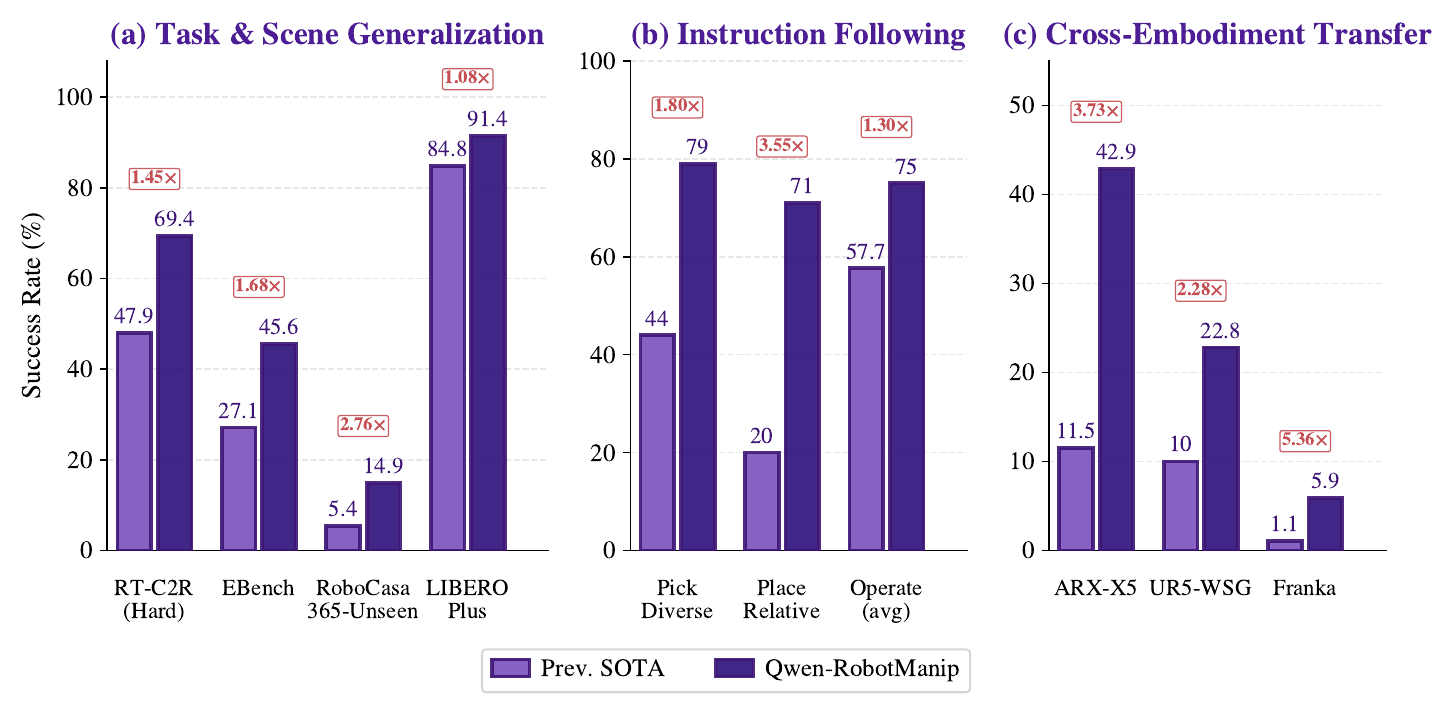}
    \caption{\textbf{OOD generalization summary.} \ours vs.\ previous state-of-the-art VLA model across three generalization axes. (a)~Task and scene generalization under controlled perturbations. (b)~Instruction following with held-out language templates. (c)~Zero-shot cross-embodiment transfer. \ours outperforms previous models on every OOD benchmark, with the gap widening on harder settings.}
    \label{fig:ood_summary}
\end{figure}

Figure \ref{fig:ood_summary} summarizes results across benchmarks evaluated under OOD settings, where \ours consistently outperforms prior state-of-the-art models by a substantial margin. \ours outperforms $\pi_{0.5}$ on all benchmarks evaluated, with the gap widening as the evaluation becomes more challenging.

On task and scene generalization, \ours surpasses $\pi_{0.5}$ by 7.0 points on LIBERO-Plus (91.4 vs.\ 84.4), by 21.5 points on the most demanding RoboTwin-C2R Hard setting (69.4 vs.\ 47.9), and by 18.5 points on EBench (45.6 vs.\ 27.1). On RoboCasa365, the advantage reaches 19.0 points (35.9 vs.\ 16.9).

On instruction following, \ours scores 72.2\% on RoboTwin-IF against 49.6\% for $\pi_{0.5}$---a large improvement of 22.6 points. This result is particularly significant because instruction following is the capability most prone to degradation during VLA training: fine-tuning a VLM on action prediction can erode the language conditioning pathway, causing the policy to collapse into a visually-triggered default behavior that ignores the instruction entirely (\cref{sec:training}). The fact that \ours maintains strong instruction following across all five RoboTwin-IF suites---including tasks requiring target-object grounding, spatial-relation understanding, and multi-way verb discrimination---indicates that the dual-stream co-training strategy and the diversity of the pretraining corpus together preserve genuine language-conditioned control.

The most revealing comparison is cross-embodiment transfer on RoboTwin-XE. When both models are trained exclusively on demonstrations collected on the AgileX ALOHA platform and evaluated zero-shot on unseen robots, \ours achieves 23.9\% using camera-frame EEF actions---3.2$\times$ the 7.5\% achieved by $\pi_{0.5}$. This gap validates our camera-frame alignment strategy for action representation: by expressing actions in the visual domain, physically similar motions become numerically proximate across morphologically distinct robots, enabling effective cross-embodiment transfer.


A complementary signal comes from the comparison with models trained from scratch. On LIBERO-Plus, \ours-scratch scores 78.3 versus \ours's 89.0 and \ours-Context's 91.4. The gap is even more pronounced on RoboTwin-C2R, where \ours-scratch collapses from 71.6 (Easy) to 22.6 (Hard), retaining about 30\% of its Easy performance, while \ours degrades far more gracefully from 73.2 to 62.6, retaining roughly 86\% (compared to 66\% for $\pi_{0.5}$). 
A consistent pattern emerges across benchmarks. The pretrained VLM backbone already provides robustness to some of the visual and linguistic variations such as background, lighting, and language perturbations. However, the capabilities that distinguish genuine generalization from in-distribution memorization, including spatial reasoning under novel viewpoints, robustness to unseen robot states, and attention to task-relevant objects in cluttered scenes, specifically require the large-scale cross-embodiment pretraining that \ours provides.


\subsubsection{Detailed Analysis}

\paragraph{LIBERO-Plus.}

\begin{table}[!t]
\centering
\caption{\textbf{Out-of-distribution robustness evaluation on LIBERO-Plus across seven perturbation dimensions.}}
\label{tab:libero_plus}
\small
\tabcolsep 3pt
\begin{tabular}{lcccccccc}
\toprule
& \textbf{Camera} & \textbf{Robot} & \textbf{Language} & \textbf{Light} & \textbf{Background} & \textbf{Noise} & \textbf{Layout} & \textbf{Total} \\
\midrule
$\pi_0$~\citep{black2024pi0} & 13.8 & 6.0 & 58.8 & 85.0 & 81.4 & 79.0 & 68.9 & 53.6 \\
$\pi_{0.5}$~\citep{pi05} & 78.4 & 73.6 & 80.8 & 96.2 & 94.1 & 89.0 & 84.5 & 84.4 \\
StarVLA~\citep{community2026starvla} & 52.5 & 49.8 & \textbf{88.5} & 95.7 & 95.7 & 73.0 & 76.9 & 74.1 \\
Abot-M0~\citep{yang2026abot} & 60.4 & 67.9 & 86.4 & 96.2 & 91.6 & 86.4 & 82.6 & 80.5 \\
Cosmos-Policy~\citep{kim2026cosmos} & 75.8 & 63.3 & 81.7 & 96.5 & 88.9 & 92.7 & 82.2 & 82.2 \\
Being-H0.7~\citep{beingh07} & 82.0 & 59.0 & 82.8 & 97.8 & 90.0 & 93.5 & \textbf{88.5} & 84.8 \\
\midrule
\ours-scratch & 70.4 & 44.9 & 88.1 & 95.8 & 95.5 & 84.4 & 79.1 & 78.3  \\
\ours & 87.2 & 75.5 & 85.6 & 96.6 & 97.7 & 97.7 & 87.3 & 89.0 \\
\ours-Context & \textbf{89.9} & \textbf{83.9} & {86.5} & \textbf{98.6} & \textbf{99.9} & \textbf{97.9} & {87.5} & \textbf{91.4}  \\
\bottomrule
\end{tabular}
\end{table}

Table~\ref{tab:libero_plus} reports per-dimension results. \ours achieves 89\% and \ours-Context achieves 91.4\% overall, surpassing all baselines. Beyond the aggregate score, the per-dimension breakdown reveals which capabilities benefit from large-scale robot data pretraining and which are already provided by the pretrained VLM backbone. Models without robot data pretraining (StarVLA, \ours-scratch) match most pretrained models on Language, Light, and Background perturbations, indicating that the VLM's visual and linguistic representations are inherently robust to these variations. In contrast, these same models suffer sharp degradation under Robot perturbation (49.8\% and 44.9\% vs.\ 75.5\% for \ours), where the policy must generalize across unseen initial robot states that lie outside the VLM's purview. 
\ours-Context further raises Robot perturbation robustness to 83.9\%, a +8.4 point gain over \ours, indicating that in-context history provides an implicit kinematic prior that helps the policy adapt to unfamiliar initial robot configurations within the episode.
Camera viewpoint perturbation exhibits a similar pattern (+34.7 and +16.8 over the scratch models), suggesting that robust spatial reasoning under novel camera poses requires the grounding that VLA pretraining on diverse robot setups provides, rather than the appearance-level robustness already captured by the VLM.

\begin{table}[!t]
\centering
\caption{\textbf{Out-of-distribution evaluation on RoboTwin-Clean2Rand.} Models are fine-tuned on the Clean dataset only and tested under various environmental randomizations. }
\label{tab:robotwin_e2h}
\small
\tabcolsep 3pt
\begin{tabular}{lcccccc}
\toprule
& \textbf{Easy} & \textbf{Background} & \textbf{Light} & \textbf{Clutter} & \textbf{Height} & \textbf{Hard} \\
\midrule
StarVLA~\citep{community2026starvla} & 58.1 & 27.1 & 50.9 & 24.2 & 48.4 & 10.6 \\
GR00T-N1.7~\citep{bjorck2025grootn1} & 43.6 & 40.4 & 41.9 & 27.1 & 39.0 & 20.7 \\
$\pi_{0.5}$~\citep{pi05} & {73.1} & 67.0 & 69.2 & 57.9 & 67.6 & 47.9 \\
Abot-M0~\citep{yang2026abot} & 70.7 & 56.5 & 68.8 & 46.0 & 56.3 & 36.0 \\
\midrule
\ours-scratch & 71.6 & 60.6 & {70.7} & 24.6 & 63.6 & 22.6 \\
\ours (joint) & 73.2 & 74.6 & 68.4 & 61.3 & 71.0 & 62.6 \\
\ours (eef) & 74.0 & 75.8 & 70.1 & 59.8 & 69.4 & 60.8 \\
\ours-Context (joint) & 84.7 & \textbf{82.4} & 84.2 & \textbf{75.4} & 79.5 & \textbf{69.4} \\
\ours-Context (eef) & \textbf{85.0} & \textbf{82.4} & \textbf{84.7} & 66.8 & \textbf{82.9} & 64.0 \\
\bottomrule
\end{tabular}
\end{table}

\paragraph{RoboTwin-Clean2Rand.}

Table~\ref{tab:robotwin_e2h} reports results across individual and compound perturbation axes. \ours achieves the highest success rate on the Hard setting under both joint-space control (62.6\%) and end-effector control (60.8\%), with both modes performing similarly across all perturbation dimensions. Among all methods, \ours exhibits the smallest degradation from Easy to Hard, retaining roughly 86\% of its Easy performance compared to 66\% for $\pi_{0.5}$ and about 30\% for models without robot data pretraining. The Clutter perturbation further highlights this gap: \ours-scratch collapses from 71.6\% to 24.6\% when distractor objects are introduced, suggesting that the ability to attend to task-relevant objects amid clutter requires large-scale robotic data pretraining rather than in-domain fine-tuning alone. Notably, \ours scores higher under Background randomization (74.6\%) than under the Easy setting (73.2\%), likely because diverse real-world scenes in the pretraining data make randomized backgrounds more in-distribution than the plain white background of the Easy mode.
\ours-Context further amplifies these gains, reaching 84.7\% (Easy) and 69.4\% (Hard) under joint control---an improvement of +11.5 and +6.8 points over \ours. This substantial boost suggests that conditioning on intra-episode execution history provides complementary robustness to visual perturbations, as the policy can dynamically calibrate its actions based on observed outcomes rather than relying solely on the current observation.

\paragraph{RoboCasa365.}

\begin{table}[!t]
\centering
\caption{\textbf{Evaluation on RoboCasa365 across atomic and long-horizon manipulation tasks.}}
\label{tab:robocasa365}
\small
\tabcolsep 5pt
\begin{tabular}{lcccc}
\toprule
& \textbf{Atomic} & \textbf{Composite-Seen} & \textbf{Composite-Unseen} & \textbf{Total} \\
\midrule
$\pi_0$~\citep{black2024pi0} & 36.3 & 5.2  & 0.7 & 15.0 \\
$\pi_{0.5}$~\citep{pi05} & 39.6 & 7.1 & 1.2 & 16.9 \\
GR00T-N1.5~\citep{bjorck2025grootn1} & 50.7 & 14.8 & 2.7 & 23.9 \\
GR00T-N1.6~\citep{bjorck2025grootn1} & 51.1 & 9.4 & 1.7 & 21.9 \\
RLDX-1~\citep{kim2026rldx} & 63.0 & \textbf{27.5} & 5.4 & 33.2 \\
\midrule
\ours & \textbf{68.6} & 20.1 & \textbf{14.9} & \textbf{35.9} \\
\ours-Context & 63.9 & {22.6} & 11.2 & 33.8 \\
\bottomrule
\end{tabular}
\end{table}

Table~\ref{tab:robocasa365} reports results across the three evaluation suites. \ours achieves 35.9\% overall, surpassing the previous state-of-the-art RLDX-1 (33.2\%). On the Atomic suite, \ours achieves the highest score (68.6\%) among all methods, reflecting strong generalization across diverse manipulation primitives. On Composite-Unseen---where the robot must complete long-horizon tasks in OOD scenes---\ours achieves 14.9\%, nearly tripling the next-best result (5.4\% for RLDX-1), demonstrating substantially stronger out-of-distribution compositional generalization. 

\begin{table}[!t]
\centering
\caption{\textbf{Evaluation on EBench across three splits.} Each split reports both success rate (SR) and the EBench composite score (Score).}
\label{tab:ebench}
\small
\tabcolsep 2.5pt
\begin{tabular}{lcccccccc}
\toprule
& \multicolumn{2}{c}{\textbf{Table Top}} & \multicolumn{2}{c}{\textbf{Simple PnP}} & \multicolumn{2}{c}{\textbf{Long Horizon}} & \multicolumn{2}{c}{\textbf{Overall}} \\
\cmidrule(lr){2-3} \cmidrule(lr){4-5} \cmidrule(lr){6-7} \cmidrule(lr){8-9}
& SR & Score & SR & Score & SR & Score & SR & Score \\
\midrule
$\pi_{0}$~\citep{black2024pi0} & 15.7 & 30 & 35.0 & 39 & 17.0 & 41 & 23.6 & 37 \\
$\pi_{0.5}$~\citep{pi05} & 12.9 & 32 & 45.0 & 50 & 18.1 & 39 & 27.1 & 41 \\
X-VLA~\citep{zheng2025xvla} & 8.6 & 24 & 50.0 & 54 & 6.2 & 25 & 23.7 & 36 \\
InternVLA-A1~\citep{cai2026internvla} & 4.3 & 11 & 43.0 & 47 & 17.9 & 46 & 23.9 & 36 \\
\midrule
\ours & \textbf{50.0} & \textbf{70} & \textbf{56.5} & 60 & \textbf{29.9} & \textbf{55} & \textbf{45.6} & \textbf{60} \\
\ours-Context & 49.3 & 56 & 55.0 & \textbf{66} & 26.6 & \textbf{55} & 43.6 & 59 \\
\bottomrule
\end{tabular}
\end{table}

\begin{figure}[!t]
    \centering
    \includegraphics[width=1.0\linewidth]{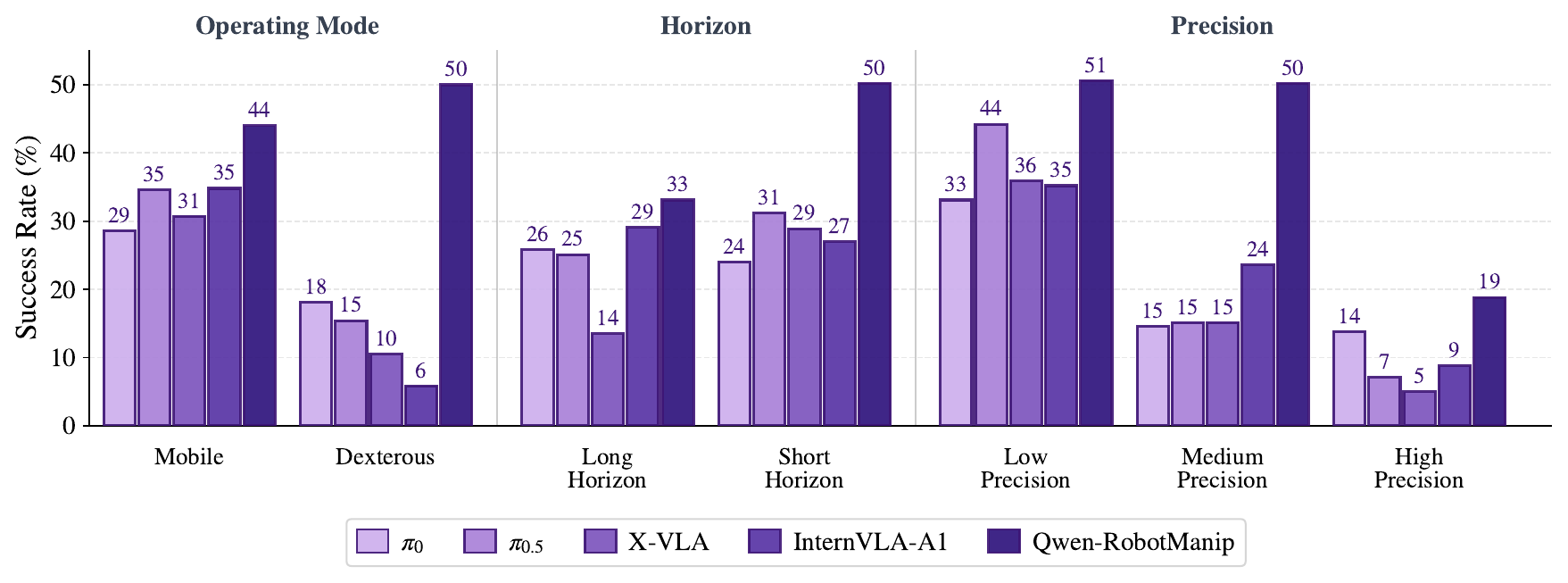}
    \caption{\textbf{EBench performance across operating mode, horizon, and precision.} }
    \label{fig:ebench_topline}
\end{figure}

\begin{figure}[!t]
    \centering
    \includegraphics[width=1.0\linewidth]{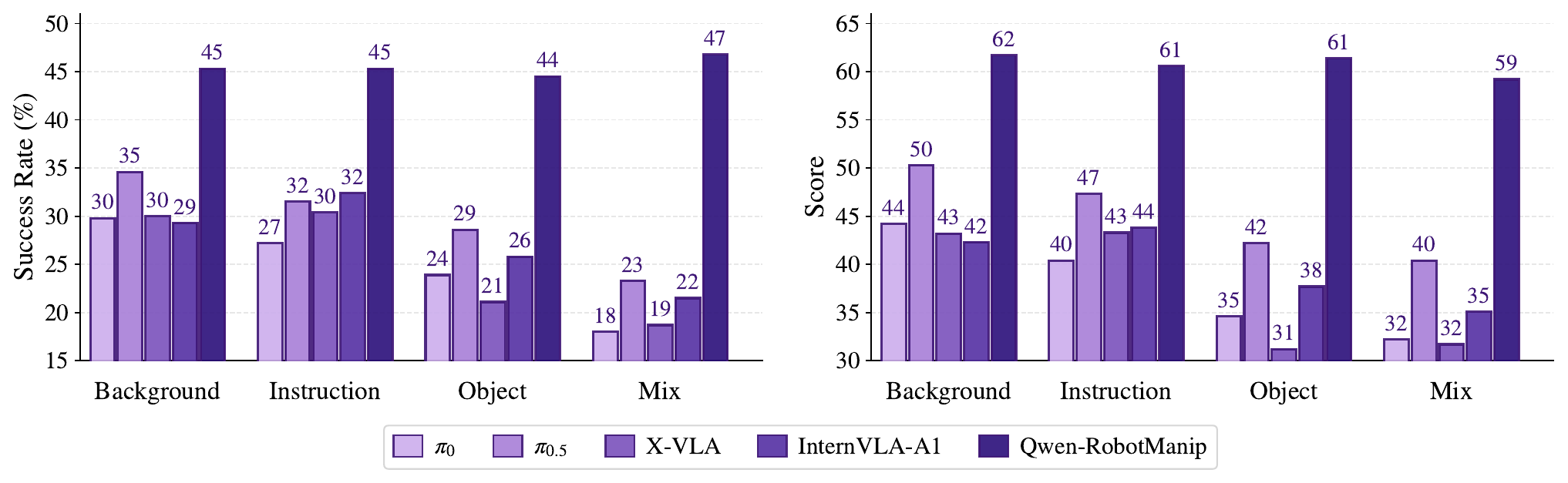}
    \caption{\textbf{Per-dimension generalization breakdown on EBench.} }
    \label{fig:ebench_generalize}
\end{figure}

\paragraph{EBench.}
Table~\ref{tab:ebench} and Figure~\ref{fig:ebench_topline} report the results on EBench~\citep{ebench2026}, an Isaac Sim-based indoor manipulation benchmark that evaluates 26 task types on a dual-arm mobile platform (Lift2 + R5a) across three splits: dexterous tabletop tasks (Table Top), pick-and-place with mobile manipulation (Simple PnP), and extended multi-step sequences (Long Horizon). 
\ours achieves 45.6\% overall success rate and a composite score of 60, outperforming $\pi_{0.5}$ (27.1\% / 41) and all other baselines by a large margin across every split.

The gains are most striking on the dexterous split. 
On Table Top, \ours achieves 50.0\% SR and a score of 70, nearly quadrupling $\pi_{0.5}$'s success rate (12.9\%) and more than doubling its score (32). 
On Simple PnP, \ours leads at 56.5\% SR while the next-best baseline is X-VLA at 50.0\%.
On Long Horizon, which requires mobile manipulation over extended sequences, \ours reaches 29.9\% SR and a score of 55, improving over $\pi_{0.5}$ by +11.8 SR and +16 score.
Figure~\ref{fig:ebench_topline} further breaks down performance by operating mode, horizon length, and precision level, showing consistent advantages across all seven dimensions.

The per-dimension generalization breakdown (Figure~\ref{fig:ebench_generalize}) further confirms robustness to distribution shift. 
As perturbation complexity increases, baseline performance degrades substantially---$\pi_{0.5}$'s SR drops from 34.6 under Background to 23.3 under Mix (a 33\% decline). 
In contrast, \ours remains remarkably stable across all dimensions (SR 44.5--46.8), with virtually no degradation under compounded perturbations.
The gap over the next-best method widens from Background (+10.7 SR) and Instruction (+12.9), to Object (+15.9) and Mix (+23.5). 
Under the most demanding Mix condition---where background, instruction, and object perturbations are applied simultaneously---\ours achieves 46.8\%, even surpassing its own Background score (45.3\%), whereas $\pi_{0.5}$ drops from 34.6 to 23.3 ($-$33\%). 
This near-uniform performance under diverse distribution shifts demonstrates strong generalization robustness.



\paragraph{RoboTwin-IF.}

\begin{table}[!t]
\centering
\caption{\textbf{Instruction-following evaluation on RoboTwin-IF.} Models are fine-tuned on RoboTwin Clean dataset and evaluated with held-out unseen instruction templates. }
\label{tab:robotwin_if}
\small
\tabcolsep 2.5pt
\begin{tabular}{lcccccc}
\toprule
& \textbf{Pick-Diverse} & \textbf{Place-Rel.}  & \textbf{Ope.-Mic-Dr.} & \textbf{Ope.-Stapler} & \textbf{Ope.-Table} & \textbf{Average} \\
\midrule
StarVLA~\citep{community2026starvla} & 11 & 13  & 0 & 49 & 74 & 29.4 \\
GR00T-N1.7~\citep{bjorck2025grootn1} & 20 & 17  & 0 & 14 & 32 & 16.6 \\
$\pi_{0.5}$~\citep{pi05} & 44 & 20 & 15 & \textbf{92} & 66   & 49.6 \\
\midrule
\ours & \textbf{79} & 57 & \textbf{42} & 90 & \textbf{93} & \textbf{72.2} \\
\ours-Context & {77} & \textbf{71} & 33 & 89 & {90} & {72.0} \\ 
\bottomrule
\end{tabular}
\end{table}

Table~\ref{tab:robotwin_if} reports per-suite results. \ours achieves 72.2\% average versus $\pi_{0.5}$'s 49.6\%, a gap of 22.6 points. The improvement happens on four of five suites, with large gains on Pick-Diverse (+35), Place-Relative (+37), Operated-Mic-Drawer (+27), and Operated-Tabletop (+27)---tasks where the instruction must be parsed to determine the correct action among multiple plausible alternatives in the same scene. This comprehensive advantage confirms that \ours has learned genuine language-conditioned control rather than relying on visual shortcuts.

\paragraph{Zero-Shot Cross-Embodiment.}

Table~\ref{tab:cross_embodiment} reports zero-shot transfer for both joint-space and camera-frame relative EEF representations on RoboTwin-XE. 
Joint-space control transfers poorly: joint configurations are robot-specific, so actions meaningful for one morphology produce near-random behavior on another---UR5 and Franka stay below 5\% for both methods. 
Switching to camera-frame EEF dramatically improves transfer: \ours reaches 42.9\% on ARX, 22.8\% on UR5 ($5.6\times$ the joint result), and 5.9\% on Franka, for a cross-embodiment average of 23.9\%. 
This confirms that the camera-frame representation successfully abstracts away morphological differences, allowing the policy to reason in shared Cartesian space. 
Our model consistently outperforms $\pi_{0.5}$ in both action spaces, with the largest gap on UR5 (22.8 vs.\ 10.0). The performance gradient (ARX $>$ UR5 $>$ Franka) correlates with visual and kinematic similarity to the training embodiment: ARX-X5 shares a visually similar 6-DOF form factor and similar reach to AgileX, UR5 differs in appearance and joint arrangement but has a similar workspace, while Franka's distinct 7-DOF morphology and larger reach present the greatest mismatch.

\begin{figure}[!t]
\centering
\includegraphics[width=0.8\columnwidth]{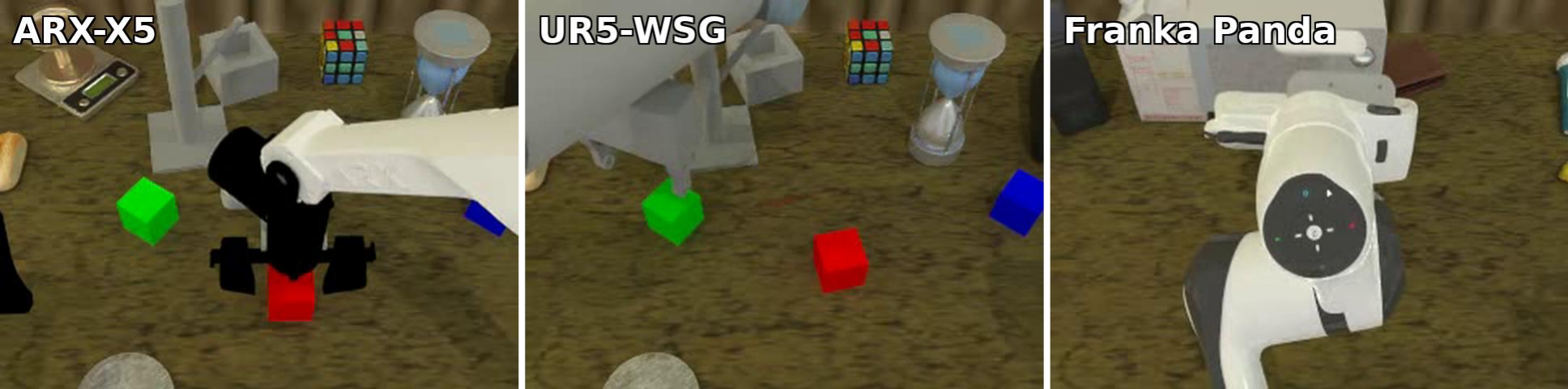}
\caption{\textbf{Zero-shot cross-embodiment evaluation on RoboTwin-XE.} 
Left: ARX-X5. Middle: UR5-WSG. Right: Franka Panda.
}
\label{fig:cross_embodiment}
\end{figure}

\begin{table}[!t]
\centering
\caption{\textbf{Zero-shot performance on RoboTwin-XE}, where models are trained on RoboTwin Clean and tested on Hard settings with unseen robot embodiments. 
}
\label{tab:cross_embodiment}
\small
\tabcolsep 5pt
\begin{tabular}{lcccc}
\toprule
& \textbf{ARX-X5} & \textbf{UR5-WSG} & \textbf{Franka Panda} & \textbf{Total} \\
\midrule
$\pi_{0.5}$ (joint)~\citep{pi05} & 24.6 & 2.2  & 0.9 & 9.2 \\
$\pi_{0.5}$ (eef)~\citep{pi05} & 11.5 & 10.0 & 1.1 & 7.5 \\
\midrule
\ours (joint) & 37.6 & 4.1 & 1.8 & 14.5 \\
\ours (eef) & \textbf{42.9} & \textbf{22.8} & \textbf{5.9} & \textbf{23.9} \\
\bottomrule
\end{tabular}
\end{table}

\subsection{Real-World Evaluation}

\subsubsection{Evaluation on ALOHA Platforms}

\begin{figure}[!t]
\centering
\includegraphics[width=\columnwidth]{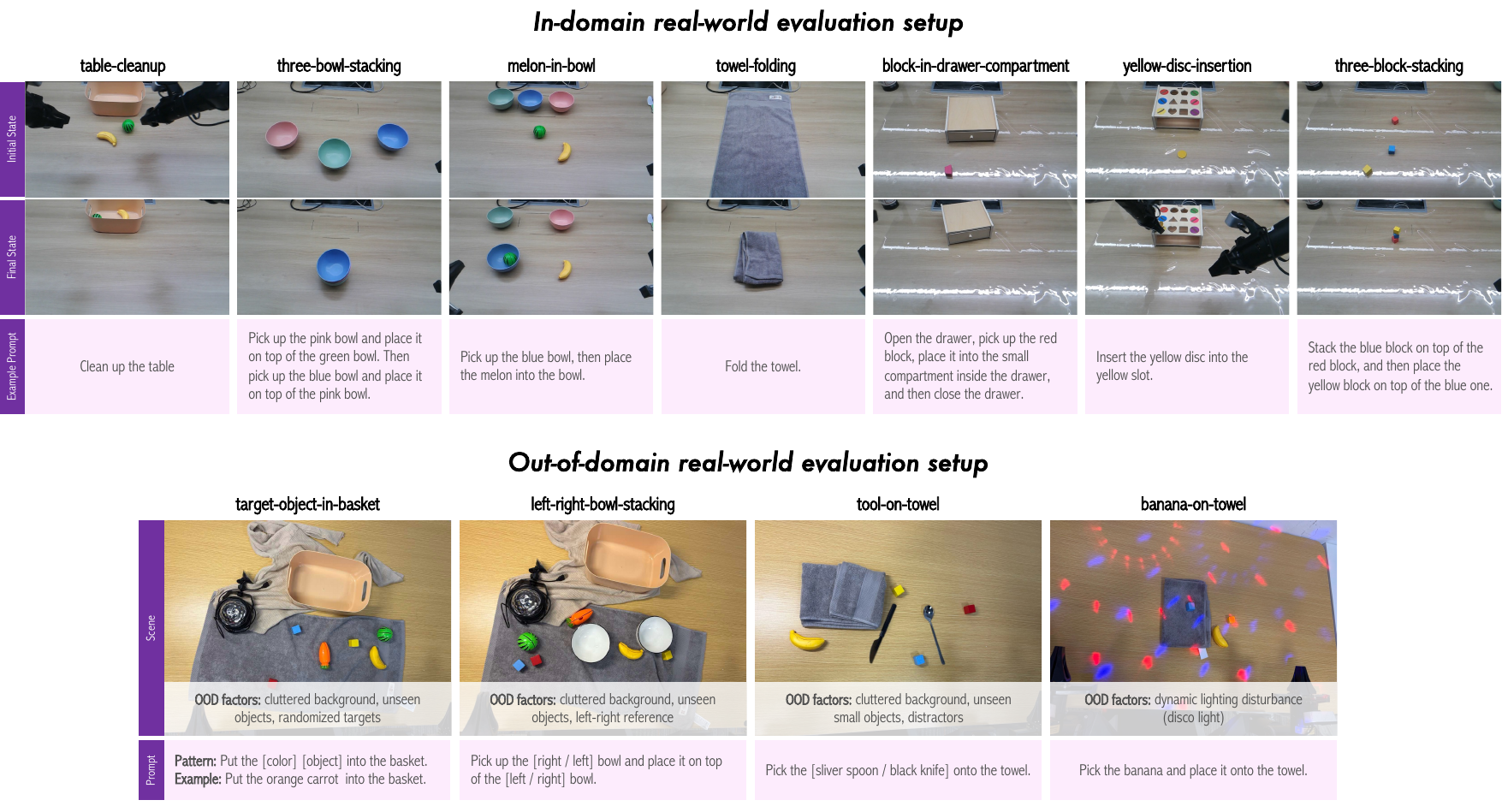}
\caption{\textbf{In-domain and out-of-domain tasks of real-world CobotMagic ALOHA platform.}}
\label{fig:real-world-cobotmagic-setup}
\end{figure}

\textbf{In-Domain (ID) and Out-of-Domain (OOD) Evaluation on CobotMagic ALOHA.}
We fine-tune \ours on 22.9 hours of teleoperated demonstrations collected on the CobotMagic ALOHA platform, covering multiple bimanual manipulation tasks. We then evaluate the fine-tuned policy on real-world ID and OOD benchmarks (Figure~\ref{fig:real-world-cobotmagic-setup}) constructed on the same platform.

The ID benchmark includes seven tasks: \textit{table-cleanup} (clearing objects from the table), \textit{three-bowl-stacking} (stacking three bowls in the specified order), \textit{melon-in-bowl} (placing a small spherical melon toy into a bowl), \textit{towel-folding} (folding a towel), \textit{place-block-into-drawer} (opening a drawer, placing a block inside, and closing the drawer), \textit{yellow-disc-insertion} (picking up a disc, performing a bimanual handover, and inserting it into a matching slot), and \textit{three-block-stacking} (stacking three blocks in sequence). These tasks span a wide range of manipulation difficulty.

At the easier end, \textit{table-cleanup}, \textit{three-bowl-stacking}, and \textit{melon-in-bowl} primarily test basic language grounding and sequential pick-and-place execution. \textit{Towel-folding} is more challenging due to the deformable nature of the object. The remaining tasks are substantially harder: \textit{place-block-into-drawer} is particularly challenging because it requires the robot to accurately grasp the small drawer handle and perform precise pulling and pushing motions to open and close the drawer, \textit{yellow-disc-insertion} requires fine-grained pose alignment for contact-rich insertion, and \textit{three-block-stacking} demands reliable relational reasoning and robust multi-step execution, while also testing the model's ability to recover from intermediate failures, since blocks may slip or collapse during the stacking process.

As shown in Table~\ref{tab:cobotmagic_id}, \ours achieves an average success rate of 88.6\%, significantly outperforming $\pi_{0.5}$ (42.9\%) and StarVLA (20.0\%). It succeeds in all 5 trials on five tasks and remains strong on \textit{towel-folding} (4/5). The only task with noticeable room for improvement is \textit{yellow-disc-insertion} (2/5), highlighting the difficulty of precise insertion on real hardware. In comparison, $\pi_{0.5}$ performs reasonably well on relatively simple tasks such as \textit{table-cleanup}, \textit{three-bowl-stacking}, and \textit{towel-folding}, but its performance degrades sharply on more challenging tasks that require long-horizon planning, precise contact-rich manipulation, or recovery from intermediate failures. StarVLA performs poorly across almost all tasks, with limited success even on the easier ones. These results indicate that \ours not only handles basic real-world manipulation reliably, but also scales much better to more challenging task settings.

We further evaluate real-world generalization in an out-of-domain (OOD) setting on CobotMagic ALOHA. Compared with the ID benchmark, these tasks introduce distribution shifts in visual scenes, object instances, and task instructions, and are designed to test whether the model can robustly generalize beyond the seen distribution.

\begin{table}[!t]
\centering
\caption{\textbf{In-domain real-world evaluation on the CobotMagic ALOHA platform.} We report task success over 5 trials for each task.}
\label{tab:cobotmagic_id}
\small
\setlength{\tabcolsep}{4pt}
\begin{tabular}{lccc}
\toprule
\textbf{Task} & \textbf{$\pi_{0.5}$} & \textbf{StarVLA} & \textbf{\ours} \\
\midrule
table-cleanup & 4/5 & 0/5 & 5/5 \\
three-bowl-stacking & 5/5 & 4/5 & 5/5 \\
melon-in-bowl & 2/5 & 0/5 & 5/5 \\
towel-folding & 4/5 & 3/5 & 4/5 \\
block-in-drawer-compartment & 0/5 & 0/5 & 5/5 \\
yellow-disc-insertion & 0/5 & 0/5 & 2/5 \\
three-block-stacking & 0/5 & 0/5 & 5/5 \\
\midrule
\textbf{Average success rate} & \textbf{42.9\%} & \textbf{20.0\%} & \textbf{88.6\%} \\
\bottomrule
\end{tabular}
\end{table}

\begin{table}[!t]
\centering
\caption{\textbf{Out-of-domain real-world evaluation on the CobotMagic ALOHA platform.} We report task success over 10 trials for each task.}
\label{tab:cobotmagic_ood}
\small
\setlength{\tabcolsep}{4pt}
\begin{tabular}{l p{4.3cm} ccc}
\toprule
\textbf{Task} & \textbf{OOD factors} & \textbf{$\pi_{0.5}$} & \textbf{StarVLA} & \textbf{\ours} \\
\midrule
target-object-in-basket & cluttered background, unseen objects, randomized targets & 8/10 & 0/10 & 10/10 \\
left-right-bowl-stacking & cluttered background, unseen objects, left-right reference & 1/10 & 0/10 & 10/10 \\
tool-on-towel & cluttered background, unseen small objects, distractors & 0/10 & 0/10 & 6/10 \\
banana-on-towel & dynamic lighting disturbance (disco light) & 6/10 & 0/10 & 9/10 \\
\midrule
\textbf{Average success rate} & -- & \textbf{37.5\%} & \textbf{0.0\%} & \textbf{87.5\%} \\
\bottomrule
\end{tabular}
\end{table}

The OOD benchmark contains four tasks: \textit{target-object-in-basket} (placing the instructed objects into a basket), \textit{left-right-bowl-stacking} (stacking one bowl onto another based on left-right references), \textit{tool-on-towel} (placing a specified tool onto a towel), and \textit{banana-on-towel} (placing a banana onto a towel). Each task targets a different aspect of generalization. \textit{Target-object-in-basket} introduces cluttered backgrounds, varied objects, and randomized target objects, testing whether the model can correctly identify and manipulate the instructed object under visual variation. \textit{Left-right-bowl-stacking} combines cluttered scenes and left-right relational references, requiring robust spatial language understanding. \textit{Tool-on-towel} further increases difficulty by introducing unseen and physically small objects (i.e., knife and spoon) together with distractors, making both target identification and stable grasping more difficult. Finally, \textit{banana-on-towel} evaluates robustness to severe illumination changes induced by a disco light in the real-world environment.

As shown in Table~\ref{tab:cobotmagic_ood}, \ours generalizes substantially better than the baselines in the OOD setting, achieving an average success rate of 87.5\%, compared with 37.5\% for $\pi_{0.5}$ and 0.0\% for StarVLA. In particular, \ours attains perfect success on \textit{target-object-in-basket} and \textit{left-right-bowl-stacking}, demonstrating strong robustness to cluttered scenes, attribute-based grounding, and left-right spatial references. It also remains effective on the more difficult \textit{tool-on-towel} task (6/10), where the robot must identify novel target objects under object distractors, and on \textit{banana-on-towel} (9/10) under lighting variation.

In contrast, $\pi_{0.5}$ retains some robustness on relatively simpler OOD variations, achieving 8/10 on \textit{target-object-in-basket} and 6/10 on \textit{banana-on-towel}. However, its performance collapses on tasks requiring stronger compositional and relational generalization, obtaining only 1/10 on \textit{left-right-bowl-stacking} and 0/10 on \textit{tool-on-towel}. StarVLA fails on all four OOD tasks. These results suggest that \ours preserves much stronger visual-linguistic representation and is considerably more robust to scene clutter, novel objects, spatial references, and lighting changes in real-world settings.




\begin{figure}[!t]
\centering
\includegraphics[width=\columnwidth]{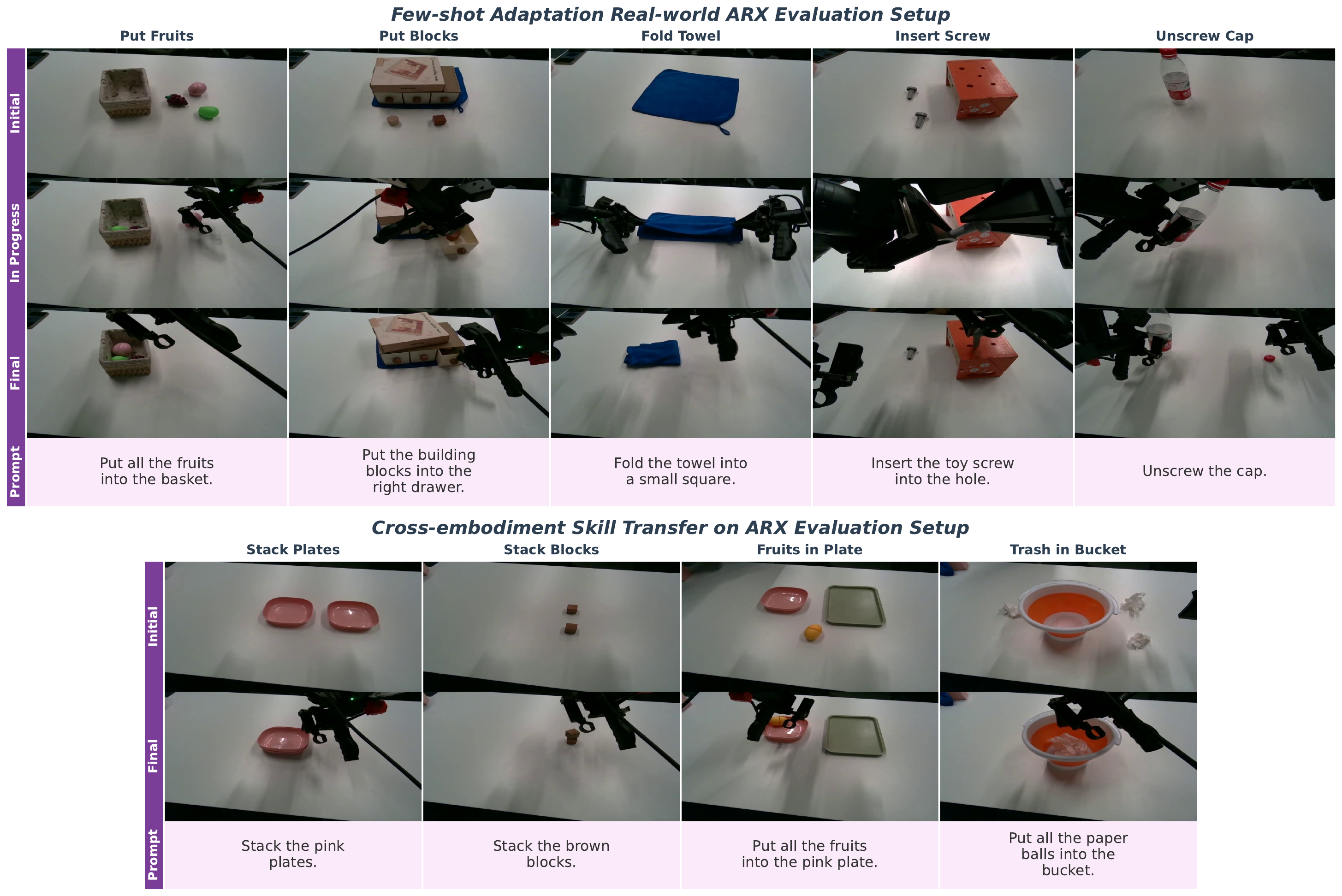}
\caption{\textbf{Real-world evaluation setup on the ARX ALOHA platform.}}
\label{fig:real-world-arx-setup}
\end{figure}

\begin{table}[!t]
\centering
\caption{\textbf{Few-shot adaptation on ARX ALOHA.} Sub-step success over 10 trials. All methods are jointly fine-tuned on 130 teleoperated demonstrations (50 for Unscrew Cap, 20 for each other task).}
\label{tab:real_arx}
\small
\setlength{\tabcolsep}{4pt}
\begin{tabular}{ll ccc}
\toprule
\textbf{Task} & \textbf{Sub-step} & \textbf{StarVLA} & \textbf{$\pi_{0.5}$} & \textbf{\ours} \\
\midrule
\multirow{4}{*}{Put Fruits} 
& Place fruit 1 & 3/10 & 9/10 & 9/10 \\
& Place fruit 2 & 1/10 & 5/10 & 5/10 \\
& Place fruit 3 & 0/10 & 2/10 & \textbf{3/10} \\
& \cellcolor{gray!10}\textit{Avg. success} & \cellcolor{gray!10}13.3\% & \cellcolor{gray!10}53.3\% & \cellcolor{gray!10}\textbf{56.7\%} \\
\midrule
\multirow{5}{*}{Put Blocks}
& Open drawer & 1/10 & 4/10 & \textbf{5/10} \\
& Place block 1 & 1/10 & 2/10 & \textbf{4/10} \\
& Place block 2 & 0/10 & 2/10 & \textbf{3/10} \\
& Close drawer & 0/10 & 2/10 & \textbf{3/10} \\
& \cellcolor{gray!10}\textit{Avg. success} & \cellcolor{gray!10}5.0\% & \cellcolor{gray!10}25.0\% & \cellcolor{gray!10}\textbf{37.5\%} \\
\midrule
\multirow{3}{*}{Fold Towel}
& First fold & 0/10 & 3/10 & 3/10 \\
& Second fold & 0/10 & 1/10 & \textbf{3/10} \\
& \cellcolor{gray!10}\textit{Avg. success} & \cellcolor{gray!10}0.0\% & \cellcolor{gray!10}20.0\% & \cellcolor{gray!10}\textbf{30.0\%} \\
\midrule
\multirow{3}{*}{Insert Screw}
& Handover screw & 0/10 & 2/10 & 2/10 \\
& Insert screw & 0/10 & 0/10 & 0/10 \\
& \cellcolor{gray!10}\textit{Avg. success} & \cellcolor{gray!10}0.0\% & \cellcolor{gray!10}10.0\% & \cellcolor{gray!10}10.0\% \\
\midrule
\multirow{4}{*}{Unscrew Cap}
& Grasp bottle & 4/10 & 9/10 & 9/10 \\
& Unscrew cap & 0/10 & 2/10 & \textbf{4/10} \\
& Place down & 0/10 & 1/10 & \textbf{3/10} \\
& \cellcolor{gray!10}\textit{Avg. success} & \cellcolor{gray!10}13.3\% & \cellcolor{gray!10}40.0\% & \cellcolor{gray!10}\textbf{53.3\%} \\
\bottomrule
\end{tabular}
\end{table}

\textbf{Few-shot Adaptation on ARX ALOHA.}
We compare \ours against two baselines on five real-world manipulation tasks (Figure~\ref{fig:real-world-arx-setup}, top; Table~\ref{tab:real_arx}): $\pi_{0.5}$~\citep{pi05}, a pretrained open-source VLA, and StarVLA~\citep{community2026starvla} trained from scratch without pretraining. 
All methods use joint positions and EEF pose as state and predict actions in EEF space, jointly fine-tuned on the same 130 teleoperated demonstrations (50 for Unscrew Cap, 20 for each other task). 
The five tasks span multi-object pick-and-place (Put Fruits), long-horizon sequential manipulation (Put Blocks), deformable object handling (Fold Towel), bimanual precision assembly (Insert Screw), and fine-grained rotational control (Unscrew Cap).

\ours outperforms both baselines on four of five tasks. The largest gain appears on Put Blocks (15/40 vs.\ 10/40 for $\pi_{0.5}$), where \ours maintains higher sub-step success through all four stages---open drawer, place two blocks, and close drawer---indicating more robust long-horizon execution. 
On Unscrew Cap, \ours doubles the cap-removal rate over $\pi_{0.5}$ (4/10 vs.\ 2/10) and triples the full-task completion (3/10 vs.\ 1/10), showing stronger fine-grained rotational control. 
Fold Towel improves notably on the second fold (3/10 vs.\ 1/10). 
Insert Screw remains challenging for all methods, with no model completing a full insertion (0/10), though both $\pi_{0.5}$ and \ours succeed at the handover sub-step (2/10). 
StarVLA, without pretraining, achieves near-zero across tasks, confirming the importance of large-scale pretraining for real-world manipulation.

\begin{table}[!t]
\centering
\caption{\textbf{Cross-embodiment skill transfer on ARX.}}
\label{tab:cross_embodiment_transfer}
\small
\tabcolsep 3pt
\begin{tabular}{lccccc}
\toprule
& \textbf{Stack Plates} & \textbf{Stack Blocks} & \textbf{Fruits in Plate} & \textbf{Trash in Bucket} & \textbf{Avg.} \\
\midrule
\ours w/o UnifiedSpace & 0/10 & 0/10 & 3/10 & 0/10 & 7.5\% \\
\ours w/o UnifiedEEF & 0/10 & 0/10 & 5/10 & 0/10 & 12.5\% \\
\midrule
\ours & \textbf{3/10} & \textbf{5/10} & \textbf{7/10} & \textbf{7/10} & \textbf{55.0\%} \\
\bottomrule
\end{tabular}
\end{table}

\textbf{Cross-embodiment transfer.}
We also investigate cross-embodiment skill transfer (Figure~\ref{fig:real-world-arx-setup}, bottom; Table~\ref{tab:cross_embodiment_transfer}). 
A single policy is jointly fine-tuned on 6K CobotMagic and 130 ARX demonstrations, using the same state-action parameterization as above, then evaluated on four novel tasks on the ARX platform: stacking two plates, stacking blocks, placing fruits into a designated pink plate with distractor plates, and collecting paper balls into a bucket. 
ARX has \emph{zero demonstrations} for any of these tasks---the relevant manipulation skills (stacking, precise placement, object collection) must generalize from related but not identical CobotMagic behaviors across a kinematically different embodiment. 
We ablate two key components: \ours w/o UnifiedSpace removes the unified action-space mapping and instead naively concatenates and zero-pads each robot's action dimensions without semantic alignment; \ours w/o UnifiedEEF removes the unified EEF representation while retaining a basic slot layout with rotation unification and gripper normalization. 
Both variants fail almost entirely (7.5\% and 12.5\%), showing that neither surface-level dimension alignment nor partial unification can bridge the combined embodiment and task gap. 
\ours achieves 55.0\%---over 4$\times$ the best variant---succeeding on all four tasks including Stack Blocks (5/10) and Trash in Bucket (7/10). 
The full unified action space and EEF representation enable \ours to learn a stronger manipulation representation through large-scale diverse pretraining, enabling skill-level transfer: a policy can leverage skills learned from one embodiment to execute novel tasks on another embodiment with minimal training data and no task-specific demonstrations.



\subsubsection{Table30-v1 Challenge}

\begin{table*}[!t]
\centering
\caption{\textbf{Per-task results on the RoboChallenge Table30 v1 Generalist Track.} Each cell reports success rate (\%) / process score. Best results per task are in \textbf{bold}.}
\label{tab:table30_results}
\resizebox{\textwidth}{!}{
\begin{tabular}{llccccc}
\toprule
Robot & Task & \ours & DM0\_generalist & pi05\_generalist & GR00T-MULTI & pi0\_generalist \\
\midrule
\multicolumn{2}{l}{\textbf{Average (all)}} & \textbf{45 / 59.83} & 37 / 48.43 & 17.67 / 31.27 & 15.33 / 32.29 & 9 / 20.22 \\
\midrule
\multirow{12}{*}{ARX5}
& arrange flowers & \textbf{30 / 64} & 20 / 49 & 0 / 30.5 & 20 / 57 & 0 / 13.5 \\
& arrange paper cups & \textbf{70 / 83} & 10 / 54 & 0 / 31 & 0 / 19 & 0 / 15 \\
& fold dishcloth & \textbf{30 / 48} & 10 / 10.5 & 0 / 0 & 0 / 17 & 0 / 0 \\
& open the drawer & 0 / 47 & \textbf{90 / 95} & 50 / 80 & 0 / 50 & 0 / 20 \\
& place shoes on rack & 70 / 85 & \textbf{100 / 98.5} & 0 / 20 & 40 / 54.5 & 0 / 16.5 \\
& put cup on coaster & 100 / 99 & \textbf{100 / 100} & 70 / 63 & 80 / 92 & 0 / 0 \\
& search green boxes & 90 / 92 & \textbf{100 / 95.5} & 0 / 3 & 30 / 37.5 & 0 / 0 \\
& sort electronic products & \textbf{50 / 60.4} & 0 / 18.4 & 0 / 22.5 & 10 / 34.8 & 0 / 22.5 \\
& turn on light switch & 70 / 69.5 & \textbf{70 / 70.5} & 10 / 25 & 0 / 5 & 20 / 29 \\
& water potted plant & 0 / 9 & 0 / \textbf{33.5} & 0 / 0 & 0 / 6 & 0 / 0 \\
& wipe the table & 0 / \textbf{72.5} & 0 / 47.5 & 10 / 28 & 0 / 66 & 0 / 29 \\
\addlinespace[2pt]\cmidrule{2-7}\addlinespace[2pt]
& \textit{Avg (ARX5)} & \textbf{\textit{46.4 / 66.3}} & \textit{45.5 / 61.1} & \textit{12.7 / 27.5} & \textit{16.4 / 39.9} & \textit{1.8 / 13.2} \\
\midrule
\multirow{12}{*}{ALOHA}
& clean dining table & 20 / 57.5 & 0 / 12 & \textbf{30 / 62} & 10 / 11.5 & 0 / 25.5 \\
& make vegetarian sandwich & \textbf{10 / 46.5} & 0 / 15 & 0 / 0 & 0 / 7 & 0 / 0 \\
& plug in network cable & \textbf{30 / 51} & 10 / 26 & 0 / 0 & 0 / 3 & 0 / 0 \\
& pour fries into plate & \textbf{30 / 56} & 0 / 6 & 0 / 0 & 0 / 36 & 0 / 0 \\
& put opener in drawer & 0 / 0 & 10 / 10 & \textbf{20 / 38} & 0 / 0 & 0 / 0 \\
& put pen into pencil case & \textbf{90 / 95} & 20 / 40 & 50 / 63.5 & 20 / 58 & 0 / 14.5 \\
& scan QR code & 10 / 13 & 0 / 0 & 0 / 7 & \textbf{30 / 26.5} & 0 / 3 \\
& stack bowls & \textbf{100 / 98.5} & 70 / 71 & 80 / 83 & 40 / 55.5 & 40 / 53.5 \\
& stick tape to box & 0 / \textbf{22.5} & 0 / 14 & 0 / 16 & 0 / 2 & 0 / 0 \\
& sweep the rubbish & \textbf{70 / 84.5} & 30 / 40 & 10 / 46 & 10 / 10 & 0 / 17 \\
& turn on faucet & \textbf{100 / 100} & 70 / 84.5 & 60 / 56 & 20 / 44 & 60 / 67.5 \\
\addlinespace[2pt]\cmidrule{2-7}\addlinespace[2pt]
& \textit{Avg (ALOHA)} & \textbf{\textit{41.8 / 56.8}} & \textit{19.1 / 29.0} & \textit{22.7 / 33.8} & \textit{11.8 / 23.0} & \textit{9.1 / 16.5} \\
\midrule
\multirow{7}{*}{UR5}
& arrange fruits in basket & \textbf{80 / 89.5} & 70 / 87 & 0 / 9 & 30 / 54.5 & 0 / 11.5 \\
& hang toothbrush cup & 60 / 80 & \textbf{90 / 95} & 50 / 71 & 70 / 85 & 20 / 62 \\
& set the plates & \textbf{100 / 87.5} & 60 / 62 & 40 / 49.5 & 0 / 24 & 50 / 69.5 \\
& shred scrap paper & 0 / 15 & \textbf{30 / 45} & 20 / 36 & 0 / 0 & 20 / 27 \\
& sort books & 0 / 9 & 0 / 8.5 & 0 / 24 & 0 / 6.5 & \textbf{10 / 26.5} \\
& stack color blocks & 70 / 84.5 & \textbf{100 / 100} & 10 / 30 & 0 / 44.5 & 30 / 39 \\
\addlinespace[2pt]\cmidrule{2-7}\addlinespace[2pt]
& \textit{Avg (UR5)} & \textit{51.7 / 60.9} & \textbf{\textit{58.3 / 66.2}} & \textit{20.0 / 36.6} & \textit{16.7 / 35.8} & \textit{21.7 / 39.2} \\
\midrule
\multirow{3}{*}{Franka}
& move objects into box & \textbf{70 / 75.5} & 50 / 64.5 & 20 / 40 & 50 / 62 & 20 / 44.5 \\
& press three buttons & 0 / 0 & 0 / 0 & 0 / \textbf{4} & 0 / 0 & 0 / 0 \\
\addlinespace[2pt]\cmidrule{2-7}\addlinespace[2pt]
& \textit{Avg (Franka)} & \textbf{\textit{35.0 / 37.8}} & \textit{25.0 / 32.2} & \textit{10.0 / 22.0} & \textit{25.0 / 31.0} & \textit{10.0 / 22.2} \\
\bottomrule
\end{tabular}
}
\end{table*}

To evaluate the generalization capability of \ours, we submit to the RoboChallenge Table30 v1 benchmark under the generalist track. This benchmark comprises 30 manipulation tasks distributed across 4 robot embodiments, and offers two evaluation tracks: the \textit{specialist} track, where a dedicated policy is trained for each individual task (requiring 30 separate models), and the \textit{generalist} track, where a single unified policy is trained per embodiment to handle all tasks associated with that robot. The generalist setting is substantially more challenging, as it demands that the policy generalize across diverse manipulation skills within each embodiment rather than overfitting to a single task. We focus on this track as it better reflects real-world desideratum of building versatile manipulation policies.

Specifically, we post-train \ours on the demonstration data provided by Table30 v1 using joint control, and submit under the anonymous identity \texttt{Lira\_generalist}\footnote{See \texttt{Lira\_generalist} in \url{https://robochallenge.cn/home}.}. \ours achieves a success rate of 45\% and a process score of 59.83. Notably, \ours surpasses \texttt{DM0\_generalist} (37\% success rate, 48.43 process score) by 8 percentage points in success rate and 11.4 in process score, demonstrating its strong multi-task generalization ability. We highlight three key findings from detailed analysis of the benchmark results.

\begin{figure}[!t]
    \centering
    \includegraphics[width=\linewidth]{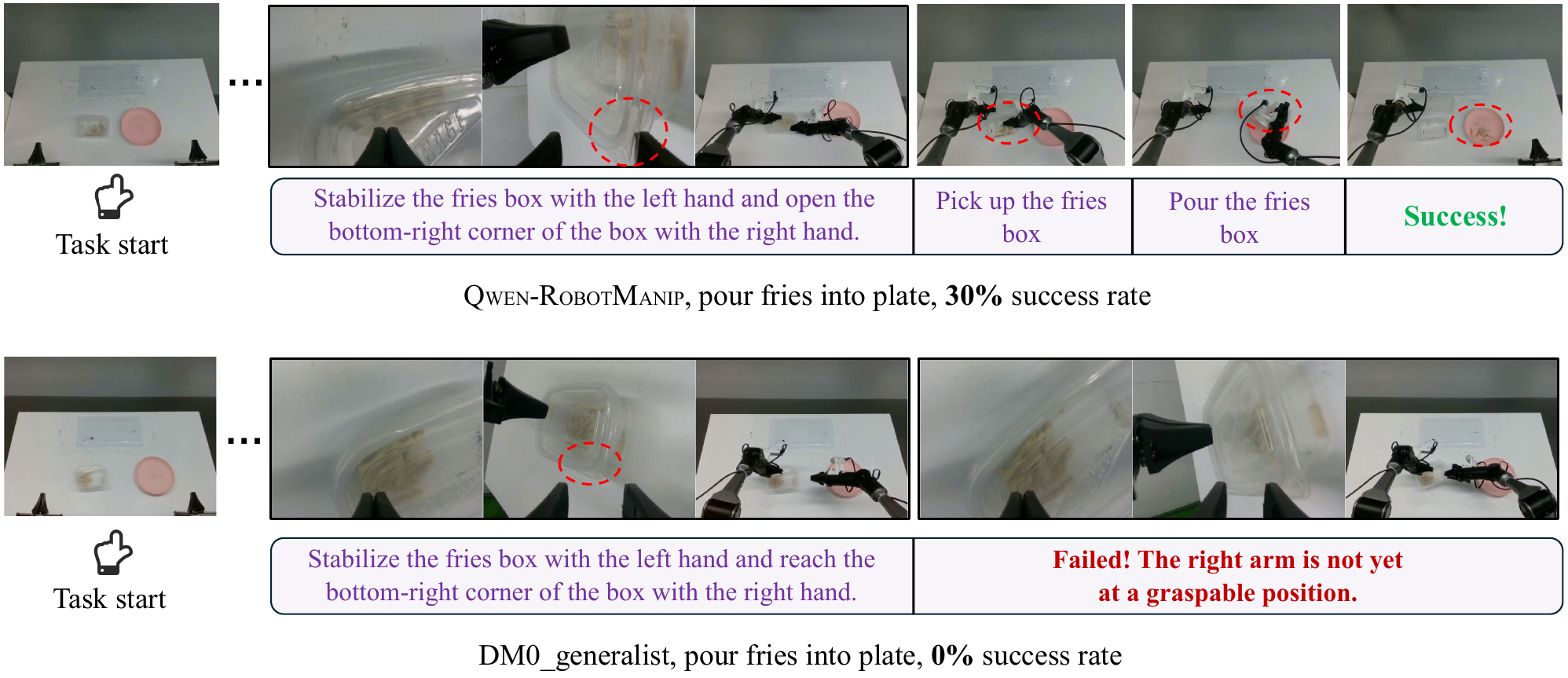}
    \caption{\textbf{Case study on \textit{pour fries into plate} (ALOHA).} \ours (top) coordinates both arms to stabilize, open, pick up, and pour the fries box, completing the task successfully. DM0\_generalist (bottom) fails at the initial stage as the right arm never reaches a graspable position.}
    \label{fig:case_bimanual}
\end{figure}

\begin{figure}[!t]
    \centering
    \includegraphics[width=\linewidth]{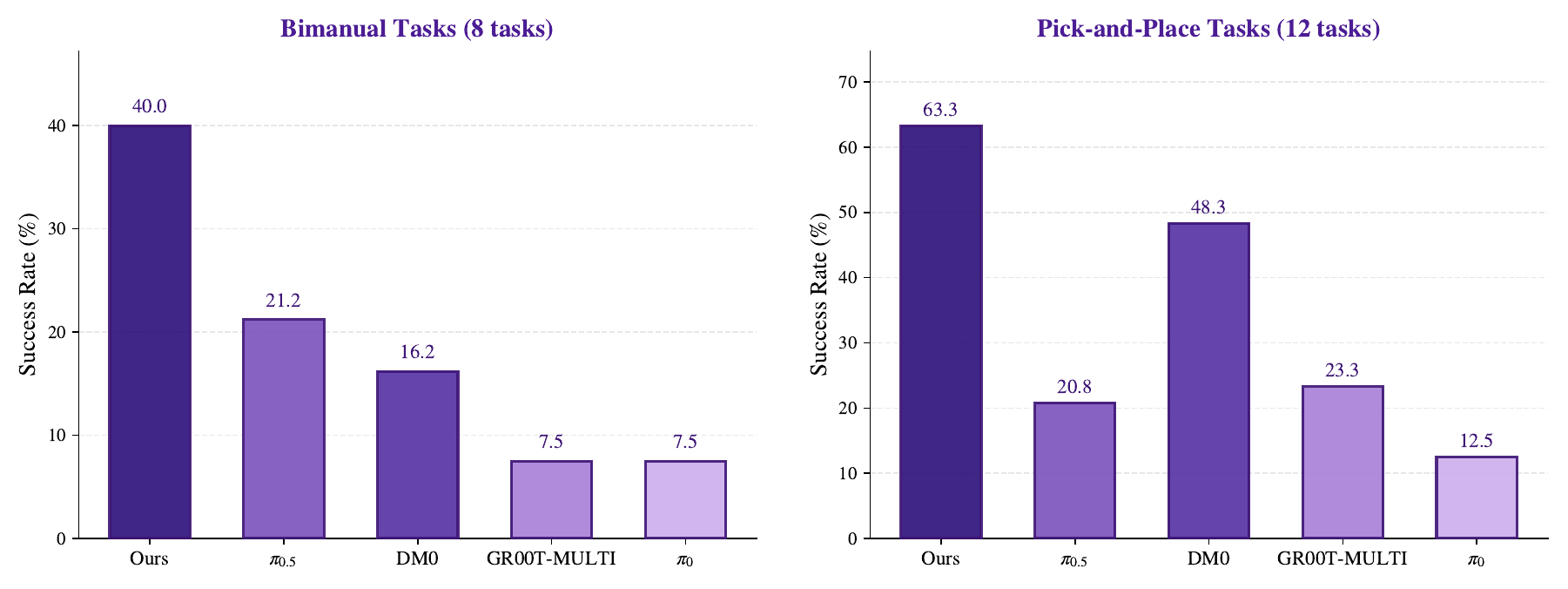}
    \caption{\textbf{Average success rate on bimanual coordination tasks (left, 8 tasks) and pick-and-place tasks (right, 12 tasks).} \ours substantially outperforms all baselines in both categories.}
    \label{fig:bimanual_pickplace_bar}
\end{figure}

\paragraph{Strong bimanual coordination.}
Among the 30 benchmark tasks, 8 require tight bimanual coordination on the ALOHA platform\footnote{The 8 bimanual tasks: \textit{clean dining table}, \textit{make vegetarian sandwich}, \textit{pour fries into plate}, \textit{put opener in drawer}, \textit{put pen into pencil case}, \textit{stick tape to box}, \textit{sweep the rubbish}, \textit{turn on faucet}.}, where the two arms must jointly stabilize, transport, and manipulate objects. \ours achieves an average success rate of 40\% on these tasks, far exceeding $\pi_{0.5}$ (21.2\%), DM0 (16.2\%), GR00T-MULTI (7.5\%), and $\pi_{0}$ (7.5\%) (Figure~\ref{fig:bimanual_pickplace_bar}, left). Notably, \ours is the \emph{only} model to succeed on \textit{pour fries into plate} (30\% vs.\ 0\% for all baselines), a task demanding sequential bimanual steps---stabilizing the fries box with the left arm, opening it with the right arm, picking it up, and pouring the contents onto the plate. As illustrated in Figure~\ref{fig:case_bimanual}, \ours completes this full sequence, while DM0\_generalist fails at the initial coordination stage: the right arm never reaches a graspable position on the box, resulting in immediate task failure. We attribute this strong bimanual performance to two factors: (1)~our pre-training corpus contains a substantial proportion of bimanual demonstration robot data, enabling the model to effectively learn coordinated dual-arm control primitives; and (2)~our Human2Robot pipeline, which synthesizes bimanual robot data from egocentric human videos, further expands the effective bimanual pre-training data and exposes the model to diverse manipulation strategies beyond those captured in teleoperated demonstrations alone.

\begin{figure}[!t]
    \centering
    \includegraphics[width=\linewidth]{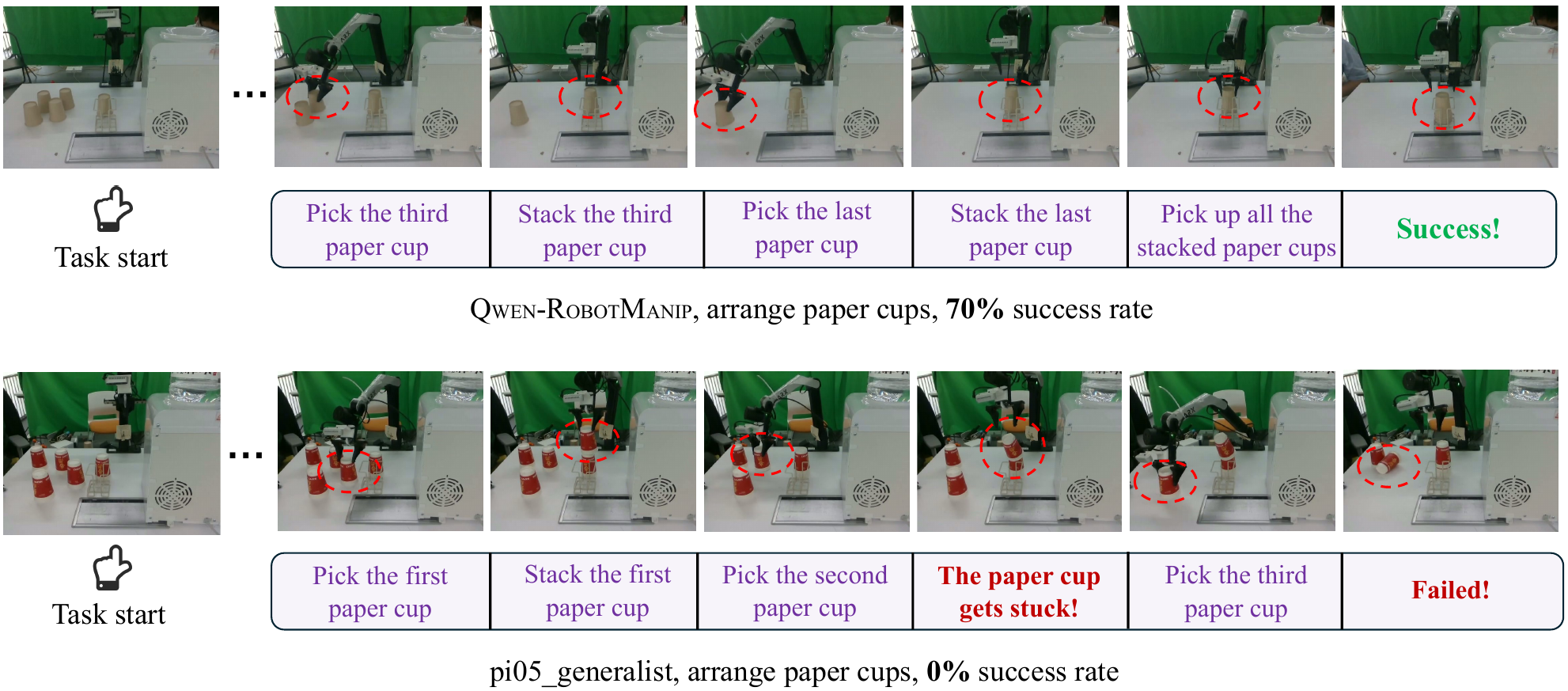}
    \caption{\textbf{Case study on \textit{arrange paper cups} (ARX5).} \ours (top) sequentially picks, stacks, and collects all paper cups with precise placement. $\pi_{0.5}$\_generalist (bottom) encounters a stuck cup during stacking and fails to recover.}
    \label{fig:case_pickplace}
\end{figure}

\paragraph{Robust pick-and-place across embodiments.}
We identify 12 tasks across all four platforms that center on pick-and-place primitives\footnote{The 12 pick-and-place tasks are: \textit{arrange flowers}, \textit{arrange fruits in basket}, \textit{arrange paper cups}, \textit{clean dining table}, \textit{move objects into box}, \textit{set the plates}, \textit{sort books}, \textit{sort electronic products}, \textit{stack bowls}, \textit{place shoes on rack}, \textit{put cup on coaster}, and \textit{stack color blocks}.}, ranging from single-object grasping (\textit{put cup on coaster} and \textit{stack color blocks}) to multi-step sequential manipulation involving 4--5 objects (\textit{arrange paper cups} and \textit{sort electronic products}). As shown in Figure~\ref{fig:bimanual_pickplace_bar}~(right), \ours achieves 63.3\% average success rate on these tasks, surpassing the next-best baseline DM0 (48.3\%) by 15.0 percentage points.  We attribute this capability to two factors: (1)~the large-scale cross-embodiment pre-training data encodes abundant pick-and-place patterns, and (2)~the unified action space enables knowledge sharing of fundamental spatial skills across different robot morphologies. Figure~\ref{fig:case_pickplace} shows a representative comparison on \textit{arrange paper cups}, a task requiring sequential pick-and-place of multiple cups followed by stacking. \ours (70\% SR) successfully picks each cup in sequence, stacks them precisely, while $\pi_{0.5}$\_generalist (0\% SR) encounters a stuck cup during stacking and fails to recover, leading to cascading errors in subsequent steps.

\begin{figure}[!t]
    \centering
    \includegraphics[width=\linewidth]{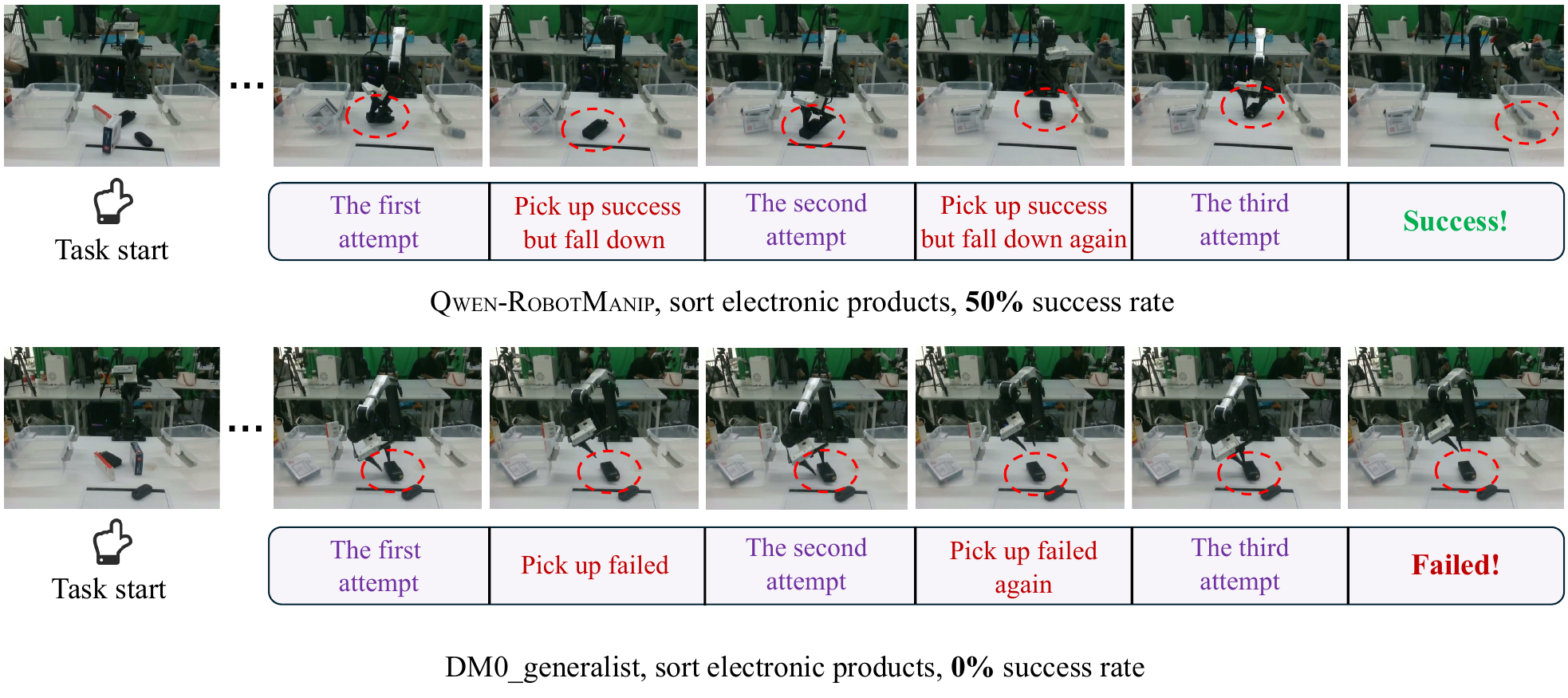}
    \caption{\textbf{Case study on \textit{sort electronic products} (ARX5).} \ours (top) picks up the object but it falls twice; the policy autonomously retries and succeeds on the third attempt. DM0\_generalist (bottom) also attempts three times but never achieves a secure grasp.}
    \label{fig:case_retry}
\end{figure}

\paragraph{Emergent retry behavior.}
A recurring pattern we observe during real-robot evaluation is that \ours exhibits spontaneous retry behavior: when an initial manipulation attempt fails (e.g., a grasp slips or a placement misses) the policy autonomously re-attempts the action rather than proceeding to the next step or stalling. While this behavior is difficult to quantify with a single metric, we observe it consistently across diverse action primitives including picking, placing, pouring, folding, wiping, and sweeping. This self-corrective capability significantly improves fault tolerance, allowing \ours to recover from intermediate failures that would otherwise cause task-level failures. Figure~\ref{fig:case_retry} provides a vivid example on \textit{sort electronic products}: \ours successfully picks up the object on its first attempt, but the object falls; it then re-attempts, and the object falls again; on the third attempt, \ours finally completes the grasp and places the object into the target bin, achieving task success (50\% SR). In contrast, DM0\_generalist (0\% SR) also attempts the pick three times but fails every attempt---the gripper never achieves a secure grasp, leading to task failure. We hypothesize that this retry behavior emerges from the diversity of the pre-training data, where demonstrations naturally contain imperfect attempts followed by corrections, enabling the model to learn recovery strategies as part of its policy.

\begin{figure}[!t]
\centering
\includegraphics[width=0.95\textwidth]{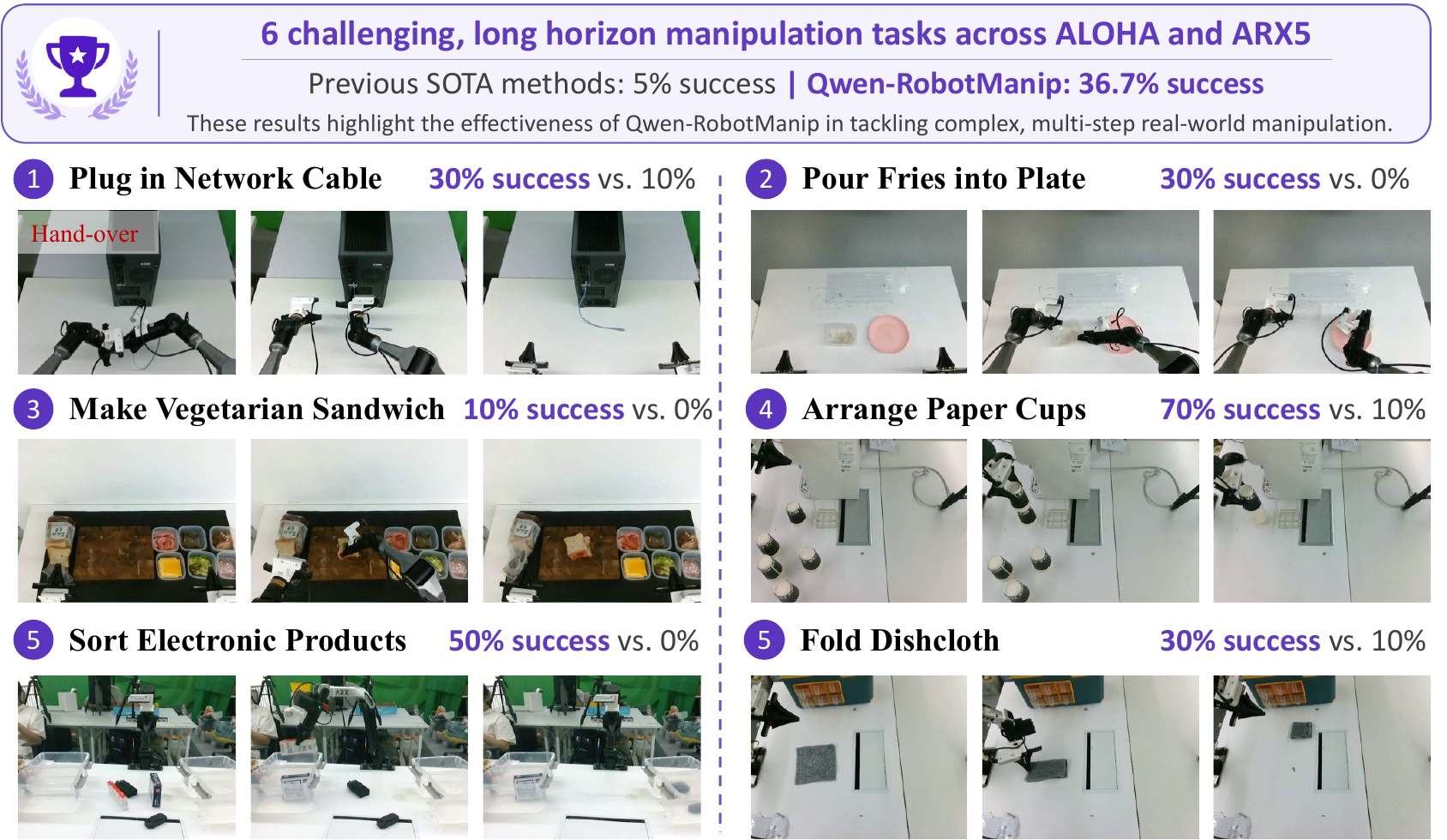}
\caption{\textbf{\ours on six challenging long-horizon tasks from RoboChallenge Table30-v1}, spanning bimanual dexterous manipulation, sequential ingredient stacking, multi-object arrangement, and deformable object handling. Prior SOTA generalist methods achieve an average of 5\% success across these tasks. \ours achieves 36.7\%, with substantial margins on every task.}
\label{fig:challenging_tasks}
\end{figure}


\textbf{Breakthrough on challenging tasks.}
The above three capabilities jointly enable \ours to achieve leading performance on several high-difficulty tasks where most generalist baselines fail entirely (see Figure~\ref{fig:challenging_tasks}). On the ALOHA platform, \textit{plug in network cable} (30\%) and \textit{pour fries into plate} (30\%) both demand precise bimanual coordination with fine posture adjustment, yet \ours is the only model achieving $\geq$30\% success rate. On \textit{make vegetarian sandwich} (10\%)---a long-horizon task requiring sequential ingredient stacking---\ours is the only model to achieve a non-zero success rate. On ARX5, spatial precision and self-correction combine to yield strong results on \textit{arrange paper cups} (70\%), \textit{sort electronic products} (50\%), and \textit{fold dishcloth} (30\%), outperforming the next-best method by 60, 40, and 20 percentage points respectively. These results demonstrate that cross-embodiment pre-training enables the interleaved acquisition of bimanual coordination, spatial precision, and self-corrective strategies, which collectively unlock complex manipulation skills that remain out of reach for existing generalist policies.


\subsection{Ablation Study}

We ablate the core design choices of \ours: alignment strategies (state-action representation, in-context adaptation, architecture) and data recipes for scaling (human-to-robot synthesis, vision-language co-training). All variants within each group share identical training conditions at reduced pretraining scale; configurations may differ from the main results.

\textbf{Action space alignment for data scaling. } A central promise of foundation models is that performance should improve predictably as training data grows. For vision-language-action models trained on heterogeneous cross-embodiment data, verifying this property requires careful design of how data from distinct robots is combined. If the action representations across embodiments are misaligned, simply adding more data may not yield increasing performance, because the model must spend capacity reconciling conflicting conventions rather than learning transferable manipulation structure.

To investigate this, we construct a controlled data-scaling experiment. Starting from the full cross-embodiment pre-training mixture, we create nested subsets at 1\%, 5\%, 10\%, 25\%, 50\%, and 100\% of the original dataset by sampling at the per-embodiment task level, ensuring that smaller subsets are strict subsets of larger ones. All models are evaluated on a fixed held-out OOD evaluation set spanning 15 embodiment types and 154 tasks that are entirely absent from any training set. For each data percentage and model variant, we report the best validation MSE achieved across all training checkpoints, measured in a unified action representation to ensure fair comparison across model variants.

\begin{figure}[!t]
    \centering
    \includegraphics[width=1.0\linewidth]{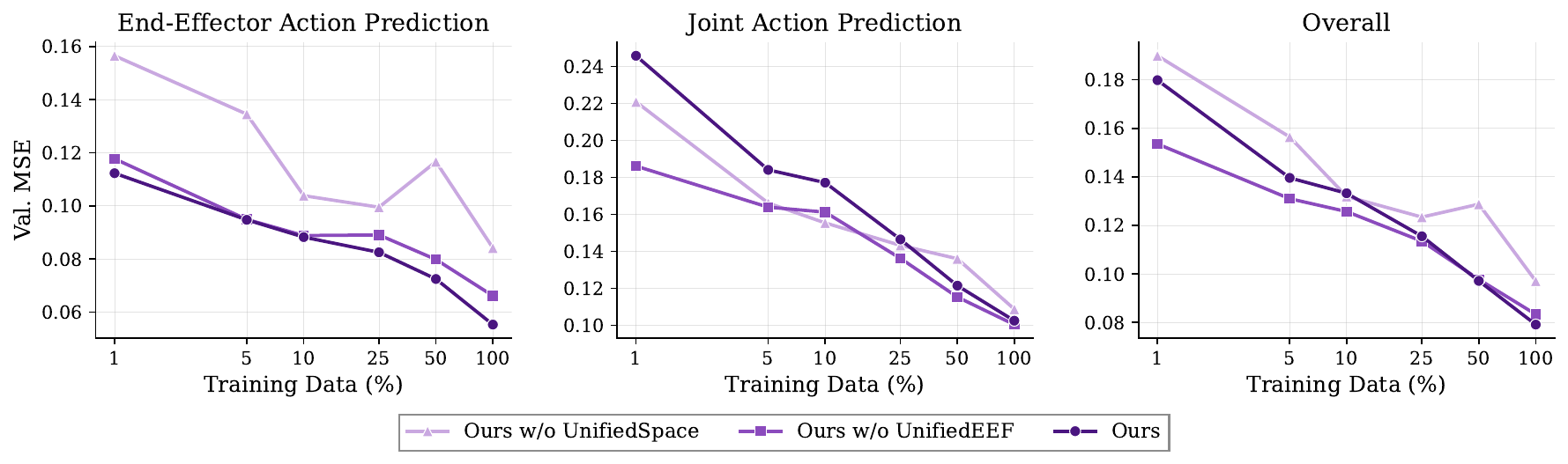}
    \caption{\textbf{Data scaling curves for models with varied state and action representations,} evaluated on held-out validation datasets. Each point reports the best validation MSE across all training checkpoints for a given training data percentage.}
    \label{fig:scaling_curves}
\end{figure}


\begin{figure}[!t]
    \centering
    \includegraphics[width=1.0\linewidth]{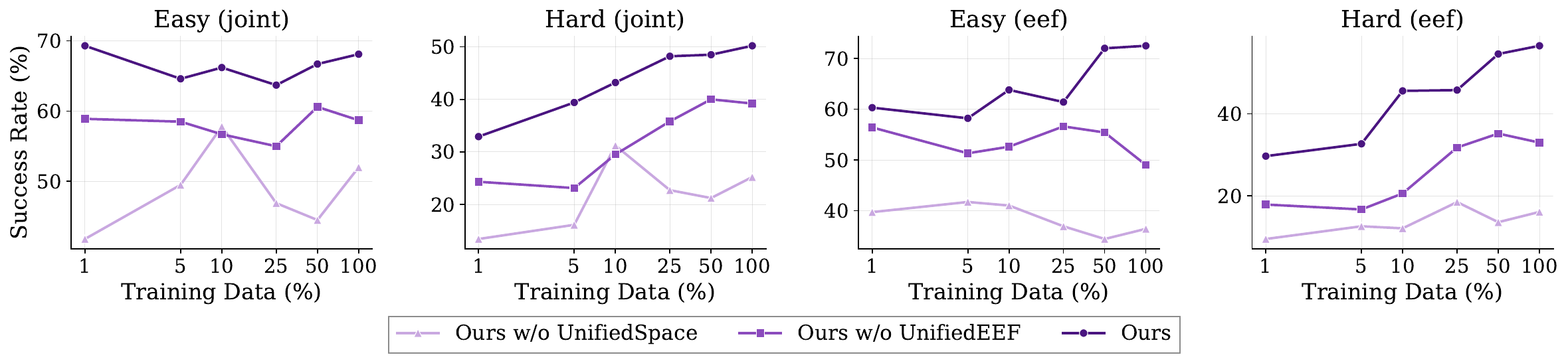}
     \caption{\textbf{Downstream performance on RoboTwin-C2R after fine-tuning models pre-trained with varied data percentages and action
  representations.} Each subplot reports success rate under a different control mode (joint or EEF) and evaluation setting (Easy or Hard).}
    \label{fig:scaling_curves_downstream}
\end{figure}

Figure~\ref{fig:scaling_curves} compares the scaling behavior of three action space designs. \textbf{Ours w/o UnifiedSpace} concatenates each embodiment's raw action fields and zero-pads to 80 dimensions without any cross-embodiment structural alignment. \textbf{Ours w/o UnifiedEEF} maps joints, end-effectors, grippers, and hands into semantically fixed positions within the 80-dimensional canonical vector (\cref{sec:unified_state_action}), with end-effector actions expressed as axis-angle deltas relative to the initial end-effector pose. \textbf{Ours} further unifies end-effector motion by expressing it as camera-frame delta poses (\cref{subsec:cam_delta_action}), grounding actions in the visual observation frame.

Models with unified representations, Ours w/o UnifiedEEF and Ours, both exhibit a clear data scaling law: their best validation MSE decreases approximately log-linearly with training data volume across the full 1\%--100\% range, confirming that with proper action-space alignment, scaling up cross-embodiment data consistently reduces OOD prediction error. Ours w/o UnifiedSpace, by contrast, shows notably different behavior. For action prediction on end-effector dimensions (Figure~\ref{fig:scaling_curves}, left), Ours w/o UnifiedSpace produces an unstable scaling curve with substantially higher MSE than the two unified variants. Ours achieves the lowest end-effector MSE on average, demonstrating that camera-frame alignment enables the most effective cross-embodiment transfer for end-effector control. 


These offline prediction improvements translate directly into downstream task performance. Figure~\ref{fig:scaling_curves_downstream} reports success rates on RoboTwin-C2R after fine-tuning each pre-trained variant on the RoboTwin Clean dataset for 80k steps, evaluated under four settings that combine two control modes (joint-space and end-effector) with two perturbation settings (Easy and Hard). Colors from dark to light correspond to progressively less aligned action representations: Ours, Ours w/o UnifiedEEF, and Ours w/o UnifiedSpace.
On the OOD Hard settings (second and fourth subplots), Ours exhibits a clear data scaling property: downstream success rate increases steadily as pretraining data grows from 1\% to 100\%, reaching 50.2\% (joint) and 56.6\% (EEF) at full scale. Ours consistently outperforms both ablated variants across all data percentages, while Ours w/o UnifiedEEF and Ours w/o UnifiedSpace show noisier scaling curves with less consistent improvement. In contrast, on the in-distribution Easy settings (first and third subplots), none of the three variants shows a clear upward trend with increasing pretraining data. This corroborates the finding in \cref{subsec:are_std_bench_enough} that standard in-distribution evaluation fails to capture the benefits of large-scale pretraining, and that OOD evaluation is needed to reveal the genuine scaling behavior. 
A further notable pattern emerges in the control-mode comparison: Ours achieves comparable or higher success rates under EEF control than under joint control on both Easy and Hard, while both ablated variants show the opposite trend. This confirms that camera-frame alignment produces a strong EEF-space policy, a capability essential for cross-embodiment transfer.

\textbf{Ablation study on embodiment prompt design and in-context policy adaptation.}
We ablate the embodiment prompt design and in-context policy adaptation mechanism using an early checkpoint of \ours trained on a smaller data subset, evaluated on RoboTwin-Clean2Rand under out-of-distribution settings. Results are reported in Table~\ref{tab:ablation_prompt_context}.

\begin{table}[!t]
\centering
\caption{\textbf{Ablation study on embodiment prompt design and in-context policy adaptation.} Out-of-distribution evaluation on RoboTwin-Clean2Rand (joint). The speed is set to 500.}
\label{tab:ablation_prompt_context}
\small
\tabcolsep 5pt
\begin{tabular}{l|cccc|ccc}
\toprule
& \textbf{Emb. Tag} & \textbf{FPS} & \textbf{Context} & \textbf{Denoise Steps} & \textbf{Easy} & \textbf{Hard} & \textbf{Avg} \\
\midrule
\ours w/o UnifiedEEF        & \texttimes & \texttimes & \texttimes & 4 & 71.2 & 54.2 & 62.7 \\
+ Soft-Prompt        & soft       & \texttimes & \texttimes & 4 & 70.2 & 52.1 & 61.2 \\
+ Language Prompt    & lang.      & 15 & \texttimes & 4 & 71.7 & 55.1 & 63.4 \\
+ Structure Prompt   & \checkmark & 15         & \texttimes & 4 & 73.4 & 58.3 & 65.9 \\
\midrule
\ours-Context        & \checkmark & 15 & \checkmark & 4  & 72.1 & 54.4 & 63.3 \\
\ours-Context        & \checkmark & 15 & \checkmark & 10 & 80.1 & 61.6 & 70.9 \\
\ours-Context        & \checkmark & 15 & \checkmark & 20 & 79.8 & 62.1 & 71.0 \\
\bottomrule
\end{tabular}
\end{table}

On the prompt side, encoding embodiment identity as a learnable soft prompt slightly degrades performance relative to the na\"ive baseline. A natural language prompt recovers this and yields a modest gain, suggesting that semantic embodiment labels carry some useful signal, but the improvement remains small, indicating that coarse identity information alone is insufficient to meaningfully adjust behavior across embodiments. The structured embodiment prompt, which additionally conditions the model on the temporal properties of the episode via the FPS field, produces consistent improvements of 2.2-3.2 points.
The structured prompt provides robustness under distribution shift rather than overfitting to a fixed temporal configuration, but is most effective when its fields faithfully reflect the deployment context.

Prompt-level conditioning, however, captures only static episode metadata. The in-context policy adaptation mechanism introduces dynamic intra-episode information in the form of a single historical observation-action chunk, enabling the model to observe the robot's actual behavioral dynamics rather than relying on coarse category labels. Without context, the policy action distribution is smooth and 4 denoising steps is already sufficient, producing stable behavior with no signs of jitter. Introducing execution history increases the complexity of the action distribution, causing the policy to produce jittery motion at the same 4-step budget, which largely negates the benefit of the context signal and yields performance near the na\"ive baseline. Increasing the denoising budget to 10 steps resolves the instability and unlocks the full benefit of the context mechanism, reaching 70.9 average and a 5.0-point improvement over the structure prompt baseline, a margin that dwarfs the contribution of any prompt design variant. Further increasing to 20 steps yields no additional gain. These results support the view that in-context history functions as an implicit embodiment identifier reflecting intra-episode kinematic signatures rather than as episodic memory, and that this signal is qualitatively distinct from what static prompt conditioning can provide, provided the action head has sufficient capacity to decode it faithfully. 

We do note one practical limitation observed in real-robot deployment: at the start of an episode, the context consists entirely of zero-padded placeholders, and the model, having learned to condition on quiescent history, tends to hesitate before initiating motion. We therefore release both the context-conditioned and context-free variants of \ours, allowing users to select appropriate configuration depending on whether rapid response or richer intra-episode adaptation is the priority.

\textbf{Ablation study on human-to-robot synthetic data.}
To isolate the contribution of egocentric human data, we compare three pretraining configurations on RoboTwin-Clean2Rand (eef) and LIBERO-Plus at a fixed 7:3 robot-to-auxiliary ratio (Tables~\ref{tab:h2r_robotwin} and~\ref{tab:h2r_libero}): \textbf{Robot-only} uses robot data only; \textbf{+Ego} mixes in raw egocentric data; \textbf{+H2R} replaces the raw data with pipeline-synthesized robot demonstrations from the same ego sources. 
All three share identical training steps, hyperparameters, and finetuning data, so any performance difference is attributable solely to the auxiliary data source.

On RoboTwin-Clean2Rand, the Hard setting---where all perturbations are applied simultaneously---shows the clearest separation: +H2R reaches 58.7\%, a +4.0 gain over Robot-only (54.7\%) and +3.7 over +Ego (55.0\%). 
Per-dimension gains are consistent, with the largest on Light (+3.0) and Height (+3.2), where the varied camera perspectives and lighting in ego data naturally augment robustness. 
The Easy setting also sees a modest improvement (72.9 $\to$ 73.4 $\to$ 74.2).

On LIBERO-Plus, +H2R raises the average success rate from 87.1\% to 89.0\%. 
The Camera dimension shows the largest improvement (+7.2 over Robot-only, 72.8 $\to$ 80.0), as egocentric data provides diverse viewpoint coverage that robot-only data lacks. 
The Robot dimension also benefits (+2.0), suggesting that the diverse manipulation trajectories in H2R data improve robustness to initial pose variation.
The monotonic Robot-only $\to$ +Ego $\to$ +H2R progression across both benchmarks confirms that raw ego data contributes through visual diversity, while the H2R pipeline unlocks additional gains via action and visual alignment.

\begin{table}[!t]
\centering
\caption{\textbf{Human2Robot ablation on RoboTwin-Clean2Rand (eef).}}
\label{tab:h2r_robotwin}
\small
\tabcolsep 3pt
\begin{tabular}{lcccccc}
\toprule
& \textbf{Easy} & \textbf{Background} & \textbf{Light} & \textbf{Clutter} & \textbf{Height} & \textbf{Hard} \\
\midrule
\ours (robot-only) & 72.9 & 70.4 & 70.3 & 57.2 & 67.8 & 54.7 \\
\ours (+ego) & 73.4 & 70.6 & 71.7 & \textbf{59.2} & 70.2 & 55.0 \\
\ours (+h2r) & \textbf{74.2} & \textbf{71.4} & \textbf{73.3} & 58.1 & \textbf{71.0} & \textbf{58.7} \\
\bottomrule
\end{tabular}
\end{table}

\begin{table}[!t]
\centering
\caption{\textbf{Human2Robot ablation on LIBERO-Plus.}}
\label{tab:h2r_libero}
\small
\tabcolsep 3pt
\begin{tabular}{lcccccccc}
\toprule
& \textbf{Camera} & \textbf{Robot} & \textbf{Language} & \textbf{Light} & \textbf{Background} & \textbf{Noise} & \textbf{Layout} & \textbf{Total} \\
\midrule
\ours (robot-only) & 72.8 & 78.2 & 88.7 & 97.5 & 97.6 & 95.4 & \textbf{85.4} & 87.1 \\
\ours (+ego) & 77.7 & 79.0 & 88.5 & \textbf{98.8} & \textbf{98.7} & \textbf{97.3} & 84.9 & 88.4 \\
\ours (+h2r) & \textbf{80.0} & \textbf{80.2} & \textbf{89.3} & 98.5 & 98.2 & 96.6 & 85.2 & \textbf{89.0} \\
\bottomrule
\end{tabular}
\end{table}

\textbf{Effect of VL data co-training during pre-training.} 
We study the role of VL data in the pre-training stage by comparing our full model with a variant pre-trained without any VL data mixture (Table~\ref{tab:vl_data_in_pretraining_ablation}). On LIBERO and LIBERO-Plus, removing VL data causes relatively small drops of 0.9 and 1.2 points, respectively. In contrast, the performance degradation is much larger on the more challenging RoboTwin benchmarks: RoboTwin-Clean2Rand (easy) drops by 6.7 points, RoboTwin-Clean2Rand (hard) by 8.2 points, and RoboTwin-IF by 7.0 points. This trend indicates that VL co-training becomes increasingly important as task diversity, scene complexity, and distribution shift grow.

We attribute these gains to the complementary benefits of different VL data sources. General-domain VL data helps preserve the VLM's broad perceptual and linguistic capabilities, reducing catastrophic forgetting during VLA training. Spatial grounding and reasoning data improves the model's ability to localize objects, understand spatial relations, and reason over complex scenes, which is crucial for precise manipulation. Embodied-centric VL data further aligns the VLM's representations with the semantic structure of embodied tasks. Together, these data sources yield VLM representations that are better suited for downstream alignment with the action expert.

\begin{table}[!t]
\centering
\caption{\textbf{Ablations on VL data co-training in the pre-training and post-training stage.} By default, \ours is pre-trained with VL data and post-training without it. (``RT'' is the abbreviation of ``RoboTwin''.)}
\label{tab:vl_data_in_pretraining_ablation}
\small
\tabcolsep 2.5pt
\begin{tabular}{lccccc}
\toprule
& \textbf{LIBERO} & \textbf{LIBERO-Plus} & \textbf{RT-C2R (easy)} & \textbf{RT-C2R (hard)} & \textbf{RT-IF} \\
\midrule
\ours & \textbf{99.1} & 90.1 & 73.2 & \textbf{62.6} & 71.6 \\
\quad - without VL data in pre-training & 98.2 & 88.9 & 66.5 & 54.4 & 64.6 \\
\quad - with VL data in post-training & 98.6 & \textbf{91.4} & \textbf{74.0} & 62.5 & \textbf{73.1} \\
\bottomrule
\end{tabular}
\end{table}

\begin{figure}[!t]
    \centering
    \includegraphics[width=0.8\linewidth]{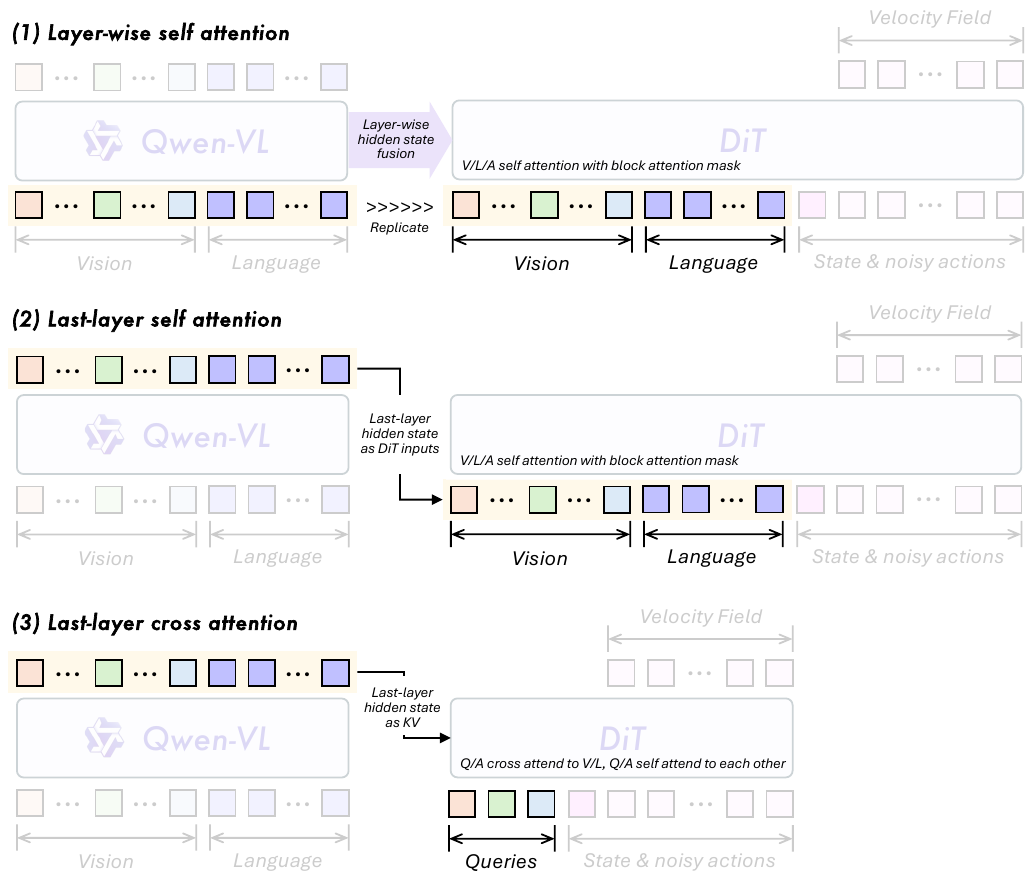}
    \caption{\textbf{Ablation study of model architecture variants.} }
    \label{fig:ablate-nn-arch}
\end{figure}

\textbf{Architecture design. }
We conduct ablation experiments on three network architecture variants (Figure~\ref{fig:ablate-nn-arch}):
\begin{enumerate}[leftmargin=*,topsep=2pt,itemsep=1pt]
    \item The first variant replicates the VLM's vision and language hidden states as input to the DiT, and employs a pure self-attention architecture within the DiT. After each self-attention layer in the DiT, the vision and language features are fused with the corresponding per-layer vision and language features from the VLM via a weighted residual combination.
    \item The second variant feeds only the VLM's \emph{last-layer} hidden states into the DiT. The DiT likewise uses a pure self-attention architecture, but without any per-layer feature fusion with the VLM's intermediate representations.
    \item The third variant introduces a small set of learnable query tokens~\citep{ICLR2024Registers} that act as condensed proxies for the VLM's vision and language tokens. 
    These query tokens are concatenated with the state and action tokens; all tokens jointly cross-attend to the VLM's last-layer hidden states and self-attend among themselves. The query tokens' outputs are discarded after processing.
\end{enumerate}

All three variants are pretrained exclusively on robot data and then fine-tuned on the LIBERO dataset. During both pretraining and fine-tuning, we exclude the structured embodiment prompt (retaining only the original task instruction), ECoT data, and in-context policy adaptation, isolating the camera-frame delta pose representation as the sole alignment design retained in this comparison.

The results are presented in Table~\ref{tab:libero_plus_diff_arch}. On LIBERO-Plus, the third variant (last-layer cross-attention) achieves the highest average success rate (87.5\%) while incurring the lowest computational cost, as it avoids both per-layer feature fusion and storing the full set of VLM vision-language tokens in the DiT. The pure self-attention architectures, whether with per-layer or last-layer fusion, do not exhibit a clear advantage for vision-language-action feature interaction. We therefore adopt the third architecture in all subsequent experiments.

\begin{table}[!t]
\centering
\caption{\textbf{Evaluation on LIBERO-Plus for different network architecture variants.}}
\label{tab:libero_plus_diff_arch}
\small
\tabcolsep 3pt
\begin{tabular}{lcccccccc}
\toprule
& \textbf{Camera} & \textbf{Robot} & \textbf{Language} & \textbf{Light} & \textbf{Background} & \textbf{Noise} & \textbf{Layout} & \textbf{Total} \\
\midrule
layer-wise self attention & 74.5 & 74.3 & 86.8 & 98.6 & 96.0 & 94.9 & 86.1 & 86.4 \\
last-layer self attention & 75.5 & 71.4 & 87.2 & 98.3 & 98.9 & 96.1 & 88.1 & 87.0 \\
last-layer cross attention & 76.9 & 73.4 & 86.9 & 98.9 & 97.7 & 97.3 & 87.3 & {\textbf{87.5}} \\
\bottomrule
\end{tabular}
\end{table}


\subsection{New Features after Alignment}

\subsubsection{Post-train with VL Data and VLA Data in Pre-training Recipe}
\label{sec:mixposttrain}

\textbf{Setting 1: Post-training with VL data co-training.} In our main experiments, to match the standard fine-tuning setting, VL data is used only during pre-training and excluded from post-training. Here, we further investigate whether incorporating VL data into the post-training stage can improve generalization.

As shown in Table~\ref{tab:vl_data_in_pretraining_ablation} (see ``- with VL data in post-training''), adding VL data during post-training leads to improved performance on several benchmarks. In particular, it improves LIBERO-Plus from 90.1 to \textbf{91.4}, RoboTwin-Clean2Rand (easy) from 73.2 to \textbf{74.0}, and RoboTwin-IF from 71.6 to \textbf{73.1}, while the performance on RoboTwin-Clean2Rand (hard) remains nearly unchanged (62.6 vs.\ 62.5). These results suggest that post-training with VL data is particularly helpful for out-of-distribution generalization.

A closer analysis shows that the gain is most evident on \textit{language-related generalization benchmarks}. For example, on LIBERO-Plus, the success rate under language perturbations increases from 86.9\% for \ours to 93.9\% when VL data is included during post-training. Similarly, on RoboTwin-IF, post-training with VL data improves the success rate from 76\% to 81\% on the ``Pick-Diverse-Object'' suite. In contrast, RoboTwin-Clean2Rand (hard) keeps the instructions fixed and instead introduces visual distractors, background changes, and lighting variations, so the benefit of VL data is less pronounced in this setting.


We attribute these improvements to better preservation of the base VLM's foundational capabilities. Fine-tuning solely on action prediction can erode the pretrained VLM's language understanding and visual grounding abilities due to catastrophic forgetting, thereby weakening its capacity to interpret novel instructions at test time. In contrast, mixing VL data into post-training helps maintain these capabilities, enabling the model to better parse unseen instructions, ground referring expressions to visual entities, and translate linguistic intent into appropriate actions.

\begin{figure*}[!t]
  \centering
  
  \begin{minipage}{0.48\textwidth}
    \centering
    \includegraphics[width=0.85\linewidth]{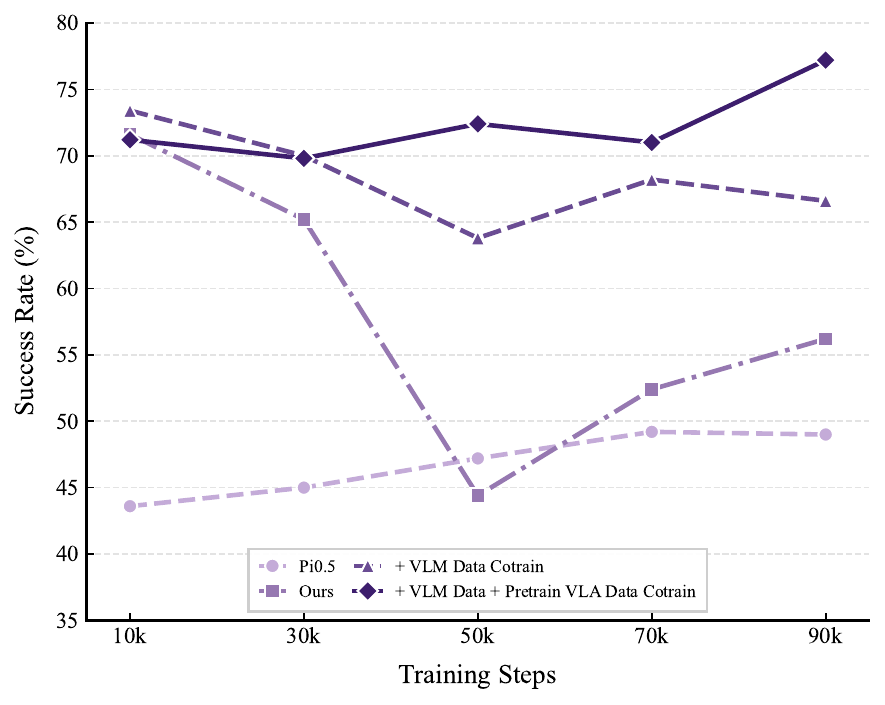}
    \caption{
      \textbf{Performance comparison under different proportions of training data in RoboTwin-IF.}
      We report the score at 10k, 30k, 50k, 70k, and 90k training steps.
      Compared with the baseline \texttt{pi05}, our method benefits from
      incorporating VLM data and further improves when co-training with both
      VLM data and VLA pretraining data.
    }
    \label{fig:mixposttrain}
  \end{minipage}\hfill
  \begin{minipage}{0.48\textwidth}
    \centering
    \includegraphics[width=0.85\linewidth]{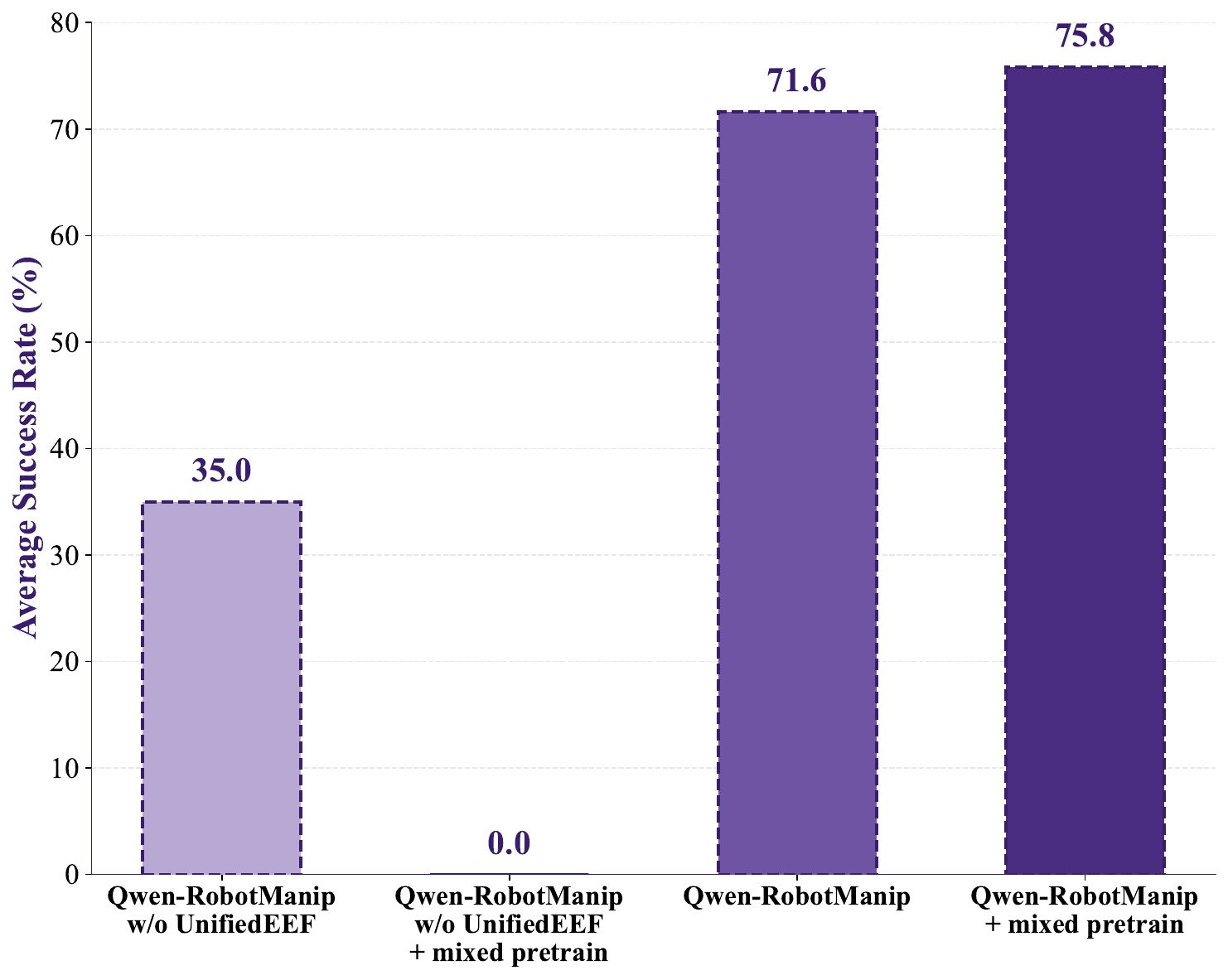}
    \caption{
      \textbf{Instruction-following success rates across configurations.} The chart highlights the critical necessity of the UnifiedEEF module, showing severe degradation without it (35.0\%) and total failure when combined with mixed pretraining (0.0\%). The optimal configuration utilizes both the base architecture and mixed pretraining (75.8\%).
    }
    \label{fig:mixedposttrain_ablation}
  \end{minipage}
  
\end{figure*}



\textbf{Setting 2: Post-training with VL and auxiliary VLA data co-training.} When transferring VLA models to a specific downstream domain via post-training such as a novel simulation environment, robot embodiment, or operational scene—it is standard practice to rely exclusively on clean, domain-specific datasets. However, this paradigm introduces a prominent bottleneck in VLA adaptation: severe domain overfitting. In most scenarios, rather than acquiring genuine, generalizable instruction-following capabilities, the model essentially memorizes the specific scene dynamics. Leveraging our proposed unified action space representation, we introduce a mixed post-training strategy to systematically enhance the model's generalization capacity within the target domain. Specifically, during the curation of the post-training dataset, we augment the target-domain data by integrating a mixture of general Vision-Language (VL) data and pre-training VLA data. Crucially, this mixed-data approach effectively mitigates overfitting without compromising the model's low-level physical execution proficiency in the target domain.

Specifically, our augmented dataset comprises two primary categories. The first consists of Vision-Language (VL) data, predominantly derived from standard Vision-Language Model (VLM) training corpora, encompassing tasks such as object detection, spatial pointing, and trajectory prediction. The second category entails VLA data, which aggregates trajectories from auxiliary simulators alongside large-scale, real-world demonstrations from morphologically similar embodiments. To facilitate effective cross-embodiment transfer, we align and optimize this auxiliary VLA data within the highly transferable end-effector (EEF) action space.

We evaluate our approach on RoboTwin-IF, the Out-of-Distribution benchmark within our target domain, RoboTwin. As shown in Figure~\ref{fig:mixposttrain}, when fine-tuning exclusively on the domain-specific RoboTwin-Clean dataset, our model initially achieves performance that significantly surpasses that of $\pi_{0.5}$. However, as the number of training steps increases, severe overfitting emerges, leading to a progressive performance decay on the RoboTwin-IF evaluation. By integrating VL data (accounting for 10\% of the training mixture), both the instability of the peak performance and the subsequent degradation are substantially mitigated. Furthermore, when we expand the mixture to include the auxiliary VLA data (comprising 75\% of the total vla data in the training mixture), the overfitting phenomenon is entirely eradicated. Notably, as training progresses, the model's performance on RoboTwin-IF exhibits continuous improvement rather than decay, ultimately achieving comparable peak performance while maintaining robust OOD generalization. Besides, as shown in Figure~\ref{fig:mixedposttrain_ablation}, the most critical insight is that the \textit{UnifiedEEF} module serves as the \textbf{exclusive prerequisite} for mixed post-training. While simply removing the module degrades baseline performance (from 71.6\% to 35.0\%), attempting to introduce mixed post-training data without it triggers a complete representation collapse (0.0\% success rate). This catastrophic failure fundamentally proves that the model is entirely incapable of assimilating mixed VLM/VLA data on its own. It is \textbf{only} through the structural alignment provided by \textit{UnifiedEEF} that the architecture can unlock the benefits of mixed post-training, successfully pushing the performance ceiling to 75.8\%. 

\subsubsection{EEF control \& Cross-Embodiment Transfer}
Camera-frame delta EEF expresses actions as pose deltas in the visual observation coordinate system, making the action representation inherently shared across embodiments regardless of their underlying kinematics. We evaluate this design along three complementary dimensions---in-distribution control quality, compositional skill transfer, and zero-shot cross-embodiment generalization---and consolidate the key results in Table~\ref{tab:eef_summary}.

\begin{table}[!t]
\centering
\caption{\textbf{Systematic advantages of camera-frame delta EEF} across three evaluation dimensions, progressing from in-distribution to zero-shot settings.}
\label{tab:eef_summary}
\small
\resizebox{\columnwidth}{!}{
\begin{tabular}{llccc}
\toprule
\textbf{Dimension} & \textbf{Evaluation} & \textbf{Best baseline} & \textbf{\ours} & \textbf{Gain} \\
\midrule
EEF control quality & RT-C2R Easy / Hard (eef mode) & 49.0 / 33.0 (w/o UnifiedEEF) & \textbf{72.5 / 56.6} & +23.5 / +23.6 \\
Skill composition & CobotMagic$\to$ARX, 4 novel tasks & 12.5\% (w/o UnifiedEEF) & \textbf{55.0\%} & $4.4\times$ \\
Zero-shot transfer & AgileX$\to$ARX, UR5, Franka (avg) & 14.5\% (joint) & \textbf{23.9\%} (eef) & $1.65\times$ \\
\bottomrule
\end{tabular}
}
\end{table}

\textbf{In-distribution EEF control.} As the first row of Table~\ref{tab:eef_summary} shows, \ours with camera-frame EEF achieves 72.5\% / 56.6\% on RoboTwin-C2R Easy / Hard under EEF-mode execution, compared with 49.0\% / 33.0\% for the best alternative action-space design (\ours w/o UnifiedEEF). More notably, \ours is the \emph{only} variant where EEF-mode execution surpasses its own joint-mode performance (72.5\% vs.\ 68.1\% Easy, 56.6\% vs.\ 50.2\% Hard, see Figure~\ref{fig:scaling_curves_downstream}); all other action-space designs degrade when switching from joint to EEF control. This reversal indicates that camera-frame alignment produces a genuinely strong EEF-space policy rather than merely an alternative control interface. Experiments for data scaling (Figure~\ref{fig:scaling_curves}) corroborates this finding: under the unified EEF representation, cross-embodiment data follows a clean log-linear scaling law, whereas the ablated method yields an erratic curve with substantially higher prediction error.

\textbf{Compositional skill transfer.} Camera-frame EEF decouples manipulation skills from robot-specific kinematics, enabling skill-level composition across embodiment--task combinations. In a joint-training experiment with 6K CobotMagic and 130 ARX demonstrations, the policy is evaluated on four novel ARX tasks for which zero target-task demonstrations exist. As the second row of Table~\ref{tab:eef_summary} shows, \ours achieves 55.0\%---$4.4\times$ the best ablated variant (\ours w/o UnifiedEEF, 12.5\%)---while \ours w/o UnifiedSpace manages only 7.5\%. Because camera-frame deltas map the same manipulation primitive to a consistent numerical pattern regardless of the executing robot, the model acquires embodiment-agnostic skill representations that compose freely with new embodiment--task pairings (per-task breakdown in Table~\ref{tab:cross_embodiment_transfer}).

\textbf{Zero-shot cross-embodiment transfer.} The most demanding test deploys a policy trained exclusively on AgileX to three unseen embodiments---ARX, UR5, and Franka---without any target-embodiment data. As the third row of Table~\ref{tab:eef_summary} shows, EEF control achieves an average success rate of 23.9\%, far exceeding the 14.5\% of joint control. The improvement is especially pronounced on UR5, where EEF mode reaches 22.8\% versus 4.1\% for joint mode ($5.6\times$). Joint-space actions are inherently robot-specific and produce near-random behavior on unseen morphologies, whereas camera-frame deltas abstract away kinematic differences and enable meaningful transfer in shared Cartesian space (per-embodiment results in Table~\ref{tab:cross_embodiment}).

Taken together, the three rows of Table~\ref{tab:eef_summary} reveal a coherent progression: camera-frame delta EEF first strengthens in-distribution EEF control, then enables cross-embodiment skill composition, and finally supports zero-shot deployment on unseen robots---confirming that it provides an embodiment-agnostic action interface whose benefits compound across increasingly challenging transfer settings.

\clearpage

\section{Conclusion}
\label{sec:conclusion}

This report set out to investigate whether the scaling recipe behind large language and multimodal models, aligning heterogeneous data under a unified formulation and training at scale, can be applied to robotic manipulation to achieve genuine generalization. We present \ours as an affirmative answer, built on the principle that alignment and scale are not independent engineering challenges but tightly coupled prerequisites: without a unified cross-embodiment formulation, scaling data produces conflicts rather than synergy; without sufficient data diversity, even a well-aligned model cannot generalize beyond its training distribution.

Several findings from this work carry implications beyond the specific system we describe. First, the unified alignment framework, spanning a canonical state-action representation, camera-frame delta pose parameterization, and in-context policy adaptation, proves critical not merely for accommodating diverse embodiments but for enabling data scaling itself. Our ablations show that na\"ive representations fail to exhibit scaling behavior; alignment is what converts additional data volume into improved capability. Second, the fact that \ours constructs a $\sim$38,100-hour corpus and achieves emergent generalization capabilities using \emph{only} open-source robotic manipulation datasets and egocentric human videos, without any proprietary data collection, suggests that the data barrier for manipulation foundation models may be lower than commonly assumed, provided the right synthesis and curation infrastructure is in place. Third, our systematic comparison between in-domain and out-of-distribution evaluation reveals that standard benchmarks consistently fail to distinguish models whose pretraining contributes genuine generalizable structure from those that succeed through in-distribution pattern matching. We believe the OOD evaluation settings introduced in this work provide a more faithful measure of robotic foundation model capability, and we hope they serve as useful diagnostics for the community.

Limitations remain. The human-to-robot synthesis pipeline, while scalable, introduces distributional gaps from retargeting approximations and inpainting artifacts that bound the effective quality of synthesized data. Our OOD evaluations, though substantially more challenging than standard benchmarks, are still predominantly simulation-based. Broader real-world evaluation across deployment conditions is needed. The fixed action chunk length and inference latency of the current system also constrain applicability to tasks requiring reactive sub-second control.

Looking forward, the alignment-then-scale paradigm demonstrated here naturally extends in several directions: incorporating more robot morphologies and task domains into the pretraining corpus, improving synthesis fidelity through more accurate hand-robot retargeting and physically grounded rendering, and incorporating agentic systems toward longer-horizon reasoning and manipulation. Moreover, we hope this work contributes to a shift in how the community evaluates VLA models, from in-domain benchmark rank toward the out-of-distribution generalization that ultimately determines whether a model can serve as a genuine foundation for real-world deployment.

\section{Authors}

\textbf{Core Contributors:} Haoqi Yuan$^*$, Zhixuan Liang$^*$, Anzhe Chen$^*$, Ye Wang$^*$, Haoyang Li$^*$, Pei Lin$^*$, Yiyang Huang$^*$, Zixing Lei$^*$, Tong Zhang$^*$, Jiazhao Zhang, Jie Zhang, Jingyang Fan, Gengze Zhou, Qihang Peng, Chenxu Lv, Xiaoyue Chen, An Yang, Fei Huang, Junyang Lin, Dayiheng Liu, Jingren Zhou, Chenfei Wu$^\dagger$, Xiong-Hui Chen$^{\ddagger\dagger}$

$^*$Equal Contribution.
$^\dagger$ Corresponding Author.
$^\ddagger$ Project Lead.

\textbf{Contributors:} Jinhui Ye, Sicheng Xie, Hale Yin, 
Xudong Guo, Shuai Bai, Lulu Hu, Minying Zhang, Shurui Li, Wenhu Xiao, Yue Wang, Kun Yan, Xiao Xu, Jiahao Li, Xuancheng Ren

\textbf{Acknowledgments:} We acknowledge the National Pilot Base for Embodied Intelligence Applications for providing the real-robot experimental environment and equipment. We thank AgileX Robotics for their hardware support. We also thank Prof. Hao Dong and Prof. Yao Mu for their support.

\clearpage
\bibliography{colm2024_conference}
\bibliographystyle{colm2024_conference}

\end{document}